\documentclass[twoside,11pt]{article}

\usepackage{amsmath}
\usepackage{amsfonts}
\usepackage{amssymb}
\usepackage{bm}
\usepackage{bbm}
\usepackage{mathrsfs}
\usepackage{color}

\usepackage[preprint]{jmlr2e}
\setcitestyle{numbers,square,comma} % numbered citations [1], [2, Theorem 13.2]

\usepackage{lastpage}
\ShortHeadings{Beyond Hidden States}{Gibson and Reisinger}

\begin{document}
\firstpageno{1}

\title{Outputs Converge Faster than Hidden States in Trained Deep ResNets}

\author{\name William Gibson \email william.gibson@maths.ox.ac.uk \\
       \addr Mathematical Institute\\
  University of Oxford\\
  Oxford, OX2 6GG, UK
       \AND
       \name Christoph Reisinger \email christoph.reisinger@maths.ox.ac.uk \\
       \addr Mathematical Institute\\
  University of Oxford\\
  Oxford, OX2 6GG, UK}
% \editor{My editor}

\maketitle

\begin{abstract}%   <- trailing '%' for backward compatibility of .sty file
We consider ResNets of depth $L$, whose residual blocks are two-layer perceptrons of width $M$ acting on an embedding of dimension $D$, trained by clipped gradient descent in the Maximal Local Update (MLU) parameterisation, and compare their outputs with those of the infinitely wide and deep limit. For convex losses, we prove the conjectured upper bound of Chaintron, Chizat and Maass on sharp quantitative convergence rates in depth and width at fixed embedding dimension. Convexity is used only to measure the discrepancy between model outputs in the geometry induced by the Hessian of the loss, which we argue is more appropriate than the Euclidean distance for many learning tasks. With high probability, the resulting pseudometric distance is at most of order $\frac1L+\frac1{\sqrt{LM}}+\frac D{LM}$, up to logarithmic factors depending only on the number of training steps $K$ and the failure probability, for Gaussian initialisation. For non-Gaussian initialisations admitting a square-integrable Stein kernel, an additional $\mathcal O\big(\sqrt{\log D/(LM)}\big)$ term appears. The result holds under the shape constraints $D=\mathcal O\big(M\wedge (LM^{2/3})\big)$, which admit the scaling $D\asymp M\asymp L^2$ consistent with architectures used in practice.
\end{abstract}

\begin{keywords}
Deep Residual Networks, Scaling Limits, Mean-field Limit, Feature Learning, Stein Kernels
\end{keywords}

\section{Introduction}
The development of large neural networks has been largely motivated by empirical scaling laws, which describe how performance improves with model size, data and computational resources \cite{kaplan2020scaling,hoffmann2022training}. These laws tell practitioners how to allocate their fixed computational budget, but their mathematical foundations remain incomplete. In particular, they offer limited explanation of how depth, width and embedding dimension should be balanced. One approach is to study the convergence of finite networks towards their infinite-dimensional limits. Then, optimising sharp convergence rates under a fixed parameter budget provides a theoretical criterion for allocating parameters across the architecture. In this paper, we pursue this approach for ResNets, and show that the resulting scaling laws depend crucially on whether convergence is measured in the hidden states or in the network's predictions.

\paragraph{}We consider ResNets with two-layer perceptron residual blocks \cite{he2016deep}. A network of embedding dimension $D$, width $M$ and depth $L$ embeds an input $x\in \R^{d_{\mathrm{in}}}$ via $W_\mathrm{in}\in \R^{D\times d_{\mathrm{in}}}$, and its hidden states at training step $k$ evolve according to
\[
h_{0,k}=W_{\mathrm{in},k}x,\qquad h_{l,k}=h_{l-1,k}+\frac{1}{LM}\Phi_k^l\bm{\rho}\left(\frac{1}{D}(\Psi_k^l)^\top h_{l-1,k}\right),
\qquad y_k=\frac1D W_{\mathrm{out},k}^\top h_{L,k},
\]
with parameters $\Phi^l,\Psi^l\in\mathbb R^{D\times M}$, where $\rho$ acts coordinatewise. The output of the model is read out from the final hidden state via $W_\mathrm{out}\in \R^{D\times d_\mathrm{out}}$. We train these networks by clipped gradient descent in the maximal local update (MLU) parameterisation (see Section \ref{sec:Setup} and \cite[Section~4.2]{chizat2025hiddenwidthdeepresnets}). In this regime, the residual parameters are initialised at scale $\sqrt D$ and the learning rates are of order one, so that both the decrease of the loss and the updates of the pre-activations $\langle\psi^{lm},h_{l-1}\rangle_{\bar2}$ of the $LM$ neurons are of order one, uniformly in $L$, $M$ and $D$. Accordingly, this network learns features, rather than evolving in the lazy regime, in which these local updates vanish \cite[Section~4.2]{chizat2025hiddenwidthdeepresnets}.

\paragraph{}As the depth and width tend to infinity, the hidden states converge to the solution of a mean ODE, Section \ref{sec:LimitingModel}, whose output at training step $k$ we denote by $Y_k$. Our main theorem is an improved quantitative, high-probability estimate of the discrepancy between $y_k$ and $Y_k$, uniformly over a fixed number of training steps. The choice of discrepancy between $y_k$ and $Y_k$ is important. In the pretraining of large language models, for example, only a small fraction of the vocabulary consists of plausible next tokens, so the coordinates of the output should not be weighted equally, and thus the Euclidean distance between outputs is not appropriate.  Instead, we measure errors using a (pseudo)metric, defined in Section \ref{sec:BD}, that depends on the loss function, whilst remaining independent of the training data. Our choice of pseudometric is natural as it keeps the RMS distance on models trained to minimise a mean squared error loss (Example \ref{ex:MSE}), whereas the metric of Section \ref{sec:BD} becomes the Fisher--Rao metric for the Cross-Entropy loss, and thus respects the fact that the outputs of such models are logits of a probability distribution (Example \ref{ex:CE}). 

\begin{theorem}[Main Result, Informal]\label{thm:informal}
Suppose that the assumptions of Theorem~\ref{thm:MainTheorem} hold, and that the residual parameters are initialised with i.i.d.\ Gaussian entries. Then there exists a constant $C$, independent of $L$, $M$, $D$ and $\delta$, such that, with probability at least $1-\delta$,
\[
\max_{k\le K}d_\ell(y_k,Y_k)\le C\left(\frac1L+\sqrt{\frac{\mathfrak l_\delta}{LM}}+\frac{D}{LM}\right),
\]
where $\mathfrak l_\delta$ is of order $\log((K+1)/\delta)$. For general initialisations with a square-integrable Stein kernel, the same bound holds up to additional terms, given in Theorem~\ref{thm:MainTheorem}, which are proportional to the standard deviation of the Stein kernel of the initialisation. This result holds under the shape and failure probability constraints of \eqref{eqn:shape_main}.
\end{theorem}
The corresponding tight error bounds for the convergence of the hidden states in the RMS norm are larger, of order $L^{-1}+ D^{\frac{1}{2}}(LM)^{-\frac{1}{2}}$, \cite[Theorem~3]{chizat2025hiddenwidthdeepresnets}. The improvement in the output convergence rate arises because the hidden-state error enters the outputs only through finitely many directions. Let $\lambda_{j,k}$ and $u_{j,k}$ be the eigenvalues and eigenvectors of the Hessian of the loss, evaluated at the output $Y_k$ of the limiting model, which exist as $\ell$ is assumed to be $C^2$. The leading order of $d_\ell(y_k,Y_k)$ is then
\[\Big(\sum_j\lambda_{j,k}\,\big\langle W^\infty_{\mathrm{out},k}u_{j,k},\,h_{L,k}-H_{1,k}\big\rangle_{\bar2}^2\Big)^{1/2},\]
a weighted mean of the projections of the hidden-state error onto the directions $W^\infty_{\mathrm{out},k}u_{j,k}$ of the limiting model (Section~\ref{sec:ProofMain}). Each projected error is of order $\frac1L+\frac1{\sqrt{LM}}$, and the $\lambda$-weighted sum of $\|W^\infty_{\mathrm{out},k}u_{j,k}\|_{\bar2}^2$ is bounded, so that the mean inherits this rate. In contrast, the RMS norm of the hidden-state error is the supremum of its projections over all unit directions, which costs the factor $\sqrt D$. The remaining $\mathcal O(\frac D{LM})$ term is the square of the hidden-state error. However, projections onto the eigenvectors of the Hessian alone do not close the recursion over the training steps at this rate, since the errors at the earlier steps re-enter through projections onto the hidden states and adjoint variables of the limiting model. Therefore, the induction in Section~\ref{sec:ProofThmMain} is conducted on the smallest family of projections that closes the argument, namely those onto the limiting hidden states, adjoint variables and output directions, which arise from the propagation estimates of Section~\ref{sec:PROP}.

\subsection{Literature Review}
The convergence of ResNets with residual branches scaled by $L^{-1}$ to the solution of a neural (mean) ODE is well known. However, much of this literature requires correlated parameter initialisation \cite{avelin2020neuralodesdeeplimit,marion2024implicit}, provides loose or no quantitative rates \cite{ding2,gao2024global} or a proportion of the parameters train lazily, such as the neural tangent kernel regime \cite{huang2020deep,hayou2021stable}.

\paragraph{}A fundamental question in the study of large neural networks is how to scale their parameters and learning rates so that feature learning persists in the infinite-dimensional limit. In standard parameterisations, stability may require learning rates that vanish with width, causing the features to remain close to their initial values. The maximal update parameterisation ($\mu$P) of \cite{pmlr-v139-yang21c} instead balances the scales of the initial weights and their updates so that every layer undergoes order-one changes in its pre-activations as width increases. Depth-$\mu$P \cite{yang2024tensor} extends this principle to increasing depth, balancing feature learning with diversity between layers. More recently, CompleteP \cite{dey2026don} addresses the additional possibility of lazy training in deep transformers, and shows that a depth scaling $\mathcal O(L^{-1})$ is required to prevent any non-negligible proportion of the parameters from training lazily. These parameterisations also permit stable hyperparameter transfer across network sizes. We work in the Maximal Local Update (MLU) parameterisation of \cite{chaintron2026resnets,chizat2025hiddenwidthdeepresnets}, which allows depth, width and embedding dimension to grow jointly while retaining nontrivial feature-learning dynamics. These papers \cite{chaintron2026resnets,chizat2025hiddenwidthdeepresnets} establish tight quantitative bounds for the convergence of ResNets in the MLU regime in each limit $D,L,M\to \infty$. However, they observed numerically that the outputs converge faster than the hidden states, and conjectured \cite[Section~1.3]{chaintron2026resnets} that the error in the outputs is
\begin{equation}\label{eqn:conjecture}
\Theta\left(\frac1L+\frac{1}{\sqrt{LM}}+\frac D{LM}+\frac1{\sqrt D}\right).
\end{equation}
Theorem~\ref{thm:informal} establishes the upper bound in the conjectured convergence rate in depth and width at fixed embedding dimension $D$. The additional $D^{-1/2}$ term, arising from the large-embedding-dimension limit, is discussed in Section~\ref{sec:joint_limit} and Remark \ref{rem:joint_D}.

\paragraph{}Each of these results is subject to constraints on the shape of the network. Those of \cite{chizat2025hiddenwidthdeepresnets,chaintron2026resnets} require $D\lesssim M$ and $\log L\lesssim M$ when the backward pass is clipped, and $D\lesssim\sqrt M$ when it is not. Independently of, and contemporaneously with, the present work, \cite{chaintron2026infiniteresnetsgaussianmean} formulated the problem without gradient clipping and deduced that the convergence only requires $D\lesssim LM$, up to logarithmic factors. However, their bound on the outputs is of the form
\[
\frac1L+\Big(\log\frac{LM}{\delta}\Big)^{\frac{K+2}2}\sqrt{\frac D{LM}},
\]
which falls short of the conjectured rate by a factor $\sqrt D$, and in which the power of the logarithmic term grows with the number of training steps. The faster rate is restated there as \cite[Conjecture~1]{chaintron2026infiniteresnetsgaussianmean}. Furthermore, our bounds contain no logarithmic factor in $L$ or $M$. For Gaussian initialisation, the only logarithmic factor is $\log((K+1)/\delta)$. For non-Gaussian initialisation, the terms arising from the Stein kernel carry an additional factor~$\log(D(K+1)/\delta)^{1/2}$. In neither case does the power of a logarithm depend on $K$. We relax the gradient clipping requirements of \cite{chizat2025hiddenwidthdeepresnets,chaintron2026resnets} by removing the clipping in the backward pass. At the same time, we improve the shape constraints to $D\lesssim M$ and $D\lesssim LM^{2/3}$, relaxing the requirement $D\lesssim M^{1/2}$ for the unclipped backward pass in \cite{chizat2025hiddenwidthdeepresnets}, subsequently used in \cite{chaintron2026resnets}, and removing the restriction $\log L\lesssim M$. These constraints admit the scaling $D\asymp M\asymp L^2$ of Section~\ref{sec:joint_limit}. 

\subsection{The Joint Limit and the Shape of the Network}\label{sec:joint_limit}
In addition to the large depth limit, the embedding dimension can also be taken to infinity, provided the width diverges similarly, in which case the joint limit in $D$, $L$ and $M$ is a DMFT-type limit \cite[Theorem~2.5]{chaintron2026resnets}. For sub-Gaussian initialisations, this convergence holds at the rate $\mathcal O(D^{-\frac{1}{2}})$ typical of the Central Limit Theorem.\footnote{The rate is that of the Central Limit Theorem, rather than the law of large numbers, because of the MLU parameterisation.} Informally, combining this with Theorem~\ref{thm:informal} through the triangle inequality, after adapting the arguments of \cite[Theorem~2.5]{chaintron2026resnets} to our assumptions, gives the upper bound in the conjecture \eqref{eqn:conjecture}, in the metric $d_\ell$ and up to a logarithmic factor (Remark~\ref{rem:joint_D}). 

\paragraph{}As motivation, and following \cite{chaintron2026resnets}, we optimise the heuristic bound \eqref{eqn:conjecture} under a fixed parameter budget $P=2DLM+D(d_{\mathrm{in}}+d_{\mathrm{out}})\asymp LMD$, for fixed input and output dimensions. Since $D\lesssim LM$ is required for convergence, the term $(LM)^{-1/2}$ is dominated by $D^{-1/2}$. The embedding dimension $D$ is the only variable with competing effects on the error, and optimising over it gives $D\asymp P^{2/5}$ and $LM\asymp P^{3/5}$. The remaining term $L^{-1}$ is at most of order $P^{-1/5}$ provided $L\gtrsim P^{1/5}$. Thus, the optimal order of the output convergence bound under a fixed parameter budget is $P^{-1/5}$.

\paragraph{}By comparison, the slower hidden-state convergence bound of \cite[Theorem~2.5]{chaintron2026resnets} has optimal order $P^{-1/6}$, attained, for example, when $L\asymp M\asymp D\asymp P^{1/3}$. The precise architectural scaling can then be chosen according to computational efficiency considerations. In particular, minimising the depth while retaining the optimal output rate, so as to maximise parallelism as in \cite{chaintron2026resnets}, gives $D\asymp M\asymp L^2$. This relation $D\asymp L^2$ is close to the shapes of large models used in practice \cite{grattafiori2024llama3herdmodels}.

\subsection{Our Contributions}
\begin{itemize}
    \item \emph{Output rate.} We prove the upper bound in the conjecture of \cite{chaintron2026resnets} for fixed embedding dimension. With high probability, and uniformly over a fixed number of training steps, the outputs of the ResNet converge to those of the infinitely deep and wide limit at the rate of Theorem~\ref{thm:MainTheorem}. Compared with the hidden states, the fluctuation term improves by a factor $\sqrt D$. The $L^{-1}$ and $(LM)^{-\frac{1}{2}}$ terms are the expected orders of the error from discretising the limiting ODE in depth and of the Monte Carlo error from sampling $LM$ neurons, respectively, while the factor $D$ in the last term cannot be reduced in general, as noted in \cite[Section~3.3.1]{chaintron2026infiniteresnetsgaussianmean}.
    \item \emph{Output metric.} We establish a loss-induced pseudometric to measure model predictions that reflects the task at hand (Section~\ref{sec:BD}).
    \item \emph{Logarithmic factors.} Our bounds contain no logarithmic factor in $L$ or $M$, nor one whose power grows with the number of training steps, provided that the failure probability $\delta$ satisfies \eqref{eqn:shape_main}. This is in contrast with \cite{chaintron2026infiniteresnetsgaussianmean}.
\end{itemize}
\section{ResNets at Infinity}\label{sec:Setup}
All random variables are defined on a probability space $(\Omega,\mathcal F,\mathbb P)$, and for $x,y\in\R^D$, write $\langle x,y\rangle_{\bar2}:=D^{-1}x^\top y$ and $\|x\|_{\bar2}:=\langle x,x\rangle_{\bar2}^{1/2}$ for the RMS inner product and norm. For the logarithmic factors in our high-probability bounds, define
\[
\mathfrak l_\delta:=\log\frac{e(K+1)}{\delta},
\qquad
\mathfrak l^\infty_\delta:=\log\frac{12D(K+1)}{\delta},
\]
where $K$ is the maximal training step. The remaining notation is collected in Appendix~\ref{sec:Notation}.
\subsection{Parameter Initialisation}
We consider sampling parameters at initialisation from some distribution $\pi\in \mathcal P(\R)$
\begin{align}
    \phi^{ilm}_0,\psi^{ilm}_0 &\overset{\mathrm{i.i.d.}}{\sim} \pi, \qquad 1\le i \le D,\, 1\le m \le M,\, 1\le l \le L.
\end{align}
The input and output matrices $W_{\mathrm{in},0}\in\R^{D\times d_{\mathrm{in}}}$ and $W_{\mathrm{out},0}\in\R^{D\times d_{\mathrm{out}}}$ are sampled independently of the residual parameters, and are trained alongside them (Section~\ref{sec:model_setup}).
We require that our parameters are sampled from a distribution with a Stein kernel \cite[Definition~2.7]{NOURDIN20143170}, normalised according to the MLU initialisation.
\begin{assumption}[Stein kernel in the MLU normalisation]
\label{assm:Stein}\label{assm:SteinMLU}\label{assm:MLU}
The coordinates of all initial parameters $\psi^{lm}_0,\phi^{lm}_0$
are i.i.d.\ with law $\pi$, and $\xi\sim\pi$ satisfies:
\begin{enumerate}
\item $\mathbb E[\xi]=0$ and $\|\xi\|_{\psi_2}\le\sigma\sqrt D$;
\item there exists a measurable $\kappa:\mathbb R\to\mathbb R$ with
$\mathbb E[\kappa(\xi)^2]<\infty$ such that
\[
\mathbb E[\xi F(\xi)]=D\,\mathbb E[\kappa(\xi)F'(\xi)]
\]
for every $F\in C^1(\mathbb R)$ with
$\mathbb E|\xi F(\xi)|+\mathbb E|\kappa(\xi)F'(\xi)|<\infty$.
\end{enumerate}
\end{assumption}
Taking $F(x)=x$ gives $\mathbb E[\kappa(\xi)]=\frac1D\mathbb E[\xi^2]=:\mu$, with $\mu\le C\sigma^2$ by the first item, and we write
\[
\mathbb V:=\mathrm{Var}(\kappa(\xi))=\mathbb E\big[(\kappa(\xi)-\mu)^2\big]
\]
for the variance of the Stein kernel. For centred distributions on $\R$ with densities positive on an interval and zero elsewhere, Stein kernels admit an explicit construction \cite{Stein1D}. Assumption \ref{assm:SteinMLU} holds for both the uniform and Gaussian distributions, the latter with $\mathbb V=0$, but fails for Rademacher initialisation.

\paragraph{}This is the Maximal Local Update (MLU) scaling of \cite[Section~4]{chizat2025hiddenwidthdeepresnets}. With the normalised inner product $\langle\cdot,\cdot\rangle_{\bar2}$, it makes the features $\langle\psi^{lm}_0,h\rangle_{\bar2}$ of hidden states with $\|h\|_{\bar2}=\mathcal O(1)$ of order one, and, with the learning rate $\eta$ applied to each neuron, it makes the update of each layer change the hidden states by $\mathcal O(1/L)$, uniformly in $L$, $M$ and $D$.

\subsection{Model Set-Up}\label{sec:model_setup}
Throughout, we consider a single training sample $x$ and suppress its dependence in the notation. The analysis extends to multiple training samples (\cite{chizat2025hiddenwidthdeepresnets,chaintron2026resnets,gibson2026uniform} and Remark \ref{rem:multi_data}), but we restrict to one for notational simplicity. At training step $k\ge 0$, the ResNet with a two-layer perceptron residual is given by
\begin{align*}
    h_{0,k} &= W_{\mathrm{in},k} x, \\
    h_{l,k} & = h_{l-1,k}+ \frac{1}{LM}\sum_{m=1}^M \phi^{lm}_k \rho(\langle \psi_k^{lm}, h_{l-1,k}\rangle_{\bar{2}} ),\qquad 1\le l\le L,\\
    y_k &= \frac{1}{D}W_{\mathrm{out},k}^\top\,h_{L,k},
\end{align*}
where $W_{\mathrm{in},k}\in\R^{D\times d_{\mathrm{in}}}$ and $W_{\mathrm{out},k}\in\R^{D\times d_{\mathrm{out}}}$ are the input and output embeddings, and $x\in\R^{d_{\mathrm{in}}}$.
\begin{assumption}[Initial embeddings]\label{assm:Wout_Init}\label{assm:WoutInit_2}
The initial embeddings $(W_{\mathrm{in},0},W_{\mathrm{out},0})$ are independent of the initial parameters of the residual layers, and may depend on one another. The rows of $W_{\mathrm{in},0}$ are independent, with $\|\langle(W_{\mathrm{in},0})_{i\cdot},x\rangle\|_{\psi_2}\le\sigma_{\mathrm{in}}$ for $1\le i\le D$. The entries of $W_{\mathrm{out},0}$ are independent and centred, with $\|(W_{\mathrm{out},0})_{ia}\|_{\psi_2}\le\sigma_{\mathrm{out}}$. We write $\mathcal F_W:=\sigma(W_{\mathrm{in},0},W_{\mathrm{out},0})$.
\end{assumption}
To control the growth of the hidden states and adjoint variables during training, and to compare them across models, we require the following assumption on the activation function.
\begin{assumption}\label{assm:rho}
Assume that there exists $\Lambda_\rho>0$, such that the activation function $\rho\in C^3(\R)$ has linear growth $|\rho(x)|\le\Lambda_\rho(1+|x|)$ and bounded derivatives up to the third order; $\|\rho^{(i)}\|_\infty\le\Lambda_\rho$.
\end{assumption}
The parameters $(\phi_k^{lm},\psi_k^{lm})_{l,m=1}^{L,M}$ are updated by clipped gradient descent on a loss $\ell(y,\mathfrak t)$, where $y\in\mathbb R^{d_{\mathrm{out}}}$ is the model output and $\mathfrak t\in\mathfrak T$ is an optional target, as in supervised learning. We impose the following assumption on the loss.
\begin{assumption}\label{assm:Obj}
The loss function $\ell :\R^{d_\mathrm{out}}\times \mathfrak{T}\to \R_{\ge 0}$ is convex and twice continuously differentiable in its first component, such that $G_0:=\sup_{\mathfrak t\in\mathfrak T}|\nabla\ell(0,\mathfrak t)|$ is finite. Further assume that the Hessian $\nabla^2_y\ell(y,\mathfrak t)$ with respect to the model output is independent of the target $\mathfrak t$. This last condition is only used so that the pseudometric $d_\ell$ and the Hessian-induced geometry of Section~\ref{sec:BD} do not depend on the target.  Since we are free to rescale the loss function, we take the convention that $\mathrm{tr}(\nabla^2\ell(0,\mathfrak t))=:\tau_\star=\Theta(1)$.
\end{assumption}
The adjoint variables used to update the parameters are given by the backward pass
\begin{align*}
a_{L,k}&=W_{\mathrm{out},k}\nabla\ell(y_k),\\
a_{l-1,k}&=a_{l,k}+\frac1{LM}\sum_{m=1}^M\psi^{lm}_k\,\rho'(\langle\psi^{lm}_k,h_{l-1,k}\rangle_{\bar2})\,\langle\phi^{lm}_k,a_{l,k}\rangle_{\bar2},\qquad 1\le l\le L,
\end{align*}
and the parameters are updated by
\begin{align*}
    \phi^{lm}_{k+1} &= \phi^{lm}_{k}- \eta\, a_{l,k}\,\rho(\langle \psi_k^{lm}, h_{l-1,k}\rangle_{\bar 2}),\\
    \psi^{lm}_{k+1}&= \psi^{lm}_{k}-\eta\, h_{l-1,k}\, \rho'(\langle \psi_k^{lm}, h_{l-1,k}\rangle_{\bar 2})\, \mathrm{cl}(\langle \phi_k^{lm}, a_{l,k}\rangle_{\bar 2}),
\end{align*}
where $\mathrm{cl}$ is the gradient clipping of Assumption~\ref{assm:clipp}. Here, we have implicitly rescaled the gradients with respect to $\phi,\psi$ by $LMD$ to compensate for the scaling factors in the model definition. The embeddings are likewise trained by gradient descent, with gradients rescaled by $D$ and learning rates $\eta_{\mathrm{in}},\,\eta_{\mathrm{out}}>0$ independent of $L,M,D$, from the initial embeddings $W_{\mathrm{in},0}$ and $W_{\mathrm{out},0}$ of Assumption~\ref{assm:Wout_Init}. That is, the embedding matrices are updated according to 
\begin{equation}\label{eqn:embedding_updates}
W_{\mathrm{in},k+1}=W_{\mathrm{in},k}-\eta_{\mathrm{in}}\,a_{0,k}\,x^\top,
\qquad
W_{\mathrm{out},k+1}=W_{\mathrm{out},k}-\eta_{\mathrm{out}}\,h_{L,k}\,\nabla\ell(y_k)^\top .
\end{equation}
\begin{assumption}\label{assm:clipp}
    Suppose that the gradient clipping  function $\mathrm{cl}\in C^2(\R)$ is bounded~$\|\mathrm{cl}\|_\infty\le c$, with bounded derivatives  $\|\mathrm{cl}'\|_\infty\vee\|\mathrm{cl}''\|_\infty\le c$ up to the second order, for some constant $c>0$. 
\end{assumption}

\paragraph{Histories and features.}
We call a sequence $\bm Z=(H_k,A_k)_{k=0}^K\in\R^{(K+1)\times 2D}$ a \emph{history} of hidden states and adjoint variables. Therefore, a neuron with initial parameters $(\phi_0,\psi_0)$, \emph{trained along} a history $\bm Z$, has parameters
\begin{gather*}
\phi_{k+1}(\bm Z)=\phi_k(\bm Z)-\eta\,A_k\,\rho(P_k(\bm Z)),\qquad
\psi_{k+1}(\bm Z)=\psi_k(\bm Z)-\eta\,H_k\,\rho'(P_k(\bm Z))\,\mathrm{cl}(Q_k(\bm Z)).
\end{gather*}
with \[
\phi_0(\bm Z)=\phi_0, \qquad \psi_0(\bm Z)=\psi_0,
\]
and features
\[
P_k(\bm Z):=\langle\psi_k(\bm Z),H_k\rangle_{\bar2},\qquad  Q_k(\bm Z):=\langle\phi_k(\bm Z),A_k\rangle_{\bar2}.
\]
Expanding the updates, the features satisfy the recursion
\begin{align}\label{eqn:p}
    P_k(\bm Z)&= \langle \psi_0,H_k \rangle_{\bar 2}-\eta \sum_{0\le k'< k} \Gamma^{\bm H}_{k',k}\,\rho'(P_{k'}(\bm Z))\, \mathrm{cl}( Q_{k'}(\bm Z)), \\
    Q_k(\bm Z)&= \langle \phi_0,A_k \rangle_{\bar 2}-\eta \sum_{0\le k'< k}\Gamma^{\bm A}_{k',k}\,\rho(P_{k'}(\bm Z)),\label{eqn:q}
\end{align}
where
\begin{equation}\label{eqn:Gamma}
    \Gamma^{\bm H}_{k',k}:= \langle H_{k'}, H_{k}\rangle_{\bar 2},\qquad \Gamma^{\bm A}_{k',k}:= \langle A_{k'}, A_{k}\rangle_{\bar 2}
\end{equation}
are the Gram matrices of the history. In particular, $P_k(\bm Z)$ and $Q_k(\bm Z)$ depend on the initial parameters only through $\langle\psi_0,H_{k'}\rangle_{\bar2}$ and $\langle\phi_0,A_{k'}\rangle_{\bar2}$, $k'\le k$, and on the history only through these inner products and the Gram matrices, which is a key observation used for the proof of Theorem \ref{thm:MainTheorem}. Finally, let
\[
\bm z_l:=(h_{l-1,k},a_{l,k})_{k=0}^K ,
\]
be the history of the $l$-th layer of the discrete model. We use lowercase letters to denote the quantities associated with it:
\[
p^{lm}_k:=P^{lm}_k(\bm z_l)=\langle\psi^{lm}_k,h_{l-1,k}\rangle_{\bar2},\qquad q^{lm}_k:=Q^{lm}_k(\bm z_l)=\langle\phi^{lm}_k,a_{l,k}\rangle_{\bar2},
\]
where $\phi^{lm}_k$ and $\psi^{lm}_k$ abbreviate $\phi^{lm}_k(\bm z_l)$ and $\psi^{lm}_k(\bm z_l)$, respectively.
\subsection{Output Geometry}\label{sec:BD}
This work is motivated by studying scaling limits of transformers from a theoretical perspective. In fact, much of the analysis automatically extends to the transformer (see \cite{gibson2026uniform} and \cite[Section~2.2]{chizat2025hiddenwidthdeepresnets}), and it is only the growth of the derivative of the residual with respect to the parameters that needs to be addressed. The stereotypical uses of the transformer, diffusion models \cite{peebles2023scalable} and LLMs \cite{grattafiori2024llama3herdmodels}, are trained to minimise loss functions related to the mean squared error (MSE) and cross-entropy loss (CE), respectively. This is at least true during pretraining.\\

If our model is trained to minimise an MSE loss, it is natural to measure the discrepancy of the model output using Euclidean or RMS norms. Whereas, models trained with cross-entropy produce logits that determine class probabilities. Furthermore, LLMs have a large vocabulary size, but the output probabilities concentrate on few tokens. Hence, it is unnatural to consider the Euclidean norm on the outputs of such models. Therefore, our measure of discrepancy between model outputs should be task dependent. We use the geometry in which the Hessian of the convex potential $\ell$ is the metric tensor. That is, for a 2-absolutely continuous curve $\gamma:[0,1]\to\R^{d_{\mathrm{out}}}$ (see \cite[Definition~1.1.1]{ambrosio2005gradient}), define the action
\[
\mathcal A_\ell(\gamma):=\int_0^1\dot\gamma(s)^\top\,\nabla^2\ell(\gamma(s),\mathfrak t)\,\dot\gamma(s)\,ds,
\]
We use the following pseudometric to measure discrepancy between model outputs
\[
d_\ell(y,y'):=\underset{\gamma(0)=y,\gamma(1)=y'}{\inf_{\gamma \in AC^2([0,1], \R^{d_{out}})}}\mathcal A_\ell(\gamma)^{1/2},
\]
Under Assumption~\ref{assm:Obj}, the distance $d_\ell$ does not depend on the target. Since $\ell$ is convex, its Hessian may be degenerate, so $d_\ell$ is in general only a pseudometric. It induces a metric on the quotient space obtained by identifying outputs at zero distance, which coincides with the original space when the Hessian is positive definite. For cross-entropy, these equivalence classes are $y+\operatorname{span}\{\mathbf 1\}$, since the Hessian has kernel $\operatorname{span}\{\mathbf 1\}$.

\paragraph{Hessian Induced Geometry.}By Assumption~\ref{assm:Obj}, the Hessian $\mathscr H(y):=\nabla^2\ell(y)$ is independent of the target and is positive semidefinite (PSD). To measure how the Hessian varies with the output, we use the Thompson metric on PSD matrices \cite[Definition~2.1]{lemmens2015unique},
\[
d_T(A,B):=\inf\{r\geq0:e^{-r}A\preceq B\preceq e^rA\},
\]
with $\inf\emptyset:=\infty$, so that $d_T(A,B)$ is finite if and only if $A$ and $B$ have the same kernel. This leads to the loss-induced seminorm
\[
\|u\|_\ell:=\sup_{y\in\R^{d_{\mathrm{out}}}}\sup_{s>0}
\frac{d_T\big(\mathscr H(y+su),\mathscr H(y)\big)}{s},
\qquad u\in\R^{d_{\mathrm{out}}}.
\]
It is symmetric, positively homogeneous and subadditive by the triangle inequality for $d_T$, and satisfies
\[
d_T\big(\mathscr H(y),\mathscr H(y')\big)\leq\|y-y'\|_\ell.
\]
For the MSE loss, $\|\cdot\|_\ell$ vanishes identically, while for cross-entropy it is bounded by the oscillation seminorm (Examples~\ref{ex:MSE} and~\ref{ex:CE}). We shall also use the spectral representation of the Thompson metric. Namely, for nonzero $A,B$ with the same kernel,
\[
d_T(A,B)=\log\max\left\{
\|A^{\dagger/2}BA^{\dagger/2}\|_{\mathrm{op}},
\|B^{\dagger/2}AB^{\dagger/2}\|_{\mathrm{op}}
\right\},
\]
where $\dagger$ denotes the Moore--Penrose pseudoinverse. We take the Hessian at the origin as our reference and write
\[
\mathscr H_\star:=\mathscr H(0),\qquad
\tau_\star:=\mathrm{tr}\,\mathscr H_\star,\qquad
\lambda_\star:=\|\mathscr H_\star\|_{\mathrm{op}}\le\tau_\star.
\]
For $v\in\R^{d_{\mathrm{out}}}$ and $\mathscr H\succeq0$, let $|v|_{\mathscr H}:=(v^\top\mathscr Hv)^{1/2}$. We measure an output embedding $W\in\R^{D\times d_{\mathrm{out}}}$ by
\begin{equation}\label{eqn:mathfrak f&n}
\mathfrak f(W):=\frac{\|W\mathscr H_\star^{1/2}\|_F}{\sqrt D},
\qquad
\mathfrak n(W):=\sup_{\|h\|_{\bar2}\le1}
\Big\|\frac1DW^\top h\Big\|_\ell.
\end{equation}
The first measures the size of the readout in the geometry of the loss, while the second controls the variation of the Hessian along its output directions. Writing $\widehat W:=W/\sqrt D$ for the readout acting on Euclidean-normalised hidden states, we have
\[
\mathfrak f(W)^2
=\|\widehat W\mathscr H_\star^{1/2}\|_F^2
\leq\|\widehat W\|_{\mathrm{HS}}^2
\sup_y\|\mathscr H(y)\|_{\mathrm{op}}.
\]
The assumptions of \cite{chaintron2026infiniteresnetsgaussianmean} bound the right-hand side by controlling the Hilbert--Schmidt norm of the initial output embedding and imposing a global Lipschitz bound on the loss gradient, equivalently a uniform operator-norm bound on its Hessian. By contrast, we bound the left-hand side directly and normalise the trace $\tau_\star$, yielding a weaker requirement. Both sides scale in the same way under a rescaling $\tilde\ell(y)=\ell(\alpha y)$, $\alpha>0$, so that their ratio compares the two assumptions independently of the normalisation of the loss. For a Gaussian output embedding, this ratio is approximately $d_{\mathrm{out}}\sup_y\|\mathscr H(y)\|_{\mathrm{op}}/\tau_\star$. It equals one for the mean squared error, for which the two assumptions coincide, and is of order $d_{\mathrm{out}}$ for cross-entropy. Our assumption is weaker here because we only use the Hessian at the origin rather than a global bound on its operator norm.
\begin{assumption}[Loss regularity]\label{assm:Loss}
The following loss-dependent constants are finite:
\begin{enumerate}
    \item[(L1)] $\displaystyle \varkappa_\ell:=\sup_{y\in\mathbb R^{d_{\mathrm{out}}},\,\mathfrak t\in\mathfrak T}\|\nabla\ell(y,\mathfrak t)\|_\ell$.
    \item[(L2)] $\displaystyle \Lambda_\ell:=\sup_{|v|\le1}\|v\|_\ell$.
\end{enumerate}
They depend only on the loss, and so on the dimensions only through $d_{\mathrm{out}}$. For the MSE and CE losses they are bounded independently of $d_{\mathrm{out}}$ (Examples~\ref{ex:MSE} and~\ref{ex:CE}).
\end{assumption}
A training step of the output embedding moves the outputs along $\nabla\ell$, so (L1) states that it changes the Hessian at a bounded rate. (L2) states that the Hessian changes at a bounded rate along Euclidean unit directions. Together with Assumption~\ref{assm:Wout_Init}, (L2) controls $\mathfrak n(W_{\mathrm{out},0})$ through $\Lambda_\ell$ and the Gaussian width \cite[Definition~7.5.1]{Vershynin_2026}
\[
\mathcal{GW}(\|\cdot\|_\ell):=\mathbb E\|g\|_\ell,\qquad g\sim\mathcal N(0,I_{d_{\mathrm{out}}}).
\]
We write $\mathfrak g_\ell:=\mathcal{GW}(\|\cdot\|_\ell)/\sqrt D$ for the Gaussian width relative to the embedding dimension.

\begin{lemma}[Initial embeddings]\label{lemma:embed_event}
Suppose Assumptions~\ref{assm:Wout_Init}, \ref{assm:Obj} and~\ref{assm:Loss} hold and $\delta\ge\exp(-c_1D)$, and let $\mathfrak t$ be the target of the training sample. There exist $C_{\mathrm{in}}$, $C_{\mathfrak f}$ and $C_r$, depending only on $\sigma_{\mathrm{in}}$, $\sigma_{\mathrm{out}}$, $\tau_\star$, $G_0$ and $c_1$, and $C_{\mathfrak n}$, depending only on $\sigma_{\mathrm{out}}\Lambda_\ell$, $\sigma_{\mathrm{out}}\mathfrak g_\ell$ and $c_1$, and an $\mathcal F_W$-measurable event $\mathcal E^W_\delta$ with $\mathbb P(\mathcal E^W_\delta)\ge1-\delta$, on which
\[
\|W_{\mathrm{in},0}\,x\|_{\bar2}\le C_{\mathrm{in}},\qquad
\mathfrak f(W_{\mathrm{out},0})\le C_{\mathfrak f},\qquad
\|W_{\mathrm{out},0}\nabla\ell(0,\mathfrak t)\|_{\bar2}\le C_r,\qquad
\mathfrak n(W_{\mathrm{out},0})\le C_{\mathfrak n}.
\]
\end{lemma}
The proof is given in Appendix~\ref{app:output_geometry}.\\
\begin{remark}[Dependence on the loss and the dimensions]\label{rem:gw}
The loss enters the constants of this paper only through $\tau_\star$, $\lambda_\star\le\tau_\star$, $\Lambda_\ell$, $\varkappa_\ell$, $G_0$ and the Gaussian width $\mathcal{GW}(\|\cdot\|_\ell)$. These may depend on the output dimension $d_{\mathrm{out}}$, on which the loss is defined, while the embedding dimension $D$ enters only through $\mathfrak g_\ell=\mathcal{GW}(\|\cdot\|_\ell)/\sqrt D$. In the bounds of Lemmas~\ref{lemma:embed_event}, \ref{lemma:apriori} and~\ref{lemma:boundary}, from which all later constants are built, each of these quantities appears multiplied by the initialisation scale or the learning rate of the output embedding, so that the constants depend on them only through upper bounds for
\[
\mathfrak{s}=\sigma_{\mathrm{out}}\big(\sqrt{\tau_\star}+\Lambda_\ell+\mathfrak g_\ell+G_0\big), \qquad
\varsigma=\eta_{\mathrm{out}}\big(\lambda_\star+\varkappa_\ell+G_0^2\big),
\]
and are non-decreasing in them. They also depend on the lower bound for $\tau_\star$ implicit in the normalisation $\tau_\star=\Theta(1)$ of Assumption~\ref{assm:Obj}. Choosing $\sigma_{\mathrm{out}}$ and $\eta_{\mathrm{out}}$ so that both quantities are of order one makes all constants uniform in $D$, $d_{\mathrm{out}}$ and the loss. For the CE loss, $\hbox{$\varkappa_\ell,\Lambda_\ell\le2$}$, $G_0\le 1$, $\lambda_\star=1/d_{\mathrm{out}}$ and $\mathfrak g_\ell\le2\sqrt{2\log(2d_{\mathrm{out}})/D}$, so that $\mathfrak s\le\sigma_{\mathrm{out}}(4+\mathfrak g_\ell)$ and $\hbox{$\varsigma \le 4\eta_{\mathrm{out}}$}$, and readout entries and learning rate of order one suffice whenever $\log d_{\mathrm{out}}\lesssim D$. For the mean squared error, $\Lambda_\ell=\mathfrak g_\ell=\varkappa_\ell=0$, $\lambda_\star=1/d_{\mathrm{out}}$ and $G_0=\sup_t|t|/d_{\mathrm{out}}$, which is $O(d_{\mathrm{out}}^{-1/2})$ for targets with uniformly bounded entries. As a result, $\mathfrak s\lesssim\sigma_{\mathrm{out}}$ and $\varsigma\lesssim\eta_{\mathrm{out}}/d_{\mathrm{out}}$, allowing $\sigma_{\mathrm{out}}=O(1)$ and $\eta_{\mathrm{out}}\asymp d_{\mathrm{out}}$. These dimension dependencies are discussed in Appendix~\ref{app:dout}.
\end{remark}

\paragraph{} The mean squared error and the cross-entropy loss are studied in Appendix~\ref{app:dout_examples} (Examples~\ref{ex:MSE} and~\ref{ex:CE}).

\subsection{Limiting Models}\label{sec:LimitingModel}
We write $(H_{t,k},A_{t,k})$, $t\in[0,1]$, for the hidden states and adjoint variables of the infinitely wide and deep model. Conditionally on $\mathcal F_W$ they are deterministic, and, with a single training sample, satisfy the forward-backward system of ODEs
\begin{align*}
H_{t,k}&=W^\infty_{\mathrm{in},k}x+\int_0^t\mathbb E\big[\phi^\infty_{s,k}\,\rho(P_{s,k})|\mathcal F_W\big]ds,
\\
A_{t,k}&=W^\infty_{\mathrm{out},k}\nabla\ell(Y_k)+\int_t^1\mathbb E\big[\psi^\infty_{s,k}\,\rho'(P_{s,k})\,Q_{s,k}\big|\mathcal F_W\big]ds,\\
Y_k &= \frac 1D (W^\infty_{\mathrm{out},k})^\top H_{1,k},
\end{align*}
with embeddings trained as in \eqref{eqn:embedding_updates}, along the limiting model,
\[
W^\infty_{\mathrm{in},k+1}=W^\infty_{\mathrm{in},k}-\eta_{\mathrm{in}}\,A_{0,k}\,x^\top,
\qquad
W^\infty_{\mathrm{out},k+1}=W^\infty_{\mathrm{out},k}-\eta_{\mathrm{out}}\,H_{1,k}\,\nabla\ell(Y_k)^\top,
\]
from $W^\infty_{\mathrm{in},0}=W_{\mathrm{in},0}$ and $W^\infty_{\mathrm{out},0}=W_{\mathrm{out},0}$. Suppose $\psi_0,\phi_0$ are random variables on $(\Omega,\mathcal F,\mathbb P)$, independent of each other and of $\mathcal F_W$, each with law $\pi^{\otimes D}$ (Assumption \ref{assm:MLU}). Then, the parameters $\phi^\infty_{\cdot,k},\psi^\infty_{\cdot,k}\in C([0,1],L^2)$ are recursively defined by
\[
\psi^\infty_{t,k+1}-\psi^\infty_{t,k} = -\eta H_{t,k}\rho'(\langle \psi^\infty_{t,k},H_{t,k}\rangle_{\bar 2})\mathrm{cl}(\langle \phi^\infty_{t,k},A_{t,k}\rangle_{\bar 2}), \qquad \psi^\infty_{t,0}=\psi_0.
\]
and $\phi^\infty_{t,k}$ are defined similarly. A generic neuron at layer time $s$ is trained along the history of the limiting model at that depth,
\[
\bm Z_s:=(H_{s,k},A_{s,k})_{k=0}^K,
\]
and its parameters and features are written with the depth as a subscript,
\[
\phi^\infty_{s,k}:=\phi_k(\bm Z_s),\qquad\psi^\infty_{s,k}:=\psi_k(\bm Z_s),\qquad P_{s,k}:=P_k(\bm Z_s),\qquad Q_{s,k}:=Q_k(\bm Z_s),
\]
in the notation of Section~\ref{sec:Setup}. Layer $r$ of the discrete model is compared with the limiting hidden state at depth
$(r-1)/L$ and the adjoint at depth $r/L$, that is, with the deterministic history
\[
\bm Z^\infty_r:=(H_{\frac{r-1}L,k},A_{\frac rL,k})_{k=0}^K .
\]
A hat denotes the parameters and features of a
layer trained along $\bm Z^\infty_r$, so that
\[\hat\Phi^r_k:=\Phi^r_k(\bm Z^\infty_r), \qquad  \hat P^r_k:=P^r_k(\bm Z^\infty_r),\] and similarly for $\Psi$ and $Q$.

\subsection{Statement of the Main Theorems}
Throughout, fix $c_0,c_1>0$, and let $N_0,\zeta_*>0$ be the constants constructed in Section~\ref{sec:LOLO_induction}, which depend only on $K$, $c_0$, $c_1$ and the constants in our assumptions. We impose the shape and failure-probability constraints
\begin{equation}\label{eqn:shape_main}
L\ge N_0,\qquad D\le c_0M,\qquad
\frac{D}{LM^{2/3}}\le\zeta_*,
\qquad
\delta\ge\exp\big(-c_1(D\wedge M^{1/3})\big).
\end{equation}
\begin{theorem}\label{thm:bar 2-norm}
Suppose Assumptions~\ref{assm:MLU}--\ref{assm:Loss} and the shape and failure probability constraints \eqref{eqn:shape_main} hold. Then there exists $C_1$, independent of $L,M,D,d_{\mathrm{in}}$, depending only on $K$, $\eta$, $\eta_{\mathrm{in}}|x|^2$, $\sigma$, $\sigma_{\mathrm{in}}$, $\Lambda_\rho$, $c$, $c_0$, $c_1$, and on the loss and $d_{\mathrm{out}}$ only through upper bounds on $\mathfrak s$ and $\varsigma$ (Remark~\ref{rem:gw}), such that, with probability at least $1-\delta$,
    \[
    \max_{0\le k \le K}\max_{0\le l \le L}\|H_{\frac{l}{L},k}-h_{l,k}\|_{\bar 2} \vee  \|A_{\frac{l}{L},k}-a_{l,k}\|_{\bar 2} \le C_1 \left( \frac{1}{L}+\sqrt{\frac{D+\mathfrak l_\delta}{LM}}\right).
    \]
\end{theorem}
\begin{theorem}[Main Theorem]\label{thm:MainTheorem}\label{thm:Rates}
Suppose that Assumptions~\ref{assm:MLU}--\ref{assm:Loss} and the shape and failure probability constraints \eqref{eqn:shape_main} hold.
Then, with probability at least $1-\delta$,
\begin{align*}
\max_{k\le K}d_\ell(y_k,Y_k)
&\le C_2\,
\left[\frac1L+\sqrt{\frac{\mathfrak l_\delta}{LM}}
+\frac{D}{LM}
\right]+C_2\,\sqrt{\mathbb V}\left[\sqrt{\frac{\mathfrak l^\infty_\delta}{LM}}
+\frac1L\sqrt{\frac{\mathfrak l^\infty_\delta}{D}}
\right],
\end{align*}
where $C_2$ depends only on the quantities listed in Theorem~\ref{thm:bar 2-norm} and on $\mu$ and $\mathbb V$. If $\mathbb V\ne0$, which is only possible for non-Gaussian initialisation, the bound picks up an extra $\sqrt{\log D}$ term, since the Stein-kernel terms involve $\mathfrak l^\infty_\delta\le\mathfrak l_\delta+\log(12D)$ instead of $\mathfrak l_\delta$.
\end{theorem}
The proof gives the bound in terms of the concentration rates $\mathcal T$ of Appendix~\ref{sec:Notation}, before the shape constraints are used, in \eqref{eqn:Lbar_collected}, to simplify them.
\begin{remark}[The limit $D\to\infty$, Informal]\label{rem:joint_D}
Suppose that $\Lambda:=\sup_{z\in\mathbb R^{d_{\mathrm{out}}}}\|\nabla^2\ell(z)\|_{\mathrm{op}}$ is finite. Then
\[
d_\ell(y,y')\le\sqrt{\Lambda}\,|y-y'|,
\qquad y,y'\in\mathbb R^{d_{\mathrm{out}}}.
\]
Let $Y_k^{\mathrm{DMFT}}\in\mathbb R^{d_{\mathrm{out}}}$ denote the deterministic output at training step $k$ of the dynamical mean-field limit constructed in \cite[Section~2.2]{chaintron2026resnets}. Under the additional assumptions needed for this limit, adapting \cite[Theorem~2.5]{chaintron2026resnets} to our training dynamics gives
\[
\max_{k\le K}|Y_k- Y^{\mathrm{DMFT}}_k|
\le C\frac{1+\log(2/\delta)}{\sqrt D}
\]
with probability at least $1-\delta/2$, for $\delta>2e^{-D}$. Combining this with Theorem~\ref{thm:Rates}, applied at confidence level $\delta/2$, and using the shape constraints to simplify the rates, gives, with probability at least $1-\delta$,
\[
\max_{k\le K}d_\ell(y_k,Y^{\mathrm{DMFT}}_k)
\le C\left[
\frac1L+\frac D{LM}+\frac{\mathfrak l_\delta}{\sqrt D}
\right]
+C\sqrt{\mathbb V}\sqrt{\frac{\mathfrak l^\infty_\delta}{D}}.
\]
This would yield the upper bound in the conjecture of \cite{chaintron2026resnets}, with the Euclidean norm replaced by the loss-induced pseudometric $d_\ell$.
\end{remark}
\begin{remark}[Several training samples]\label{rem:multi_data}
For simplicity, we consider a single training sample $(x,\mathfrak t)$. As in \cite[Remark~4]{chaintron2026resnets}, the analysis extends to full-batch gradient descent on a finite training set $\{(x_i,\mathfrak t_i)\}_{i=1}^{N_{\mathrm{data}}}$, at the cost of heavier notation. The histories and Gram matrices then carry an additional sample index, and the features $P,Q$ involve averages over the samples at earlier training steps. The proof follows the same structure, with the constants depending on $KN_{\mathrm{data}}$ rather than $K$ alone. Union bounds are likewise taken over training step--sample pairs, adding $\log N_{\mathrm{data}}$ to $\mathfrak l_\delta$ and $\mathfrak l^\infty_\delta$.
\end{remark}

\subsection{Sketch Proof}\label{sec:Sketch}

\paragraph{Error decomposition and decoupling.} Fix a direction $w$ independent of the residual parameters, and write
\[
e^H_{l,k}(w):=\langle w,h_{l,k}-H_{l/L,k}\rangle_{\bar 2}.
\]
Inserting at each layer the residual that would have been obtained had its neurons been trained along the limiting history gives the decomposition
\[
e^H_{l,k}(w)
=e^H_{0,k}(w)
+\langle w,\mathcal E^H_{\mathrm{T.D.},l,k}\rangle_{\bar 2}
+\langle w,\mathcal E^H_{\mathrm{m.c.},l,k}\rangle_{\bar 2}
+\langle w,\mathcal E^H_{\mathrm{prop},l,k}\rangle_{\bar 2},
\]
and an analogous decomposition for the adjoint variables. The four terms, explicitly defined in \eqref{eqn:error_decomp}, are described as the input condition, time discretisation, Monte Carlo fluctuations and propagation of errors through training. The time discretisation contributes $O(L^{-1})$ \cite[Lemma~5.4]{chizat2025hiddenwidthdeepresnets}, while the Monte Carlo error, being an average over $LM$ independent neurons, is of order $(LM)^{-1/2}$ in a fixed direction, up to logarithmic factors (Lemma \ref{lemma:MC}). Since the history along which a layer is trained depends on that layer's own parameters, we replace the history with that of the network in which the layer has been removed. We call this the leave-one-layer-out (LOLO) history. The \emph{decoupling error} is the difference between the residuals of the layers trained along their own histories and along the LOLO histories. It is controlled by the distance between the two histories (Proposition~\ref{prop:comparison} and Corollary~\ref{cor:MainLemma}). The difference between the LOLO and discrete model histories, on average over the layers, is of order $L^{-1}+D/(LM)$ (Proposition~\ref{lemma:MainLemma}). As the residual with the LOLO history is an average of $M$  i.i.d. neurons, we split the remaining propagation error into a conditional mean and a centred fluctuation.

\paragraph{Residual Fluctuation.} Conditional on the leave-one-layer-out histories, the fluctuations are averages of independent, centred neuron contributions. Each is a product of at most three sub-Gaussian variables, with scale proportional to the difference between the histories, and hence has $2/3$-sub-Weibull tails. Thus, we pick up an additional $M^{-\frac{1}{2}}$ from the law of large numbers in the width, yielding errors of order $\mathcal O((LM)^{-1/2})$ under our shape constraints.

\paragraph{Bias.} The residuals depend on the histories through their Gram matrices and their inner products with the initial parameters. Differences of Gram matrices decompose into projections of the errors onto the limiting histories, together with terms quadratic in the errors. Applying the Stein identity to the initial parameters gives the same structure for the remaining terms (Lemma~\ref{lemma:Lip_in_E}). For Gaussian initialisation, the Stein kernel is constant,  and so no additional term appears. Thus, apart from quadratic errors, the propagation error is controlled by projections onto the limiting hidden states and adjoint variables, together with projections onto the direction in which the error is being measured.

\paragraph{Non-Gaussian Initialisation.} The bias contains an additional term, equal to $\sqrt{\mathbb V/D}$ times the $\bar 2$-norm of the hidden-state error and the $\bar\infty$-norm of the projection direction $w$. For the family of directions used below, $\|w\|_{\bar\infty}=\mathcal O(\sqrt{\log D})$ (Lemma~\ref{lemma:InfNorm}).

\paragraph{Closing the projections.}
It remains to find a family of directions on which these estimates close. The bias at layer $r$ involves the projections of the errors onto the direction $w$ itself at the earlier steps, and onto the limiting hidden states and adjoint variables at the same depth, $\mathcal W_r:=\{H_{\frac rL,k},A_{\frac rL,k}:k\le K\}$ \eqref{eqn:depth_directions}. The outputs require one further family of directions. Let $\omega_{j,k}$ be the normalised image, under the limiting output embedding, of the $j$-th eigenvector of $\nabla^2\ell(Y_k)$ \eqref{eqn:output_directions}. The output error $y_k-Y_k$ decomposes into the last-layer hidden-state error, projected through $W^\infty_{\mathrm{out},k}$, and the contribution $\Delta^W_k$ from the output embedding. The Hessian-weighted norm of the former is precisely the eigenvalue-weighted root mean square, denoted $\mathcal Q_k$, of the last-layer errors projected onto the directions $\omega_{j,k}$ \eqref{eqn:Hessian_projections}. Lemma~\ref{lemma:grad_metric}, with $\bar y=Y_k$, then bounds both $d_\ell(y_k,Y_k)$ and the difference of the loss gradients by a multiple of $\mathcal Q_k+|\Delta^W_k|_{\mathscr H_\star}$ \eqref{eqn:Dk_split}, \eqref{eqn:Bregman_projections}, where $|\cdot|_{\mathscr{H}_\star}$ denotes the Mahalanobis seminorm induced by $\mathscr{H}_\star$. The input embedding contributes only earlier first-layer adjoint errors projected onto the same direction $w$ \eqref{eqn:input_projection}. The output embedding is updated by rank-one terms $h_{L,k'}\nabla\ell(y_{k'})^\top$ for $k'<k$ \eqref{eqn:DeltaWout}, so its contribution $\Delta^W_k$ to the outputs involves only earlier last-layer errors projected onto $H_{1,k}\in\mathcal W_L$ and earlier gradient differences \eqref{eqn:DeltaW_bound}. Its contribution to the terminal adjoint error in direction $w$ likewise involves only earlier last-layer errors projected onto $w$. The remaining, loss-dependent, terminal adjoint error is controlled by $\mathcal Q_k+|\Delta^W_k|_{\mathscr H_\star}$ \eqref{eqn:terminal_projections}. Thus the directions $\mathcal W_r$, $0\le r\le L$, together with $\omega_{j,k}$, all $\mathcal F_W$-measurable, form a family closed under the error dynamics.

\paragraph{Induction.}
For each of these directions $w$, let $\mathscr L^\star_k(w)$ be the sum of the time discretisation, Monte Carlo, decoupling and fluctuation errors in the direction $w$, maximised over the layers and the training steps, together with the Stein-kernel term of the bias at step $k$ and its contributions from earlier steps, weighted by $\bar\eta$ (see \eqref{s4:L_def}). Here, $\bar \eta$ is the effective learning rate of \eqref{eqn:eta_bar}. Let $\bar{\mathscr L}^\star$ collect the maximum of $\mathscr L^\star_k(w)$ over $\mathcal W_r$ and $k$, averaged over the layers, the last layer $\mathcal W_L$, and an eigenvalue-weighted root mean square over the output directions, together with the layer-wise average $\bar\varrho$ of the quadratic terms \eqref{s4:Lbar_def}. The quantity $\bar{\mathscr L}^\star$ does not depend on the training step. We prove by induction over the training steps that, at step $k$, the error in each such direction $w$ is at most $\mathsf C^{k+1}(\mathscr L^\star_k(w)+\bar{\mathscr L}^\star)$, and that $\mathcal Q_k+|\Delta^W_k|_{\mathscr H_\star}\le \mathsf C^{k+1}\bar{\mathscr L}^\star$ \eqref{s4:claim}. Errors at earlier steps enter with a factor of the effective learning rate, and are bounded by the induction hypothesis. Within the current step, the estimates are closed in three stages. The hidden-state error at layer $l$ depends on those at the layers below it, and so is bounded by a discrete Gr\"onwall inequality upwards in depth. This bound, at the last layer and in the output directions, controls $\mathcal Q_k$, and hence the terminal adjoint error. The adjoint error at layer $l$ depends on those at the layers above it, starting from the terminal condition, and so is bounded by a discrete Gr\"onwall inequality downwards in depth. In both recursions, each layer $r$ contributes with weight $1/L$, through the errors in its own directions $\mathcal W_r$. Therefore, these errors enter only through their average over the layers. We never need them to be small at every layer simultaneously, which would require a union bound over the $L$ layers and cost a factor $\log L$. It suffices that their average is small, which is why $\bar{\mathscr L}^\star$ averages over the layers. Each stage costs a constant factor, independent of $k$, and so $d_\ell(y_k,Y_k)\le C\,\mathsf C^{K+1}\bar{\mathscr L}^\star$ \eqref{eqn:S_by_Lbar}.

\paragraph{Rates.} In a fixed direction, the Monte Carlo and fluctuation errors are of order $(LM)^{-1/2}$, up to the logarithmic factor $\sqrt{\mathfrak l_\delta}$. Since they enter $\bar{\mathscr L}^\star$ only through averages over the depths and the output directions, Lemma~\ref{lemma:LogRemoval_mixed} combines them without a union bound, and so without a $\log L$ or $\log d_{\mathrm{out}}$ factor. The decoupling and quadratic errors contribute $D/(LM)$, and the time discretisation contributes $1/L$. Together with the Stein-kernel terms for non-Gaussian initialisation, this gives Theorem~\ref{thm:Rates}.

\paragraph{Proof Layout.} The necessary estimates are established in the following sections. Section~\ref{sec:PROP} develops the linear response of neuron features to changes in their training histories, the leave-one-layer-out comparison, and the bias and fluctuation bounds. Section~\ref{sec:MC} establishes the fixed-direction Monte Carlo estimates. Section~\ref{sec:ProofMain} combines these ingredients and closes the projection estimates by induction over the training steps. Auxiliary probabilistic and a priori estimates are collected in the appendices.

\section{How Differences Propagate over Training}\label{sec:PROP}
\subsection{Linear Response Matrix}\label{sec:LinearResponse}
Throughout this subsection, we fix a layer $r$ and a neuron $m$, and drop their indices. Let
$\bm Z^0=(\bm H^0,\bm A^0)$ and $\bm Z^1=(\bm H^1,\bm A^1)$ be two histories in $\R^{(K+1)\times 2D}$, in the sense of Section~\ref{sec:Setup}, and let
\[
\bm Z^s:=\bm Z^0+s(\bm Z^1-\bm Z^0)=(\bm H^s,\bm A^s),\qquad s\in[0,1],
\]
be their linear interpolation. We study the features of the neuron trained along the interpolated history,
\[
P_{k,s}:=P_k(\bm Z^s),\qquad Q_{k,s}:=Q_k(\bm Z^s),\qquad 0\le k\le K,
\]
defined by the recursion \eqref{eqn:p}--\eqref{eqn:q} with the history $\bm Z^s$ and the fixed initial parameters $(\phi_0,\psi_0)$, and we write $\Gamma^{\bm H^s}_{k',k}$ and $\Gamma^{\bm A^s}_{k',k}$ for the Gram matrices \eqref{eqn:Gamma} of $\bm Z^s$. 
After differentiating with respect to $s$, we obtain the recursion
\begin{equation}\label{eqn:LinearRecursion}
\begin{pmatrix}
    \partial_s P_{k,s} \\ \partial_s Q_{k,s}
\end{pmatrix} = \mathcal B_k(\bm Z^s)+\sum_{0\le k' <k} \mathcal M_{k',k}(\bm Z^s)
\begin{pmatrix}
    \partial_s P_{k',s} \\ \partial_s Q_{k',s}
\end{pmatrix},\qquad 0\le k\le K,
\end{equation}
where $\mathcal B_k(\bm Z^s)\in \R^2$ collects the derivatives that do not fall on the features,
\begin{align*}
    \mathcal B_k(\bm Z^s)&:= \begin{pmatrix} \langle \psi_0,H^1_k-H^0_k\rangle_{\bar 2} \\
        \langle \phi_0,A^1_k-A^0_k\rangle_{\bar 2}
    \end{pmatrix} -\eta \sum_{0\le k'< k}\begin{pmatrix}
       \rho'(P_{k',s})\,\mathrm{cl}(Q_{k',s}) \\ \rho(P_{k',s})
    \end{pmatrix}\odot  \begin{pmatrix}
        \partial_s \Gamma^{\bm H^s}_{k',k}\\ \partial_s\Gamma^{\bm A^s}_{k',k}
    \end{pmatrix},
\end{align*}
with
\[
\partial_s\Gamma^{\bm H^s}_{k',k}=\langle H^1_{k'}-H^0_{k'},H^s_k\rangle_{\bar2}+\langle H^s_{k'},H^1_k-H^0_k\rangle_{\bar2},
\]
and similarly for $\bm A$, and $\mathcal M_{k',k}(\bm Z^s)\in \R^{2\times 2}$ is
\begin{align}\label{eqn:mathcal M}
\mathcal M_{k',k}(\bm Z^s)&:=-\eta  \begin{pmatrix}
     \Gamma^{\bm H^{s}}_{k',k} \;\rho''(P_{k',s})\, \mathrm{cl}(Q_{k',s}) & \Gamma^{\bm H^s}_{k',k}\; \rho'(P_{k',s})\,\mathrm{cl}'(Q_{k',s})\\ \Gamma^{\bm A^s}_{k',k}\; \rho' (P_{k',s}) & 0
\end{pmatrix}.
\end{align}
The recursion \eqref{eqn:LinearRecursion} is a discrete Volterra system, whose solution is
\begin{equation}\label{eqn:VolterraResponse}
    \begin{pmatrix}
    \partial_s P_{k,s} \\ \partial_s Q_{k,s}
\end{pmatrix} = \sum_{0\le k' \le k} \mathcal R_{k',k}(\bm Z^s)\; \mathcal B_{k'}(\bm Z^s),
\end{equation}
where the \emph{linear response matrix} $\mathcal R_{k',k}\in \R^{2\times 2}$ is given by $\mathcal R_{k,k}:=I_2$ and, for $k'<k$,
\begin{equation}\label{eqn:mathcal R}
    \mathcal R_{k',k}:= \sum_{j=1}^{k-k'} \sum_{k' = t_0< t_1<\ldots < t_j=k} \mathcal M_{t_{j-1},t_j}\cdots \mathcal M_{t_0,t_1}.
\end{equation}
Here, the indication of the dependence of $\mathcal R$ and $\mathcal M$ on $\bm{Z}^s$ has been dropped for notational ease. The response matrix $\mathcal R_{k',k}$ propagates a perturbation of the history at step $k'$ to the features at step $k$, while $\mathcal B$ contains the perturbations themselves.

\paragraph{}There is nothing special about the interpolation parameter $s$ in this argument. Therefore, the same computation applies to the derivative with respect to any parameter on which the recursion \eqref{eqn:p}--\eqref{eqn:q} depends smoothly. In particular, for a fixed history $\bm Z$ independent of $\phi_0,\psi_0$, the derivatives of the features with respect to the coordinates $\psi^i_0$ and $\phi^i_0$, $1\le i\le D$, of the initial parameters only receive a contribution from the first term of \eqref{eqn:p}, respectively \eqref{eqn:q}, so that
\begin{align}\label{eqn:PQ_Deriv_theta}
    \begin{pmatrix}
        \partial_{\theta^i}P_k(\bm Z) \\ \partial_{\theta^i}Q_k(\bm Z)
    \end{pmatrix} = \sum_{0\le k'\le k}\mathcal R_{k',k}(\bm Z) \begin{pmatrix}
        \mathbbm{1}_{\{\theta = \psi\}}\,H^i_{k'}/D\\ \mathbbm{1}_{\{\theta = \phi\}}\,A^i_{k'}/D
    \end{pmatrix} ,\qquad \theta\in\{\psi,\phi\}.
\end{align}
\subsection{Preliminary Growth Estimates}
\paragraph{The Output Space.} Let $C_B$, $C'_{\mathfrak f}$ and $C'_{\mathfrak n}$ be the constants of Lemma~\ref{lemma:apriori}, which bound the histories, $\mathfrak f(W_{\mathrm{out},k})$ and $\mathfrak n(W_{\mathrm{out},k})$ of every model at every step. Set $R_*:=C'_{\mathfrak n}C_B$, $S_*:=C'_{\mathfrak f}C_B$, and
\[
\mathcal Y:=\big\{y\in\R^{d_{\mathrm{out}}}:\ \|y\|_\ell\le R_*,\ |y|_{\mathscr H_\star}\le S_*\big\}.
\]
It is convex and contains the origin, and, by Lemma~\ref{lemma:apriori}, on its event, it contains the outputs of the limiting model, of the discrete model and of the leave-one-layer-out networks of Section \ref{sec:LOLO} at every step $k\le K$, and the segments between them. As a result, for $y,y'\in\mathcal Y$,  $d_T(\mathscr H(y),\mathscr H(y'))\le2R_*$, and so
\begin{align}\label{eqn:Hessian_comparable}
e^{-R_*}\mathscr H_\star\preceq\mathscr H(y)\preceq e^{R_*}\mathscr H_\star,\qquad
C_{\mathscr H}^{-1}\mathscr H(y)\preceq\mathscr H(y')\preceq C_{\mathscr H}\mathscr H(y),\qquad C_{\mathscr H}:=e^{2R_*}.
\end{align}
By (L2), $d_T(\mathscr H(y),\mathscr H_\star)<\infty$ for every $y$, so all Hessians have the same kernel as $\mathscr H_\star$. Consequently, the fundamental theorem of calculus gives, for all $y,y'\in\R^{d_{\mathrm{out}}}$,
\begin{align}
\nabla\ell(y,\mathfrak t)-\nabla\ell(y',\mathfrak t)
&=\mathscr H_\star^{1/2}\widetilde M_{y,y'}\mathscr H_\star^{1/2}(y-y'),\label{eqn:grad_factor}\\
\widetilde M_{y,y'}
&:=\mathscr H_\star^{\dagger/2}
\left(\int_0^1\mathscr H(y'+s(y-y'))\,ds\right)
\mathscr H_\star^{\dagger/2}.\notag
\end{align}
For $y,y'\in\mathcal Y$, the first bound of \eqref{eqn:Hessian_comparable} gives $\|\widetilde M_{y,y'}\|_{\mathrm{op}}\le e^{R_*}$. In particular, taking $y'=0$ yields
\[
G_\ell:=\sup_{y\in\mathcal Y,\,\mathfrak t\in\mathfrak T}|\nabla\ell(y,\mathfrak t)|
\le G_0+\sqrt{\lambda_\star}\,e^{R_*}S_*.
\]
\begin{lemma}[Gradients in the output geometry]\label{lemma:grad_metric}
Suppose Assumptions~\ref{assm:Obj} and~\ref{assm:Loss} hold. Let $y,y',\bar y\in\mathcal Y$ and $w\in\R^{d_{\mathrm{out}}}$. Then
\[
d_\ell(y,y')\le\sqrt{C_{\mathscr H}}\,|y-y'|_{\mathscr H(\bar y)},
\qquad
w^\top\big(\nabla\ell(y)-\nabla\ell(y')\big)\le C_{\mathscr H}\,|w|_{\mathscr H(\bar y)}\,|y-y'|_{\mathscr H(\bar y)} .
\]
\end{lemma}
The proof is given in Appendix~\ref{app:output_geometry}.

\begin{lemma}[A priori bound]\label{lemma:apriori}
Suppose Assumptions~\ref{assm:MLU}--\ref{assm:Loss} hold. Assume further that the shape constraints and failure probabilities satisfy the second and fourth constraints of \eqref{eqn:shape_main},
\begin{equation}\label{eqn:S&FPconstraints}
    D\le c_0M, \qquad \delta\ge\exp(-c_1(D\wedge M^{1/3})),
\end{equation}
Then, there exist constants $C_B$, $C'_{\mathfrak f}\ge C_{\mathfrak f}$ and $C'_{\mathfrak n}\ge C_{\mathfrak n}$, independent of $L$, $M$ and $\delta$, and depending on the loss and $d_{\mathrm{out}}$ only through upper bounds on $\mathfrak s$ and $\varsigma$ of Remark~\ref{rem:gw}, and an event $\mathcal E^{\mathrm{op}}_\delta\subseteq\mathcal E^W_{\delta/2}$ with $\mathbb P(\mathcal E^{\mathrm{op}}_\delta)\ge1-\delta$, which only depends on the initial parameters and embeddings, such that, on $\mathcal E^{\mathrm{op}}_\delta$ and every $k\le K$,
\begin{gather*}
\max_{0\le l\le L}\|h_{l,k}\|_{\bar2}\vee\|a_{l,k}\|_{\bar2}\le C_B,
\qquad
\|W_{\mathrm{in},k}\,x\|_{\bar2}\le C_B,\\
\mathfrak f(W_{\mathrm{out},k})\le C'_{\mathfrak f},\qquad\mathfrak n(W_{\mathrm{out},k})\le C'_{\mathfrak n} .
\end{gather*}
The same bounds hold simultaneously on $\mathcal E^{\mathrm{op}}_\delta$ with $(h,a,W_{\mathrm{in}},W_{\mathrm{out}})$ replaced by their counterparts in every leave-one-layer-out network of Definition~\ref{def:LOLO}, for all removed layers\\ $l^*\in\{1,\ldots,L\}$.
\end{lemma}
\begin{proof}
We consider the discrete model, as the argument applies equally to the leave-one-layer-out networks. A leave-one-layer-out network is the same model with the residual of layer $l^*$ deleted, trained by the same updates along its own hidden states and adjoint variables. Every bound below therefore applies to it with the superscript $(-l^*)$, with $o_{l^*,j}:=0$ for the removed layer, and with the same initial parameters and embeddings. Define
\begin{align*}
o_{r,j}&:=\frac{\|\Phi^r_j\|_{\mathrm{op}}\vee\|\Psi^r_j\|_{\mathrm{op}}}{\sqrt{DM}},
\qquad
\bar o_j^2:=\frac1L\sum_{r}o_{r,j}^2,
\\
s_j&:=\mathfrak f(W_{\mathrm{out},j}),\qquad n_j:=\mathfrak n(W_{\mathrm{out},j}),\qquad r_j:=\|W_{\mathrm{out},j}\,g_0\|_{\bar2},
\end{align*}
with $g_0:=\nabla\ell(0,\mathfrak t)$ for the target $\mathfrak t$ of the training sample.

\emph{Stage 1: Growth of the Parameters:} Suppose for induction that \[\max_{0\le l\le L}\|h_{l,j}\|_{\bar2}\vee\|a_{l,j}\|_{\bar2}\le B,\] for some $B\ge0$, for all steps $j<k$. The differences in parameters from one step to the next satisfy
\begin{align*}
    \left\|\Phi^r_{j+1} - \Phi^r_{j}\right\|_{\mathrm{op}}&\le \eta\sqrt{DM} \| a_{r,j}\|_{\bar 2} \| \bm{\rho}(p^r_j)\|_{\bar 2}\le  \eta\sqrt{DM} B\Lambda_\rho(1+\|p^r_j\|_{\bar 2}), \\
     \left\|\Psi^r_{j+1} - \Psi^r_{j}\right\|_{\mathrm{op}} &\le \eta\sqrt{DM}\Lambda_\rho \| \mathrm{cl}(q^r_{j})\|_{\bar 2} \| h_{r-1,j}\|_{\bar 2}\le  \eta\sqrt{DM}\, B\Lambda_\rho c,
\end{align*}
where the final inequalities result from Assumptions~\ref{assm:rho} and~\ref{assm:clipp}. For the output embedding, the output $y_j=\frac1DW_{\mathrm{out},j}^\top h_{L,j}$ satisfies $|y_j|_{\mathscr H_\star}\le s_jB$ and $\|y_j\|_\ell\le n_jB$. Hence $d_T(\mathscr H(sy_j),\mathscr H_\star)$ is less than $n_jB$ for $s\in[0,1]$, and thus \[g_j:=\nabla\ell(y_j,\mathfrak t)=g_0+\mathscr H_\star^{1/2}\widetilde M_{y_j,0}\mathscr H_\star^{1/2}y_j\] is bounded according to
\begin{equation}\label{eqn:grad_apriori}
|g_j|\le G_0+\sqrt{\lambda_\star}\,e^{n_jB}s_jB,\qquad\|g_j\|_\ell\le \varkappa_\ell,
\end{equation}
as $\|\widetilde M_{y_j,0}\|_{\mathrm{op}}\le e^{n_jB}$. The second bound comes from (L1). The update $W_{\mathrm{out},j+1}-W_{\mathrm{out},j}$ is given by $-\eta_{\mathrm{out}}h_{L,j}g_j^\top$ where $|g_j|_{\mathscr H_\star}\le\sqrt{\lambda_\star}|g_j|$ and $|g_j^\top g_0|\le G_0|g_j|$. Applying the above bounds yields the following recursions:
\[
s_{j+1}\le s_j+\eta_{\mathrm{out}}B\sqrt{\lambda_\star}\,|g_j|,\qquad n_{j+1}\le n_j+\eta_{\mathrm{out}}B\,\varkappa_\ell,\qquad r_{j+1}\le r_j+\eta_{\mathrm{out}}BG_0\,|g_j|, 
\]
where the bound on $n_j$ uses the positive homogeneity of $\|\cdot\|_\ell$ to factor out $\langle h_{L,j},h\rangle_{\bar{2}}$ in the definition of $\mathfrak{n}$ \eqref{eqn:mathfrak f&n}. Since $\|p^r_j\|_{\bar2}\le o_{r,j} B$, we have
\[
o_{r,j+1}\le(1+\eta\Lambda_\rho B^2)\,o_{r,j}+\eta\Lambda_\rho B(1+c),
\]
and the initial condition can grow by at most $\eta_{\mathrm{in}}|x|^2B$ from one step to the next. Hence, for $j\le k$, applying \eqref{eqn:grad_apriori} and Lemma \ref{lemma:embed_event} gives
\begin{align}\label{eqn:op_growth}
&o_{r,j}\le\alpha(B)\,o_{r,0}+\beta(B),
\qquad
s_j\le s(B),\\&n_j\le n(B),\qquad r_j\le r(B),
\qquad
\|h_{0,j}\|_{\bar2}\le C_{\mathrm{in}}+\eta_{\mathrm{in}} K|x|^2B,\notag
\end{align}
on $\mathcal E^W_{\delta/2}$, where
\[
\alpha(B):=e^{\eta K\Lambda_\rho B^2},\qquad\beta(B):=\alpha(B)\,\eta K\Lambda_\rho B(1+c),
\]
and
\begin{align*}
n(B)&:=C_{\mathfrak n}+\eta_{\mathrm{out}}KB\varkappa_\ell,\qquad s(B):=e^{\eta_{\mathrm{out}}K\lambda_\star B^2e^{n(B)B}}\big(C_{\mathfrak f}+\eta_{\mathrm{out}}KB\sqrt{\lambda_\star}\,G_0\big),\\
r(B)&:=C_r+\eta_{\mathrm{out}}KBG_0\big(G_0+\sqrt{\lambda_\star}\,e^{n(B)B}s(B)B\big).
\end{align*}
By Assumption~\ref{assm:rho}, the forward and backward passes satisfy
\[
\|h_{r,k}\|_{\bar2}\le\Big(1+\frac{\Lambda_\rho o_{r,k}^2}L\Big)\|h_{r-1,k}\|_{\bar2}+\frac{\Lambda_\rho o_{r,k}}L,
\qquad
\|a_{r-1,k}\|_{\bar2}\le\Big(1+\frac{\Lambda_\rho o_{r,k}^2}L\Big)\|a_{r,k}\|_{\bar2},
\]
so that the discrete Gr\"onwall inequality and \eqref{eqn:op_growth} bound the hidden states at the step $k$ by $b_k:=e^{\Lambda_\rho\bar o_k^2}\big(C_{\mathrm{in}}+\eta_{\mathrm{in}}K|x|^2B+\Lambda_\rho\bar o_k\big)$. The gradient of the loss at the step $k$ satisfies \eqref{eqn:grad_apriori} with $B$ replaced by $b_k$, since $s_k$ and $n_k$ only depend on the earlier steps. Hence, using $\|\widetilde M_{y_k,0}\|_{\mathrm{op}}\le e^{n_k b_k}$ and $|y_k|_{\mathscr H_\star}\le s_k b_k$, the terminal condition satisfies
\[
\|a_{L,k}\|_{\bar2}\le r_k+\big\|W_{\mathrm{out},k}\mathscr H_\star^{1/2}\widetilde M_{y_k,0}\mathscr H_\star^{1/2}y_k\big\|_{\bar2}\le r(B)+e^{n(B)b_k}s(B)^2b_k,
\]
and the backward pass gives
\begin{equation}\label{eqn:apriori_det}
\max_{0\le l\le L}\|h_{l,k}\|_{\bar2}\vee\|a_{l,k}\|_{\bar2}
\le b_k\vee e^{\Lambda_\rho\bar o_k^2}\big(r(B)+e^{n(B)b_k}s(B)^2b_k\big).
\end{equation}

\emph{Stage 2: The Event.} Each $o_{r,0}$ is the normalised operator norm of the initial parameters of a single layer. By Assumption~\ref{assm:MLU} and Lemma~\ref{lemma:OpNormLayerAvg}, applied to $\Phi^r_0$ and $\Psi^r_0$, for every $t\ge0$,
\[
\mathbb P\Big(o_{r,0}>a+\frac{C\sigma t}{\sqrt M}\Big)\le2e^{-t^2},\qquad a:=C\sigma\Big(1+\sqrt{\frac DM}\Big),
\]
for an absolute constant $C$. Write the excess of $o_{r,0}$ over $a$ as
\[
X_r:=\max\{o_{r,0}-a,\,0\},\qquad\text{so that}\qquad o_{r,0}^2\le2a^2+2X_r^2 .
\]
The variables $X_r$ are independent across the layers. By the tail bound, $\|X_r\|_{\psi_2}\le C\sigma/\sqrt M$ \cite[Proposition~2.6.1]{Vershynin_2026}. Hence $X_r^2$ is sub-exponential \cite[Lemma~2.8.6]{Vershynin_2026}, with
\[
\mathbb EX_r^2\le\frac{C\sigma^2}M,\qquad\big\|X_r^2-\mathbb EX_r^2\big\|_{\psi_1}\le\frac{C\sigma^2}M .
\]
Bernstein's inequality \cite[Theorem~2.9.1]{Vershynin_2026} for the average of the centred sub-exponential variables gives, with probability at least $1-\delta/2$,
\[
\left|\frac1L\sum_{r=1}^LX_r^2 -\E[X_r^2]\right|\le\frac{C\sigma^2}M\Big(\sqrt{\frac{\log(4/\delta)}{L}}+\frac{\log(4/\delta)}L\Big).
\]
Combining the last two inequalities, and using the restrictions on $D$ and $\delta$ in the statement,
\[
\bar o_0^2\le2a^2+\frac2L\sum_{r=1}^LX_r^2\le C\sigma^2\Big(1+\frac DM+ \sqrt{\frac{\log(4/\delta)}{M^2L}}+\frac{\log(4/\delta)}{LM}\Big)\le C\sigma^2(1+c_0+c_1)=:R^2 .
\]
Let $\mathcal E^{\mathrm{op}}_\delta$ be the intersection of this event with $\mathcal E^W_{\delta/2}$ of Lemma~\ref{lemma:embed_event}. Since $\delta/2\ge\exp(-(c_1+\log2)D)$, that lemma applies at the level $\delta/2$ with $c_1$ replaced by $c_1+\log2$, so that $\mathbb P(\mathcal E^{\mathrm{op}}_\delta)\ge 1-\delta$.

\emph{Stage 3: Induction over the training steps.} A leave-one-layer-out network has the same initial parameters, without those of the removed layer, so its $o_{r,0}$ are those of the full network, and the bound below on its average $\bar o_k^2$, a sum over the remaining layers $r\ne l^*$ with the normalisation $1/L$, is at most the same bound for the full network. Since $\mathcal E^{\mathrm{op}}_\delta$ only involves $\bar o_0$ and the initial embeddings, the bounds hold for every $l^*$ on this single event, without a union bound over $l^*$. Hence, by \eqref{eqn:op_growth}, under the hypothesis of Stage~1 and on $\mathcal E^{\mathrm{op}}_\delta$, every discrete network parameters satisfy
\[
\bar o_k^2\le\frac1L\sum_{r=1}^L\big(\alpha(B)\,o_{r,0}+\beta(B)\big)^2\le2\alpha(B)^2R^2+2\beta(B)^2 .
\]
Let $G(B)$ be the RHS of \eqref{eqn:apriori_det} with $\bar o_k^2$ replaced by this bound, and set
\[
B_0:=G(0),\qquad B_k:=B_{k-1}\vee G(B_{k-1}),\quad 1\le k\le K.
\]
At training step $k = 0$, there are no earlier steps, so the hypothesis of Stage 1 holds with $B = 0$. Therefore, the history at step 0 is bounded by $G(0) = B_0$. For $k \ge 1$, the $B_j$ are non-decreasing, so a history bounded by $B_j$ at each step $j < k$ is bounded by $B_{k-1}$ at all of them, and Stages~1 and~2 bound it by $G(B_{k-1}) \le B_k$ at step $k$. By induction over $k$, the lemma holds with, $C_B:=B_K$, $C'_{\mathfrak f}:=s(C_B)$ and $C'_{\mathfrak n}:=n(C_B)$.
\end{proof}

\begin{lemma}[The limiting model]\label{lemma:C1}
Suppose Assumptions~\ref{assm:MLU}--\ref{assm:Loss} hold. There exists $C_{\mathrm{lim}}$, depending only on $\eta K$, $\eta_{\mathrm{in}}|x|^2K$, $\eta_{\mathrm{out}}(\lambda_\star+\varkappa_\ell+G_0^2)K$, $\Lambda_\rho$, $\sigma$, $c$, $\lambda_\star$, $\varkappa_\ell$, $G_0$, $|x|$ and the constants $C_{\mathrm{in}}$, $C_{\mathfrak f}$, $C_r$ and $C_{\mathfrak n}$ of Lemma~\ref{lemma:embed_event}, such that, whenever the initial embeddings satisfy the bounds of Lemma~\ref{lemma:embed_event}, and in particular on $\mathcal E^W_{\delta/2}$ with $c_1$ replaced by $c_1+\log 2$, the limiting model of Section~\ref{sec:LimitingModel} is well defined, $\mathcal F_W$-measurable, with $\phi^\infty_{\cdot,k},\psi^\infty_{\cdot,k},P_{\cdot,k},Q_{\cdot,k}\in C([0,1],L^2)$ for every $k\le K$, and on $\mathcal E^W_{\delta/2}$ satisfies
\[
\max_{0\le k\le K}\sup_{t\in[0,1]}\|H_{t,k}\|_{\bar2}\vee\|A_{t,k}\|_{\bar2}\le C_{\mathrm{lim}},
\qquad
\mathfrak f(W^\infty_{\mathrm{out},k})\le s(C_{\mathrm{lim}}),\qquad\mathfrak n(W^\infty_{\mathrm{out},k})\le n(C_{\mathrm{lim}}),
\]
with $s$ and $n$ as in \eqref{eqn:op_growth}. Moreover, the constants of Lemma~\ref{lemma:apriori} may be chosen so that $C_{\mathrm{lim}}\le C_B$, and so, on $\mathcal E^{\mathrm{op}}_\delta$, the limiting model also satisfies the bounds of Lemma~\ref{lemma:apriori}.
\end{lemma}
\begin{proof}
We work on $\mathcal E^W_{\delta/2}$, on which Lemma \ref{lemma:embed_event} applies with $c_1$ replaced by $c_1+\log2$, and all expectations are over the initial parameters $(\phi_0,\psi_0)$ of a single neuron, conditionally on $\mathcal F_W$. The limiting model of Section~\ref{sec:LimitingModel} is the discrete model with the average over the $M$ neurons of a layer replaced by this expectation, and the sum over the layers replaced by an integral over the depth. We follow the proof of Lemma~\ref{lemma:apriori}, with the normalised operator norms replaced by
\[
o^\infty_{t,j}:=\sup_{u\in\mathbb B^D_2}\mathbb E\Big[\Big|\frac{u^\top\phi^\infty_{t,j}}{\sqrt D}\Big|^2\Big|\mathcal F_W\Big]^{1/2}\vee\sup_{u\in\mathbb B^D_2}\mathbb E\Big[\Big|\frac{u^\top\psi^\infty_{t,j}}{\sqrt D}\Big|^2\Big|\mathcal F_W\Big]^{1/2},
\]
so that, for every square-integrable scalar $X$ and $\mathcal F_W$-measurable $h\in\R^D$,
\begin{equation}\label{eqn:apriori_limit_op}
\big\|\mathbb E[\phi^\infty_{t,j}X|\mathcal F_W]\big\|_{\bar2}\le o^\infty_{t,j}\,\|X\|_{L^2},
\qquad
\big\|\langle\psi^\infty_{t,j},h\rangle_{\bar2}\big\|_{L^2|\mathcal F_W}\le o^\infty_{t,j}\,\|h\|_{\bar2},
\end{equation}
and similarly with $\phi$ and $\psi$ exchanged. At initialisation, the coordinates of $\phi_0$ and $\psi_0$ are i.i.d., centred, with $\psi_2$-norm $\sigma\sqrt D$ (Assumption~\ref{assm:MLU}), so that 
\[
o^\infty_{t,0}\le C\sigma=:R_\infty,
\]
for an absolute constant $C$.

\emph{Inductive Hypothesis.} Fix $0\le k\le K$ and suppose that, for some $B\ge0$, the limiting histories are well-defined, continuous in $t$, and satisfy
\[
\sup_{t\in[0,1]}\max_{j<k}
\big(\|H_{t,j}\|_{\bar2}\vee\|A_{t,j}\|_{\bar2}\big)\le B.
\]
For $k=0$, this hypothesis is vacuous. At initialisation, the neuron parameters are independent of $t$, and their initial operator bound is $R_\infty$.

\paragraph{}The parameters are updated according to
\[
\phi^\infty_{t,j+1}-\phi^\infty_{t,j}=-\eta A_{t,j}\rho(P_{t,j}),
\qquad
\psi^\infty_{t,j+1}-\psi^\infty_{t,j}=-\eta H_{t,j}\rho'(P_{t,j})\mathrm{cl}(Q_{t,j}),
\]
and so Minkowski's inequality in $L^2$ and \eqref{eqn:apriori_limit_op} give the recursion
\[
o^\infty_{t,j+1}\le(1+\eta\Lambda_\rho B^2)\,o^\infty_{t,j}+\eta\Lambda_\rho B(1+c),
\]
uniformly in $t\in[0,1]$ and $j< k$. This is the same recursion as Stage 1 in the proof of Lemma \ref{lemma:apriori} and thus
\[
o^\infty_{t,j}\le\alpha(B)R_\infty+\beta(B)=:o_\infty(B)
\]
for any $j\le k$ and $t\in[0,1]$. Furthermore, the recursions for $s_j$, $n_j$ and $r_j$ rely only on the boundedness of the hidden states and adjoint variables, and therefore also hold for the corresponding quantities of the limiting model, which have the same initial conditions.

\paragraph{Well-posedness.} By the inductive hypothesis, the histories at the earlier training steps are continuous in $t$. Since the neuron parameters are updated through continuous functions of these histories, $\phi^\infty_{\cdot,k},\psi^\infty_{\cdot,k}\in C([0,1];L^2)$, and the uniform moment bounds on $o^\infty_{t,k}$ imply that the expected forward vector field
\[
(t,H)\mapsto\mathbb E\big[\phi^\infty_{t,k}\rho(\langle\psi^\infty_{t,k},H\rangle_{\bar2})\mid\mathcal F_W\big]
\]
is continuous in $t$. By \eqref{eqn:apriori_limit_op} and Assumption~\ref{assm:rho}, it is globally Lipschitz in $H$, uniformly in $t$, with constant $\Lambda_\rho o_\infty(B)^2$. Thus Picard--Lindelöf \cite[Chapter~II, Theorem~1.1]{hartman2002ode} gives a unique continuous solution of the forward equation on $[0,1]$, since the Lipschitz constant is uniform in $t$. Given this solution, the backward equation is linear in the adjoint variable, with coefficients continuous in $t$ by the same argument, and likewise has a unique solution on $[0,1]$. The solutions are $\mathcal F_W$-measurable, since the vector fields and initial and terminal conditions are $\mathcal F_W$-measurable.

\paragraph{The Bounds.} By \eqref{eqn:apriori_limit_op} and the linear growth of $\rho$ (Assumption \ref{assm:rho}),
\[
\|\partial_tH_{t,k}\|_{\bar2}\le o_\infty(B)\,\Lambda_\rho\big(1+o_\infty(B)\|H_{t,k}\|_{\bar2}\big),
\qquad
\|\partial_tA_{t,k}\|_{\bar2}\le\Lambda_\rho\,o_\infty(B)^2\,\|A_{t,k}\|_{\bar2}.
\]
Gr\"onwall's inequality in $t$, upwards from $t=0$ and downwards from $t=1$, replaces the discrete Gr\"onwall inequality of Stage~1, and gives \eqref{eqn:apriori_det} with $\bar o_k$ replaced by $o_\infty(B)$ and the maximum over $l$ replaced by the supremum over $t$. Let $G_\infty(B)$ be its RHS, which is at most $G(B)$ with $R$ replaced by $R_\infty$. From the base case $B^\infty_0:=G_\infty(0)$ and $B^\infty_k:=B^\infty_{k-1}\vee G_\infty(B^\infty_{k-1})$. The induction of Stage~3 of the proof of Lemma \ref{lemma:apriori} then gives the bound on the history, with $C_{\mathrm{lim}}:=B^\infty_K$, which depends only on the constants in the statement. Then, by the same argument as the proof of Lemma \ref{lemma:apriori}, we have $\mathfrak f(W^\infty_{\mathrm{out},k})\le s(C_{\mathrm{lim}})$ and $\mathfrak n(W^\infty_{\mathrm{out},k})\le n(C_{\mathrm{lim}})$. Since no event other than $\mathcal E^W_\delta$ has been used, the bound is deterministic given $\mathcal F_W$.
\end{proof}
Throughout the rest of the paper, $C_B$, $C'_{\mathfrak f}$ and $C'_{\mathfrak n}$ denote the constants of Lemma~\ref{lemma:apriori}, and we work on the event $\mathcal E^{\mathrm{op}}_\delta$. Without loss of generality, by increasing $R$ in the proof of Lemma \ref{lemma:apriori}, we may take $C_B\ge C_{\mathrm{lim}}$ of Lemma \ref{lemma:C1}. As $s,n$ of \eqref{eqn:op_growth}, are increasing in $B$, $\mathfrak{f}(W^\infty_{\mathrm{out},k}), \mathfrak{n}(W^\infty_{\mathrm{out},k})$ are bounded, on $\mathcal E^W_{\delta/2}$, by $C_\mathfrak{f}',C_\mathfrak{n}'$, respectively. Thus, $C_B$ bounds every history used below: the history of the limiting model, the histories of the discrete model and of every leave-one-layer-out network, and the comparison histories of Section~\ref{sec:LOLO_notation}. Moreover, by \eqref{eqn:op_growth} with $B=C_B$, for each of these networks, every layer $r$ it contains and every $j\le K$,
\begin{equation}\label{eqn:op_growth_all}
o_{r,j}\le C'\varpi^r_2,\qquad \varpi^r_2:=1\vee o_{r,0},
\end{equation}
where $\varpi^r_2$ only depends on the initial parameters of layer $r$, which are shared by all the networks, and $C'$ only depends on $C_B$, $\eta K$, $\Lambda_\rho$ and $c$.

\paragraph{Effective learning rate.} Since $h_{0,k+1}-h_{0,k}=-\eta_{\mathrm{in}}|x|^2a_{0,k}$, and the update of $W_{\mathrm{out}}$ contributes $-\eta_{\mathrm{out}}h_{L,k}\,\nabla\ell(y_k)^\top\nabla\ell(y_{k+1})$ to $a_{L,k+1}$, the learning rate of the input embedding enters through $\eta_{\mathrm{in}}|x|^2$, and, as in Remark~\ref{rem:gw}, that of the output embedding enters through $\eta_{\mathrm{out}}(\lambda_\star+\varkappa_\ell+G_0^2)$. We denote
\begin{equation}\label{eqn:eta_bar}
\bar\eta:=\eta\vee\eta_{\mathrm{in}}|x|^2\vee\eta_{\mathrm{out}}\big(\lambda_\star+\varkappa_\ell+G_0^2\big),
\qquad
\tilde\eta_{\mathrm{out}}:=\eta_{\mathrm{out}}\big(\lambda_\star+G_0^2\big)\le\bar\eta,
\end{equation}
for the largest effective learning rate, and for the part of the learning rate of the output embedding that multiplies the curvature and the gradient of the loss at the origin, in the bounds on the output embedding.

\begin{lemma}[Boundary conditions]\label{lemma:boundary}
Suppose the assumptions of Lemma~\ref{lemma:apriori} hold. Let $(h,a,y,W_{\mathrm{in}},W_{\mathrm{out}})$ and $(h',a',y',W'_{\mathrm{in}},W'_{\mathrm{out}})$ be the hidden states, adjoint variables, outputs and embeddings of two of the following: the discrete model, a leave-one-layer-out network, and the limiting model, with $h'_{l,k}:=H_{\frac lL,k}$, $a'_{l,k}:=A_{\frac lL,k}$, $y'_k:=Y_k$ and $W'_{\cdot,k}:=W^\infty_{\cdot,k}$ for the latter. There is a constant $C_\partial$, depending only on $C_B$, $C'_{\mathfrak f}$, $R_*$, $S_*$ and $\tilde\eta_{\mathrm{out}}K$, such that, on $\mathcal E^{\mathrm{op}}_\delta$, for every $k\le K$,
\begin{align}
\|h_{0,k}-h'_{0,k}\|_{\bar2}&\le\eta_{\mathrm{in}}|x|^2\sum_{k'<k}\|a_{0,k'}-a'_{0,k'}\|_{\bar2},\label{eqn:boundary_input}\\
\|a_{L,k}-a'_{L,k}\|_{\bar2}&\le C_\partial\Big(\|h_{L,k}-h'_{L,k}\|_{\bar2}+\tilde\eta_{\mathrm{out}}\sum_{k'<k}\|h_{L,k'}-h'_{L,k'}\|_{\bar2}\Big).\label{eqn:boundary_terminal}
\end{align}
\end{lemma}
\begin{proof}
Both networks start from the same embeddings, so the update of $W_{\mathrm{in}}$ in \eqref{eqn:embedding_updates} gives \eqref{eqn:boundary_input}. \\

Write $e_k:=h_{L,k}-h'_{L,k}$, $g_k:=\nabla\ell(y_k,\mathfrak{t})$, $g'_k:=\nabla\ell(y'_k,\mathfrak{t})$, $g_0:=\nabla\ell(0,\mathfrak{t})$, and
\[
\Delta W_k:=W_{\mathrm{out},k}-W'_{\mathrm{out},k},\qquad
\mathscr F_k:=\mathfrak f(\Delta W_k),\qquad
\mathscr V_k:=\|\Delta W_k g_0\|_{\bar2},
\]
so that $\mathscr F_0=\mathscr V_0=0$. On $\mathcal E^{\mathrm{op}}_\delta$, both outputs lie in $\mathcal Y$, and hence $|g_k|\vee|g'_k|\le G_\ell$. By \eqref{eqn:grad_factor},
\[
g_k-g'_k
=\mathscr H_\star^{1/2}\widetilde M_{y_k,y'_k}
\mathscr H_\star^{1/2}(y_k-y'_k),
\qquad
\|\widetilde M_{y_k,y'_k}\|_{\mathrm{op}}\le e^{R_*}.
\]
Moreover, decomposing the output difference gives
\[
\begin{aligned}
|y_k-y'_k|_{\mathscr H_\star}
&\le \frac1D|W_{\mathrm{out},k}^\top e_k|_{\mathscr H_\star}
+\frac1D|\Delta W_k^\top h'_{L,k}|_{\mathscr H_\star}\le C'_{\mathfrak f}\|e_k\|_{\bar2}+C_B\mathscr F_k.
\end{aligned}
\]
In particular, $|g_k-g'_k|\le\sqrt{\lambda_\star}e^{R_*}|y_k-y'_k|_{\mathscr H_\star}$ and $|g_k-g'_k|_{\mathscr H_\star}\le\lambda_\star e^{R_*}|y_k-y'_k|_{\mathscr H_\star}$. Next, apply \eqref{eqn:embedding_updates} to deduce that \[\Delta W_{k+1}-\Delta W_k=-\eta_{\mathrm{out}}\big(e_kg_k^\top+h'_{L,k}(g_k-g'_k)^\top\big),\] and so
\begin{align*}
    \mathscr F_{k+1}&\le\mathscr F_k+\eta_{\mathrm{out}}\big(\sqrt{\lambda_\star}G_\ell\|e_k\|_{\bar2}+C_B\lambda_\star e^{R_*}|y_k-y'_k|_{\mathscr H_\star}\big),\\
\mathscr V_{k+1}&\le \mathscr V_k+\eta_{\mathrm{out}}G_0\big(G_\ell\|e_k\|_{\bar2}+C_B\sqrt{\lambda_\star}e^{R_*}|y_k-y'_k|_{\mathscr H_\star}\big).
\end{align*}
Inserting the bound on $|y_k-y'_k|_{\mathscr H_\star}$ into the first recursion gives\[\mathscr F_{k+1}\le\big(1+\eta_{\mathrm{out}}C_B^2\lambda_\star e^{R_*}\big)\mathscr F_k+\eta_{\mathrm{out}}\big(\sqrt{\lambda_\star}G_\ell+C_BC'_{\mathfrak f}\lambda_\star e^{R_*}\big)\|e_k\|_{\bar2},\]and the discrete Gr\"onwall inequality gives 
\[
\mathscr F_k\le\eta_{\mathrm{out}}e^{\eta_{\mathrm{out}}KC_B^2\lambda_\star e^{R_*}}\big(\sqrt{\lambda_\star}G_\ell+C_BC'_{\mathfrak f}\lambda_\star e^{R_*}\big)\sum_{k'<k}\|e_{k'}\|_{\bar2}.
\]
Since $G_\ell\le G_0+\sqrt{\lambda_\star}e^{R_*}S_*$ and $\sqrt{\lambda_\star}G_0\le(\lambda_\star+G_0^2)/2$, every coefficient in the two recursions, after multiplication by $\eta_{\mathrm{out}}$, is bounded by $C\widetilde\eta_{\mathrm{out}}$, with $C$ depending only on $C_B$, $C'_{\mathfrak f}$, $R_*$ and $S_*$. Thus, discrete Gr\"onwall, summing the recursion for $\mathscr{V}_k$ and using the bound on $|y-y'|_{\mathscr{H}_\star}$ give
\[
\mathscr F_k+\mathscr V_k
\le C\widetilde\eta_{\mathrm{out}}
\sum_{k'<k}\|e_{k'}\|_{\bar2},
\]
where $C$ depends only on $C_B$, $C'_{\mathfrak f}$, $R_*$, $S_*$ and $\widetilde\eta_{\mathrm{out}}K$. Finally, applying \eqref{eqn:grad_factor} with $y_k$ replaced by $0$, using $|y'_k|_{\mathscr H_\star}\le S_*$ and $\|W\mathscr H_\star^{1/2}v\|_{\bar2}\le\mathfrak f(W)|v|$, we obtain
\begin{align*}
\|a_{L,k}-a'_{L,k}\|_{\bar2}
&\le\|W_{\mathrm{out},k}(g_k-g'_k)\|_{\bar2}
+\|\Delta W_k\,g'_k\|_{\bar2}\\
&\le C'_{\mathfrak f}e^{R_*}|y_k-y'_k|_{\mathscr H_\star}
+\mathscr V_k+e^{R_*}S_*\mathscr F_k.
\end{align*}
Combining this with the preceding estimates gives \eqref{eqn:boundary_terminal}.
\end{proof}

\subsection{Preliminary Propagation Estimates}\label{sec:PropPrelim}
Equation~\eqref{eqn:VolterraResponse} describes how a perturbation of the history along which a layer is trained propagates to its features. The second term of $\mathcal B$ contains $\rho(P_{k',s})$, which is not bounded, so its magnitude depends on that of the features along the interpolated history. When one of the two histories is that of the discrete model, it depends on the parameters of the layer, and Lemma~\ref{lemma:p-norm_on_PQ} does not apply. Bounding the features of each layer by the operator norms of its initial parameters is always possible (see \eqref{eqn:PQ_dependent} below). However, it implicitly treats the parameters of a layer and the history as fully correlated, and, used on its own as \cite{chizat2025hiddenwidthdeepresnets} does, it leads to restrictive shape constraints. Instead, we compare the features along a dependent history $\bm Z$ with those along a history $\widetilde{\bm Z}$ that is close to $\bm Z$ but independent of the initial parameters of the layer, to which Lemma~\ref{lemma:p-norm_on_PQ} applies. Operator norms are then only used for the small differences between $\bm Z$ and $\widetilde{\bm Z}$. This is the content of Lemma~\ref{lemma:prop_error_prelim}, which we tackle after obtaining a few preliminary results.

\paragraph{Notation.} For a layer $1\le r\le L$ and $p\in[1,\infty)$, the normalised operator norm of $\R^{D\times M}$ matrix is
\[
\|(\Psi^r_0)^\top\|_{\bar2\to\bar p}=\sup_{u\in\mathbb B^D_2}\Big(\frac1M\sum_{m=1}^M\Big|\frac{u^\top\psi^{rm}_0}{\sqrt D}\Big|^p\Big)^{1/p},
\]
where $\mathbb B^D_2$ is the Euclidean unit ball of $\R^D$. Define
\[
\varpi^r_p:=1\vee\|(\Psi^r_0)^\top\|_{\bar2\to\bar p}\vee\|(\Phi^r_0)^\top\|_{\bar2\to\bar p},
\]
which only depends on the initial parameters of layer $r$, and is non-decreasing in $p$. For two histories $\bm Z^0_r,\bm Z^1_r$ of layer $r$ and $0\le k\le K$, we write
\[
\Delta^H_{r,\wedge k}:=\max_{0\le k'\le k}\|H^1_{r,k'}-H^0_{r,k'}\|_{\bar2},\quad
\Delta^A_{r,\wedge k}:=\max_{0\le k'\le k}\|A^1_{r,k'}-A^0_{r,k'}\|_{\bar2},\quad
\Delta_{r,\wedge k}:=\Delta^H_{r,\wedge k}\vee\Delta^A_{r,\wedge k}.
\]
Finally, for $d\ge 0$, $p\ge1$ and $\delta\in(0,1]$, to simplify our high probability bounds, we define
\begin{equation}\label{eqn:mathcal K}
\mathcal K^{d,M,p}_\delta := \sqrt{p}+ \frac{\sqrt{d+\log(1/\delta)}}{M^{1/p}} .
\end{equation}

Let $\bm Z$ be any history with $\|\bm Z\|_{\infty,\bar2}\le C_B$, possibly depending on the parameters, and let $p\in[1,\infty)$. In \eqref{eqn:p}--\eqref{eqn:q}, the first terms have $\bar p$-norm at most $C_B\varpi^r_p$, while $|\Gamma^{\bm H}|\vee|\Gamma^{\bm A}|\le C_B^2$, and the remaining terms grow linearly in $P,Q$. Hence, we have that
\[
\|P^r_k(\bm Z)\|_{\bar p}\vee\|Q^r_k(\bm Z)\|_{\bar p}\le  C_B\varpi^r_p+\eta \sum_{0 \le k'<k}C^2_B (c\vee 1)\Lambda_\rho(1+\|P^r_{k'}(\bm Z)\|_{\bar p}\vee\|Q^r_{k'}(\bm Z)\|_{\bar p}),
\]
which, using $\varpi^r_p\ge 1$, simplifies to
\begin{equation}\label{eqn:PQ_dependent}
\|P^r_k(\bm Z)\|_{\bar p}\vee\|Q^r_k(\bm Z)\|_{\bar p}\le C_{\mathrm f}\,\varpi^r_p,\qquad
C_{\mathrm f}:=\big(C_B+\eta K(c\vee1)\Lambda_\rho C_B^2\big)\,e^{\eta K(c\vee1)\Lambda_\rho C_B^2}.
\end{equation}

\begin{lemma} \label{lemma:DeltaP_prelim}
Suppose Assumptions~\ref{assm:rho} and~\ref{assm:clipp} hold, and let $C_B>0$, $1\le r\le L$ and $p\in[1,\infty)$. Further suppose we are given a pair of histories $\bm Z^0_r,\bm Z^1_r$ satisfying
\[
\|\bm Z^0_r\|_{\infty,\bar2}\vee\|\bm Z^1_r\|_{\infty,\bar2}\le C_B ,
\]
both of which may depend arbitrarily on the parameters. Then, there exists a constant $C_\Delta=C_\Delta(\eta K,\Lambda_\rho,c,C_B)$, non-decreasing in $C_B$, such that the following holds for every realisation of the initial parameters and for every $0\le k\le K$,
\begin{align*}
  \|P^{r}_{k}(\bm Z^1_r)-P^{r}_{k}(\bm Z^0_r)\|_{\bar p} & \le C_\Delta\,\varpi^r_p\left(\Delta^H_{r,\wedge k} +\mathbbm 1_{\{k>0\}}\Delta_{r,\wedge (k-1)} \right),\\
  \|Q^{r}_{k}(\bm Z^1_r)-Q^{r}_{k}(\bm Z^0_r)\|_{\bar p} & \le C_\Delta\,\varpi^r_p\left(\Delta^A_{r,\wedge k} +\mathbbm 1_{\{k>0\}}\Delta_{r,\wedge (k-1)} \right).
\end{align*}
Moreover, suppose that Assumption~\ref{assm:MLU} holds, and that $p\ge2$. Then there is an absolute constant $C$ such that, for every $\delta\in(0,1]$, with probability at least $1-\delta$, on an event that only depends on $(\Psi^r_0,\Phi^r_0)$, we have $\varpi^r_p\le C(1+\sigma)\mathcal K^{D,M,p}_\delta$. Hence, on this event, simultaneously for all such pairs of histories and all $0\le k\le K$,
\begin{align*}
  \|P^{r}_{k}(\bm Z^1_r)-P^{r}_{k}(\bm Z^0_r)\|_{\bar p} & \le C'\left(\Delta^H_{r,\wedge k} +\mathbbm 1_{\{k>0\}}\Delta_{r,\wedge (k-1)} \right)\mathcal K^{D,M,p}_{\delta},\\  \|Q^{r}_{k}(\bm Z^1_r)-Q^{r}_{k}(\bm Z^0_r)\|_{\bar p} & \le C'\left(\Delta^A_{r,\wedge k} +\mathbbm 1_{\{k>0\}}\Delta_{r,\wedge (k-1)} \right)\mathcal K^{D,M,p}_{\delta},
\end{align*}
with $C':=C(1+\sigma)C_\Delta$.
\end{lemma}
\begin{proof}
 We drop the index $r$ and write $\mathcal B^{m}_j(s)=(\mathcal B^{m,1}_j(s),\mathcal B^{m,2}_j(s))^\top$, where $\mathcal B$ is defined in Section~\ref{sec:LinearResponse}. Further define 
\[
\mathcal R^*_{j,k}:=\sup_{s\in[0,1]}\max_{1\le m\le M}\|\mathcal R^m_{j,k}(\bm Z^s)\|_{\mathrm{op}}\le\eta C_{\mathcal R}(1+\eta C_{\mathcal R})^{k-j-1},\qquad j<k,
\]
where $C_{\mathcal R}=C_{\mathcal R}(\Lambda_\rho,c,C_B)$ and the bound is Lemma~\ref{lemma:BoundMathcalR}. Integrating the first component of \eqref{eqn:VolterraResponse} over $s\in[0,1]$, and using $\mathcal R_{k,k}=I_2$, gives
\begin{equation}\label{eqn:ExpandDeltaP}
|P^{m}_{k}(\bm Z^1)-P^{m}_{k}(\bm Z^0)|  \le \int_0^1\Big(|\mathcal B^{m,1}_k(s)|+\sum_{0 \le j < k}\mathcal R_{j,k}^*\big(|\mathcal B^{m,1}_j(s)|+|\mathcal B^{m,2}_j(s)|\big)\Big)ds,
\end{equation}
and the same bound holds for $|Q^m_k(\bm Z^1)-Q^m_k(\bm Z^0)|$, with $\mathcal B^{m,1}_k$ replaced by $\mathcal B^{m,2}_k$ in the first term. Since $\|H^s_j\|_{\bar 2}\le C_B$, we have $|\partial_s\Gamma^{\bm H^s}_{i,j}|\le 2C_B\Delta^H_{\wedge j}$ for $i<j$, and similarly for $\bm A$. Hence, by Assumption~\ref{assm:rho},
\begin{align*}
|\mathcal B^{m,1}_j(s)|&\le|\langle\psi^m_0,H^1_j-H^0_j\rangle_{\bar2}|+2\eta\Lambda_\rho cC_B\,\Delta^H_{\wedge j}\sum_{0\le i<j}\big(1+|Q^m_{i,s}|\big),\\
|\mathcal B^{m,2}_j(s)|&\le|\langle\phi^m_0,A^1_j-A^0_j\rangle_{\bar2}|+2\eta \Lambda_\rho C_B\,\Delta^A_{\wedge j}\sum_{0\le i<j}\big(1+|P^m_{i,s}|\big).
\end{align*}
We now take the $\bar p$-norm over the neurons $m$. By \eqref{eqn:PQ_dependent} along $\bm Z^s$, Minkowski's inequality, and using $\varpi_p\ge1$, we obtain, uniformly in $s$,
\[
\|\mathcal B^{1}_j(s)\|_{\bar p}\le C_{\mathcal B}\,\varpi_p\,\Delta^H_{\wedge j},\qquad
\|\mathcal B^{2}_j(s)\|_{\bar p}\le C_{\mathcal B}\,\varpi_p\,\Delta^A_{\wedge j},\qquad
C_{\mathcal B}:=1+2\eta K\Lambda_\rho C_B(c\vee 1)(1+C_{\mathrm f}),
\]
where $\mathcal B^{i}_j(s):=(\mathcal B^{m,i}_j(s))_{m\le M}$. Since $\mathcal R^*_{j,k}$ does not depend on $m$ or $s$, taking the $\bar p$-norm over $m$ in \eqref{eqn:ExpandDeltaP} and applying Minkowski's integral inequality, we obtain
\[
\|P_k(\bm Z^1)-P_k(\bm Z^0)\|_{\bar p}\le C_{\mathcal B}\varpi_p\Big(\Delta^H_{\wedge k}+2\Delta_{\wedge(k-1)}\sum_{0\le j<k}\mathcal R^*_{j,k}\Big).
\]
Here the sum is empty if $k=0$. Otherwise, the series on $\mathcal R^*_{j,k}$ simplifies according to Lemma \ref{lemma:BoundMathcalR}, which proves the first claim. The bound on $Q$ is identical and all the constants are non-decreasing in $C_B$.

\paragraph{}For the second claim, with $p\ge 2$, notice that $\|(\Psi^r_0)^\top\|_{\bar2\to\bar p}=C^{\{r\}}_{1,p}(\Psi_0)$, in the notation \eqref{eqn:C_{p,q}^{l}} below, and similarly for $\Phi^r_0$. Under Assumption~\ref{assm:MLU}, Lemma~\ref{lemma:OpNormLayerAvg} applies with $n=1$. Since $p\ge2$, it gives $C^{\{r\}}_{1,p}(\Psi_0)\le\sigma C'\mathcal K^{D,M,p}_{\delta}$ with probability at least $1-\delta/2$, and similarly for $\Phi^r_0$. Here, the bound has been simplified by applying $\sqrt D+\sqrt{\log(2/\delta)}\le1+\sqrt{2}\sqrt{D+\log(1/\delta)}$. Therefore, with $p\ge 2$ and probability at least $1-\delta$,
\begin{equation}\label{eqn:single_layer_opnorm}
\varpi^r_p\le C(1+\sigma)\,\mathcal K^{D,M,p}_\delta ,
\end{equation}
for an absolute constant $C$. Substituting this into the first claim gives the second.
\end{proof}

The randomness in Lemma~\ref{lemma:DeltaP_prelim} only enters through $\varpi^r_p$, and keeping these random variables explicit allows the bounds to be averaged over the layers before any probability is taken, which is how the next lemma avoids $\log(L)$ factors.

\paragraph{}For a family $\theta=(\theta^{rm})_{1\le r\le L,1\le m\le M}$ of vectors in $\R^D$, a set of layers $\mathcal J\subseteq\{1,\ldots,L\}$ and $\mathfrak p,\mathfrak q\in[1,\infty]$, let
\begin{align}
         C^{\mathcal J}_{\mathfrak{p},\mathfrak{q}}(\theta) &:= \sup_{u \in \mathbb B_2^{D}} \left( \sum_{r\in\mathcal J} \left(\frac{1}{M}\sum_{m=1}^M \left|\frac{u^\top \theta^{rm}}{\sqrt{D}}\right|^\mathfrak{q}  \right)^{\mathfrak{p}/\mathfrak{q}}\right)^{1/\mathfrak{p}},\label{eqn:C_{p,q}^{l}}
\end{align}
with the maximum convention  if $\mathfrak q=\infty$ or $\mathfrak p=\infty$, and write $C^{l,l'}_{\mathfrak p,\mathfrak q}:=C^{\{l,\ldots,l'\}}_{\mathfrak p,\mathfrak q}$ and \\ $C^{l}_{\mathfrak p,\mathfrak q}:=C^{1,l}_{\mathfrak p,\mathfrak q}$. 
For a set of layers $\mathcal J$ and per-layer quantities $(x_r)_{r\in\mathcal J}$ with $x_r\in \R$, we write $x_{\mathcal J}:=(x_r)_{r\in\mathcal J}$ and $$\|x_{\mathcal J}\|_a:=(\sum_{r\in\mathcal J}|x_r|^a)^{1/a}$$ for the unnormalised $\ell^a$-norm over the layers. For vectors $X^r\in\R^M$ indexed by the neurons, we write \\$\|X^{\mathcal J}\|_{a,\bar b}:=(\sum_{r\in\mathcal J}\|X^r\|^a_{\bar b})^{1/a}$.

\begin{lemma}\label{lemma:prop_error_prelim}
Suppose Assumptions~\ref{assm:MLU}, \ref{assm:rho} and \ref{assm:clipp} hold, and let $f\in C^1(\R^2)$ be such that, for every $p,q\in \R$,
    \[
    | \partial_pf(p,q) |\le c'(1+|q|), \qquad | \partial_q f(p,q) |\le c' ,
    \]
for some constant $c'>0$. Let $C_B>0$, $p,q\in[1,\infty)$, $\delta\in(0,1]$ and let $\mathcal I\subseteq\{1,\ldots,L\}$ be a non-empty set of layers, with $n:=|\mathcal I|$. For each $r\in\mathcal I$, we take a  \emph{comparison history} $\widetilde{\bm Z}_r$ of layer $r$, that is, a history with $\|\widetilde{\bm Z}_r\|_{\infty,\bar2}\le C_B$ which is $\mathcal F_{-r}$-measurable. Then, there exist an event $\mathcal G_\delta$, which only depends on $(\Psi^r_0,\Phi^r_0,\widetilde{\bm Z}_r)_{r\in\mathcal I}$ and satisfies $\mathbb P(\mathcal G_\delta)\ge1-\delta$, and a constant $C_4=C_4(c',C_B,K,\eta K,\Lambda_\rho,c,\sigma)$, non-decreasing in $C_B$ and independent of $p,q,\mathcal I,L,M,D$ and $\delta$, such that the following holds on $\mathcal G_\delta$: Simultaneously for
\begin{itemize}
\item every pair of families of histories $\bm Z^0=(\bm Z^0_r)_{r\in\mathcal I}$ and $\bm Z^1=(\bm Z^1_r)_{r\in\mathcal I}$ with \\$\max_{r\in\mathcal I}\|\bm Z^0_r\|_{\infty,\bar2}\vee\|\bm Z^1_r\|_{\infty,\bar2}\le C_B$, which may depend arbitrarily on the parameters,
\item every family $\theta=(\theta^{rm})$ of vectors in $\R^D$, which may depend arbitrarily on the parameters, on the histories and on $k$,
\item every $\mathcal J\subseteq\mathcal I$ and every $0\le k\le K$,
\end{itemize}
the residual sums
\[
F^i_{\mathcal J,k} :=\frac{1}{LM}\sum_{r\in\mathcal J}\sum_{m=1}^M \theta^{rm} f\big(P_k^{rm}(\bm{Z}^i_r), Q_k^{rm}(\bm{Z}^i_r)\big),\qquad i\in\{0,1\},
\]
satisfy
\begin{align*}
           \| F^1_{\mathcal J,k}-F^0_{\mathcal J,k}\|_{\bar 2} \le{}& \frac{C_4\sqrt{n}}{L} \left[C_{2,p^*}^{\mathcal J}(\theta) \big(\mathcal K_\delta^{D,M,2p}\big)^2\frac{\|\widetilde{\Delta}_{\mathcal J, \wedge k}\|_{ 8}}{n^{1/8}}+C_{2,2q^*}^{\mathcal J}(\theta)\,\mathcal K_\delta^{D,M,2}\,\mathcal K_\delta^{0,M,2q}\right]\frac{\|\Delta_{\mathcal J, \wedge k}\|_{8}}{n^{1/8}}\\& + \frac{C_4\sqrt{n}}{L} C_{2,2}^{\mathcal J}(\theta)\,\mathcal K_\delta^{D,M,2}\,\frac{\| \Delta_{\mathcal J, \wedge k}\|_{ 4}}{n^{1/4}}.
\end{align*}
Here $p^*,q^*$ are the H\"older conjugates of $p,q$, $C^{\mathcal J}_{\mathfrak p,\mathfrak q}$ is defined in \eqref{eqn:C_{p,q}^{l}} and $\mathcal K$ in \eqref{eqn:mathcal K}, $\Delta_{r,\wedge k}$ is the difference between $\bm Z^1_r$ and $\bm Z^0_r$ up to the step $k$ (as in Lemma~\ref{lemma:DeltaP_prelim}), and
\[
\widetilde\Delta_{r,\wedge k}:=\max_{0\le k'\le k}\|\bm H^0_{r-1,k'}-\widetilde{\bm H}_{r-1,k'}\|_{\bar2}\vee \|\bm A^0_{r,k'}-\widetilde{\bm A}_{r,k'}\|_{\bar2}
\]
is the difference between $\bm Z^0_r$ and the comparison history. 
\end{lemma}

\begin{proof}
Throughout, $C$ denotes a constant depending only on $c',C_B,K,\eta K,\Lambda_\rho,c$ and $\sigma$, which may change from line to line. Fix $\mathcal J\subseteq\mathcal I$, $0\le k\le K$ and abbreviate
\[
Q^{rm}_i:=Q^{rm}_k(\bm Z^i_r),\quad \widetilde Q^{rm}:=Q^{rm}_k(\widetilde{\bm Z}_r),\quad \Delta Q^{rm}:=Q^{rm}_1-Q^{rm}_0,
\]
and analogously for $P$. By the mean value theorem, applied first in $q$ and then in $p$,
\[
|f(P^{rm}_1,Q^{rm}_1)-f(P^{rm}_0,Q^{rm}_0)|\le c'|\Delta Q^{rm}|+c'(1+|Q^{rm}_0|)|\Delta P^{rm}|,
\]
Accordingly, via passing through $\widetilde{Q}$, we obtain
\begin{align*}
\|&F^1_{\mathcal J,k}-F^0_{\mathcal J,k}\|_{\bar 2} = \sup_{u \in \mathbb B^D_2} \frac{1}{LM}\sum_{r\in\mathcal J}\sum_{m=1}^M \frac{u^\top \theta^{rm}}{\sqrt{D}} \left(f(P_{1}^{rm}, Q_{1}^{rm})-f(P_{0}^{rm}, Q_{0}^{rm})\right)
\\&\le\sup_{u \in \mathbb B^D_2} \frac{c'}{LM}\sum_{r\in\mathcal J}\sum_{m=1}^M\left|\frac{u^\top \theta^{rm}}{\sqrt{D}} \right| \Big(| \Delta Q^{rm}|+|\Delta P^{rm}|+|Q^{rm}_0-\widetilde Q^{rm}|\,|\Delta P^{rm}|+|\widetilde Q^{rm}|\,|\Delta P^{rm}|\Big).
\end{align*}
For each layer, we apply H\"older's inequality over the neurons with the exponents $(2,2)$ for the first two terms, $(p^*,2p,2p)$ for the third, and $(2q^*,2q,2)$ for the fourth. Over the layers, we apply H\"older's inequality with the exponents $(2,2)$ for the first two terms and $(2,4,4)$ for the last two. Identifying the resulting expression with $C^\mathcal J_{2,\mathfrak{q}}$ from \eqref{eqn:C_{p,q}^{l}} yields
   \begin{align}\label{eqn:DeltaF}
                 \|F^1_{\mathcal J,k}-F^0_{\mathcal J,k}\|_{\bar 2}\le{}&\frac{c'}{L}C^{\mathcal J}_{2,2}(\theta)\left(\| \Delta Q^{\mathcal J}\|_{ 2,\bar{2}}+\| \Delta P^{\mathcal J}\|_{ 2,\bar{2}}\right)+\frac{c'}{L}C^{\mathcal J}_{2,p^*}(\theta)\| Q^{\mathcal J}_{0}-\widetilde{Q}^{\mathcal J}\|_{ 4,\overline{2p}}\,\| \Delta P^{\mathcal J}\|_{ 4,\overline{2p}}\notag\\&+\frac{c'}{L}C^{\mathcal J}_{2,2q^*}(\theta)\,\|\widetilde{Q}^{\mathcal J}\|_{ 4,\overline{2q}}\,\| \Delta P^{\mathcal J}\|_{ 4,\bar{2}}.
   \end{align}
   For $a\ge1$ and $p'\in[1,\infty]$, let
\[
O_{p',a}:=\Big(\sum_{r\in\mathcal I}(\varpi^r_{p'})^a\Big)^{1/a},\qquad
\mathcal Q_{2q}:=\Big(\sum_{r\in\mathcal I}\max_{0\le j\le K}\|Q^r_j(\widetilde{\bm Z}_r)\|^4_{\overline{2q}}\Big)^{1/4},
\]
which only depend on $(\Psi^r_0,\Phi^r_0,\widetilde{\bm Z}_r)_{r\in\mathcal I}$. Lemma~\ref{lemma:DeltaP_prelim} can be applied to each of $\|\Delta P^r\|_{\bar{p'}}$, $\|\Delta Q^r\|_{\bar{p'}}$ and $\|Q^r_0-\widetilde Q^r\|_{\overline{2p}}$. Hence, for some constant $C_\Delta>0$, by the Cauchy--Schwarz inequality over $r\in\mathcal J$ and the inclusion $\mathcal J\subseteq\mathcal I$,
\begin{align}
\| \Delta Q^{\mathcal J}\|_{ 2,\bar{2}}+\| \Delta P^{\mathcal J}\|_{ 2,\bar{2}}&\le4C_\Delta\Big(\sum_{r\in\mathcal J}(\varpi^r_2)^2\Delta_{r,\wedge k}^2\Big)^{1/2}\le4C_\Delta\,O_{2,4}\,\|\Delta_{\mathcal J,\wedge k}\|_4,\label{eqn:DeltaPQ_layers}\\
\| Q^{\mathcal J}_{0}-\widetilde{Q}^{\mathcal J}\|_{ 4,\overline{2p}}&\le 2C_\Delta\Big(\sum_{r\in\mathcal J}(\varpi^r_{2p})^4\widetilde\Delta^4_{r,\wedge k}\Big)^{1/4}\le 2C_\Delta\,O_{2p,8}\,\|\widetilde\Delta_{\mathcal J,\wedge k}\|_8,\label{eqn:Avg_Q_over_layers}
\end{align}
with $\|\Delta P^{\mathcal J}\|_{4,\overline{2p}}$ and $\|\Delta P^{\mathcal J}\|_{4,\bar 2}$ bounded similarly. Finally, $\|\widetilde Q^{\mathcal J}\|_{4,\overline{2q}}\le\mathcal Q_{2q}$ as $\mathcal J \subseteq \mathcal I$. Substituting these bounds into \eqref{eqn:DeltaF} gives, for every realisation of the initial parameters,
\begin{align}\label{eqn:DeltaF_det}
\|F^1_{\mathcal J,k}-F^0_{\mathcal J,k}\|_{\bar 2}\le\frac{C}{L}\Big[&C^{\mathcal J}_{2,2}(\theta)\,O_{2,4}\,\|\Delta_{\mathcal J,\wedge k}\|_4+C^{\mathcal J}_{2,p^*}(\theta)\,O_{2p,8}^2\,\|\widetilde\Delta_{\mathcal J,\wedge k}\|_8\|\Delta_{\mathcal J,\wedge k}\|_8\notag\\&+C^{\mathcal J}_{2,2q^*}(\theta)\,\mathcal Q_{2q}\,O_{2,8}\,\|\Delta_{\mathcal J,\wedge k}\|_8\Big].
\end{align}
 By \eqref{eqn:single_layer_opnorm}, each $\varpi^r_{p'}$ with $p'\ge2$ has a tail of the form required by Lemma~\ref{lemma:LogRemoval_mixed}. So does $\max_{j\le K}\|Q^r_j(\widetilde{\bm Z}_r)\|_{\overline{2q}}$, by Lemma~\ref{lemma:p-norm_on_PQ}, which applies since $\widetilde{\bm Z}_r$ is bounded by $C_B$ and independent of $(\Psi^r_0,\Phi^r_0)$, and a union bound over the $K+1$ steps. Lemma~\ref{lemma:LogRemoval_mixed}, with uniform weights over $\mathcal I$, uses no independence across the layers. Applied for $(\mathfrak p,\mathfrak{q})=(2,4),(2,8)$, and $(2p,8)$ and the features at level $\delta/4$ each, it gives an event $\mathcal G_\delta$, depending only on $(\Psi^r_0,\Phi^r_0,\widetilde{\bm Z}_r)_{r\in\mathcal I}$, with $\mathbb P(\mathcal G_\delta)\ge1-\delta$, on which
\[
O_{\mathfrak{p},\mathfrak{q}}\le Cn^{1/\mathfrak{q}}\mathcal K^{D,M,\mathfrak{p}}_\delta,\qquad \mathcal Q_{2q}\le Cn^{1/4}\mathcal K^{0,M,2q}_\delta .
\]
Substituting these bounds into \eqref{eqn:DeltaF_det} gives the main estimate on $\mathcal G_\delta$. Every other bound used is deterministic given the initial parameters, so the estimate holds simultaneously for all admissible $\bm Z^0,\bm Z^1,\theta,\mathcal J$ and $k$. 
\end{proof}

\subsection{Leave-One-Layer-Out}\label{sec:LOLO}
\begin{definition}[Leave-one-layer-out network]\label{def:LOLO}
Fix a layer $l^*\in\{1,\ldots,L\}$. The \emph{leave-one-layer-out network} is the ResNet of Section~\ref{sec:Setup} with the residual of layer $l^*$ deleted, at every training step, and with the other layers initialised, and trained, as in the full network. That is, for $r\ne l^*$, the initial parameters are $\phi^{(-l^*),rm}_0:=\phi^{rm}_0$ and $\psi^{(-l^*),rm}_0:=\psi^{rm}_0$, and, for $k\ge0$,
\begin{align*}
h^{(-l^*)}_{0,k}&=W^{(-l^*)}_{\mathrm{in},k}x,\qquad
h^{(-l^*)}_{l,k}=h^{(-l^*)}_{l-1,k}+\mathbbm 1_{\{l\ne l^*\}}\frac1{LM}\sum_{m=1}^M\phi^{(-l^*),lm}_k\,\rho\big(p^{(-l^*),lm}_k\big),\\
a^{(-l^*)}_{L,k}&=W^{(-l^*)}_{\mathrm{out},k}\nabla\ell\big(y^{(-l^*)}_k\big),\qquad
a^{(-l^*)}_{l-1,k}=a^{(-l^*)}_{l,k}+\frac{\mathbbm 1_{\{l\ne l^*\}}}{LM}\sum_{m=1}^M\psi^{(-l^*),lm}_k\,\rho'\big(p^{(-l^*),lm}_k\big)\,q^{(-l^*),lm}_k,
\end{align*}
where $y^{(-l^*)}_k:=\frac1D(W^{(-l^*)}_{\mathrm{out},k})^\top h^{(-l^*)}_{L,k}$, the features are
\[
p^{(-l^*),lm}_k:=\langle\psi^{(-l^*),lm}_k,h^{(-l^*)}_{l-1,k}\rangle_{\bar2},
\qquad
q^{(-l^*),lm}_k:=\langle\phi^{(-l^*),lm}_k,a^{(-l^*)}_{l,k}\rangle_{\bar2},
\]
and the parameters of every layer $r\ne l^*$ are updated by the gradient descent of Section~\ref{sec:Setup}, along the hidden states and adjoint variables of the leave-one-layer-out network:
\begin{align*}
\phi^{(-l^*),rm}_{k+1}&=\phi^{(-l^*),rm}_k-\eta\,a^{(-l^*)}_{r,k}\,\rho\big(p^{(-l^*),rm}_k\big),\\
\psi^{(-l^*),rm}_{k+1}&=\psi^{(-l^*),rm}_k-\eta\,h^{(-l^*)}_{r-1,k}\,\rho'\big(p^{(-l^*),rm}_k\big)\,\mathrm{cl}\big(q^{(-l^*),rm}_k\big).
\end{align*}
The embeddings start from $W_{\mathrm{in},0}$ and $W_{\mathrm{out},0}$ and are updated by \eqref{eqn:embedding_updates}, along $a^{(-l^*)}_{0,k}$, $h^{(-l^*)}_{L,k}$ and $y^{(-l^*)}_k$. Layer $l^*$ has no parameters in this network. We write $\bm z^{(-l^*)}_{r,k}:=(h^{(-l^*)}_{r-1,k},a^{(-l^*)}_{r,k})$ for its history at layer $r$, and $\Phi^{(-l^*),r}_k$, $\Psi^{(-l^*),r}_k$, $p^{(-l^*),r}_k$, $q^{(-l^*),r}_k$ for the parameters and features of layer $r\neq l^*$.
\end{definition}

The main result of this section is the following estimate on the leave-one-layer-out errors, on average over the removed layer. The notation $\Delta^{(-r)}_{\wedge K}$ and the event $\mathcal E_\delta$ are introduced in Section~\ref{sec:LOLO_notation}, together with the blocks $I_j$ of length at most $\mathfrak m=\lceil L/N_0\rceil$.
\begin{proposition}[Leave-one-layer-out estimate]\label{lemma:MainLemma}
Suppose Assumptions~\ref{assm:MLU}--\ref{assm:Loss} and the shape and failure probability constraints of \eqref{eqn:S&FPconstraints} and \eqref{eqn:Final_shape_constraints} hold, as well as that $L\ge N_0$, for some $N_0$ chosen according to \eqref{eqn:N0_choice},  $\delta\ge\exp(-c_1M^{1/3})$. Then $\mathbb P(\mathcal E_\delta)\ge1-\delta$ and, on $\mathcal E_\delta\cap\mathcal E^{\mathrm{op}}_\delta$,
\begin{equation}\label{eqn:MainLemma}
\max_{0\le j<N_0}
\left(
\frac1{\mathfrak m}
\sum_{r\in I_j}
\|\Delta^{(-r)}_{\wedge K}\|_\infty^8
\right)^{1/8}
\le
C_K\left(
\frac1L+
\frac{D+\log(2N_0/\delta)}{LM}
\right),
\end{equation}
where $C_K$ and $\zeta_*$ depend on the training horizon $K$ and the constants in the assumptions, but not on $L,M,D$.
\end{proposition}

\subsubsection{Notation, Impulses and the Common Event}\label{sec:LOLO_notation}
\paragraph{Constants.} Throughout this proof, $C'>0$ denotes a constant, which may change from line to line, depending only on $C_{\mathrm{lim}}$, $C_B$, $C'_{\mathfrak f}$, $C'_{\mathfrak n}$, $C_r$, $G_0$, $\tau_\star$, $\varkappa_\ell$, $|x|$, $\eta$, $\eta_{\mathrm{in}}$, $\tilde\eta_{\mathrm{out}}$, $K$, $\Lambda_\rho$, $c$, $\sigma$, $c_0$, $c_1$ and absolute constants. In particular, $C'$ does not depend on $L,M,D,\delta$, nor on the number $N_0$ of blocks of the partition below, where all dependence on $N_0$ is kept explicit. We work on the event $\mathcal E^{\mathrm{op}}_\delta$ of Lemma~\ref{lemma:apriori} throughout, so that every history below is bounded by $C_B$.

\paragraph{The partition.} Let $N_0\ge1$ be an integer, which is fixed in \eqref{eqn:N0_choice} below, and assume $L\ge N_0$. The results of Sections~\ref{sec:LOLO_notation}--\ref{sec:LOLO_diag} hold for every $N_0\ge1$ with $L\ge N_0$, for the corresponding partition and event $\mathcal E_\delta$, with constants that do not depend on $N_0$. Hence fixing $N_0$ afterwards, in terms of these constants, is not circular. Both the forward and the backward pass use the partition of the layers into the $N_0$ blocks
\[
l_i:=\lfloor iL/N_0\rfloor,
\qquad
I_i:=\{l_i+1,\ldots,l_{i+1}\},
\qquad 0\le i<N_0.
\]
Each block is non-empty with length at most $\lceil L/N_0\rceil$, so that
\[
\mathfrak m:=\max_{0\le i<N_0}|I_i|=\lceil L/N_0\rceil,
\qquad
\varepsilon:=\frac{\mathfrak m}{L}\le\frac1{N_0}+\frac1L\le\frac2{N_0}.
\]
Averages over a block are normalised by $1/\mathfrak m$, including for blocks of length $\lfloor L/N_0\rfloor$. Where convenient, we write $N:=N_0$.

\paragraph{Errors.} Layer $r$ of a network sees the network only through the pair $(h_{r-1,k},a_{r,k})$. For a removed layer $l^*$, a step $k$ and $0\le r\le L+1$, let
\[
\Delta^{(-l^*),H}_{r,k}:=\big\|h_{r-1,k}-h^{(-l^*)}_{r-1,k}\big\|_{\bar2},
\qquad
\Delta^{(-l^*),A}_{r,k}:=\big\|a_{r,k}-a^{(-l^*)}_{r,k}\big\|_{\bar2},
\]
with the conventions $\Delta^{H}_{0,k}:=0$ and $\Delta^{A}_{L+1,k}:=0$, and $\Delta^{(-l^*)}_{r,k}:=\Delta^{(-l^*),H}_{r,k}\vee\Delta^{(-l^*),A}_{r,k}$.
% Thus the index $r$ runs over the inputs of the residual of layer $r$, and the range $0\le r\le L+1$ covers all the hidden states $h_0,\ldots,h_L$ and all the adjoint variables $a_0,\ldots,a_L$.
For a set of indices $\mathcal I\subseteq\{0,\ldots,L+1\}$ and a step $k$, write
\[
\Delta^{(-l^*)}_{\mathcal I,\wedge k}:=\max_{r\in\mathcal I}\max_{0\le k'\le k}\Delta^{(-l^*)}_{r,k'},
\qquad
\|\Delta^{(-l^*)}_{\wedge k}\|_\infty:=\Delta^{(-l^*)}_{[0,L+1],\wedge k},
\]
and similarly with the superscripts $H$ and $A$.

\paragraph{Operator norms.} Let $o_{r,j}$ be as in the proof of Lemma~\ref{lemma:apriori}, so that \eqref{eqn:op_growth_all} holds. By Assumption~\ref{assm:rho} and the uniform bound $C_B$ on the histories, for the discrete model and every leave-one-layer-out network, at every layer $r$ present in the network and every $j\le K$,
\begin{equation}\label{eqn:feature_op}
\|p^r_j\|_{\bar2}\vee\|q^r_j\|_{\bar2}\le C'\varpi^r_2,
\qquad
\|\bm\rho(p^r_j)\|_{\bar2}\vee
\|\bm\rho'(p^r_j)\odot q^r_j\|_{\bar2}
\le C'\varpi^r_2.
\end{equation}
For a block $I_i$ and $s\ge1$, let
\[
\overline\varpi_{I_i,s}:=\Big(\frac1{\mathfrak m}\sum_{r\in I_i}(\varpi^r_2)^s\Big)^{1/s}.
\]

\paragraph{Impulses.} The only quantities in this proof that depend on the
parameters of a single layer are the residual contributions
of the removed layer. For $1\le r\le L$, define the
\emph{impulses}
\begin{equation}
\label{eqn:impulse_def}
\iota^H_r:=\frac1M\max_{0\le k\le K}
\left\|\Phi^{r}_{k}\,
\bm\rho(p^{r}_{k})\right\|_{\bar2},
\qquad
\iota^A_r:=\frac1M\max_{0\le k\le K}
\left\|\Psi^{r}_{k}
\left(\bm\rho'(p^{r}_{k})\odot
q^{r}_{k}\right)\right\|_{\bar2},
\end{equation}
and $\iota_r:=\iota^H_r\vee\iota^A_r$. Removing layer $r$
perturbs the hidden state and the adjoint at layer $r$ by at
most $\iota^H_r/L$ and $\iota^A_r/L$, respectively.
By \eqref{eqn:op_growth_all} and \eqref{eqn:feature_op}, $\iota_r\le C'(\varpi^r_2)^2$ on $\mathcal E^{\mathrm{op}}_\delta$.
The impulses are kept explicit throughout the recursions below.

\paragraph{The Confidence Level.} All the events below are intersected over the $N_0$ blocks, and, within each block, over a number of applications of Lemmas~\ref{lemma:OpNormLayerAvg}, \ref{lemma:p-norm_on_PQ}, \ref{lemma:prop_error_prelim} and~\ref{lemma:LogRemoval_mixed} that depends only on $K$, each of them already holding simultaneously for all the $K+1$ steps. Let $c_K$ be this number, and apply each at the level
\[
\delta':=\frac{\delta}{c_KN_0},
\qquad\text{and set}\qquad
\varepsilon_\delta:=\frac1L+\frac{D+\log(1/\delta')}{LM}=\frac1L+\frac{D+\log(c_KN_0/\delta)}{LM}.
\]
\paragraph{Dependence on the Number of Blocks.} The number of blocks enters the estimates only with confidence $\delta'$, that is, through
\[
\lambda_N:=\frac{\log(c_KN_0)}{M^{1/3}},
\qquad
\Xi_N:=(1+\lambda_N)^{3/2}.
\]
Under $D\le c_0M$ and $\delta\ge\exp(-c_1M^{1/3})$, we have $\log(1/\delta')\le(c_1+\lambda_N)M^{1/3}$, and so
\begin{equation}\label{eqn:Leps_bound}
L\varepsilon_\delta\le1+c_0+c_1+\lambda_N\le C'(1+\lambda_N),
\qquad
\varepsilon_\delta\le C'\Big(\frac1L+\frac{D+\log(2N_0/\delta)}{LM}\Big),
\end{equation}
where $c_K$ has been absorbed into $C'$. Since $M\ge1$, also $1+\lambda_N\le C'(1+\log(2N_0))$.

\paragraph{Moments of the features.} The $\bar p$-norms of the features at the level $\delta'$ are controlled by
\[
\mathcal K_p:=\mathcal K^{D,M,p}_{\delta'},
\qquad
\mathcal K_{p,0}:=\mathcal K^{0,M,p}_{\delta'},
\qquad p\ge1,
\]
with $\mathcal K$ as in \eqref{eqn:mathcal K}. Under the same constraints, with $\zeta:=D/(LM^{2/3})$,
\begin{equation}\label{eqn:K_bounds}
\mathcal K_2+\mathcal K_{2,0}+\mathcal K_{6,0}\le C'(1+\lambda_N)^{1/2},
\qquad
\mathcal K_3^2\le C'(1+L\zeta+\lambda_N),
\qquad
L\varepsilon_\delta\le2\mathcal K_2^2\le C'L\varepsilon_\delta .
\end{equation}
Only $\mathcal K_3$ is unbounded, through $L\zeta$, under our shape constraints.

\paragraph{The comparison history.}
Lemma~\ref{lemma:prop_error_prelim} compares the features of layer $r$ along two histories with those along a reference history $\widetilde{\bm Z}_r$ which is independent of $(\Phi^r_0,\Psi^r_0)$, so that, conditionally on $\mathcal F_{-r}$,
the $M$ neurons of layer $r$ trained along it are i.i.d. The natural
choice for this reference history is the leave-layer-$r$-out history 
\[
\bm Z^{(-r)}_{r,k}:=\big(h^{(-r)}_{r-1,k},\,a^{(-r)}_{r,k}\big),\qquad 0\le k\le K
\]
It does not use the parameters of layer $r$, and so
is $\mathcal F_{-r}$-measurable. By Lemma~\ref{lemma:apriori} it is bounded by $C_B$ on $\mathcal E^{\mathrm{op}}_\delta$, but this event depends on the parameters of layer $r$, whereas Lemma~\ref{lemma:prop_error_prelim} requires a bound that holds surely.
Therefore, we truncate the history using the projection $\pi$ of each component onto $\{x:\|x\|_{\bar2}\le C_B\}$, so we set $\widetilde{\bm Z}_{r,k}:=\pi(\bm Z^{(-r)}_{r,k})$. Then $\widetilde{\bm Z}_r$ is $\mathcal F_{-r}$-measurable and bounded by $C_B$ surely. Since $\pi$ is non-expansive and $\bm z_r$ is bounded by $C_B$ on $\mathcal E^{\mathrm{op}}_\delta$, on this event
\[
\max_{k'\le k}\|\bm z_{r,k'}-\widetilde{\bm Z}_{r,k'}\|_{\bar2}
\le\Delta^{(-r)}_{r,\wedge k},
\]
the diagonal leave-one-layer-out error of \eqref{eqn:mathcal D}. 
\paragraph{The Common Event.}
All the probabilistic estimates of this proof hold on a single event $\mathcal E_\delta$. It is the intersection, over the $N_0$ blocks, of the following events, each applied at the level $\delta'$ and each holding simultaneously for all the training steps.
\begin{enumerate}
\item \emph{Initial operator norms.} The initial parameters of the layers of the block have bounded operator norms, on average over the block, both in the plain and in the mixed norms used in Section~\ref{sec:decoupling}.
\item \emph{Features along the comparison histories.} The features of each layer of the block, trained along its comparison history, have bounded moments on average over the block, by Lemma~\ref{lemma:p-norm_on_PQ}, averaged by Lemma~\ref{lemma:LogRemoval_mixed}.
\item \emph{Propagation.} The event of Lemma~\ref{lemma:prop_error_prelim} holds for the block, with the comparison histories, for each of the finitely many functions and exponents used below.
\end{enumerate}
Precisely, the first event is
\[
\overline\varpi_{I_i,16}^2\le C'L\varepsilon_\delta,
\qquad
C^{I_i}_{2,3}(\Phi_0)\vee C^{I_i}_{2,3}(\Psi_0)\le C'\Big(\sqrt{\mathfrak m}+\frac{\sqrt{D+\log(1/\delta')}}{M^{1/3}}\Big).
\]
The first bound follows from \eqref{eqn:single_layer_opnorm} and Lemma~\ref{lemma:LogRemoval_mixed}, with exponent $16$ over the block, and the second from Lemma~\ref{lemma:OpNormLayerAvg}. Each of the $c_KN_0$ events has probability at least $1-\delta'$, so $\mathbb P(\mathcal E_\delta)\ge1-\delta$.

\paragraph{No union bound over the removed layer.} The event $\mathcal E_\delta$ controls the initial operator norms simultaneously over all blocks, independently of the removed layer. Since $\iota_r\le C'(\varpi_2^r)^2$, Jensen's inequality and the bound on $\overline\varpi_{I_i,16}$ give, on $\mathcal E_\delta\cap \mathcal E^{\mathrm{op}}_\delta$,
\begin{equation}\label{eqn:impulse_avg}
\max_{0\le i<N_0}
\left(\frac1{\mathfrak m}\sum_{r\in I_i}\iota_r^8\right)^{1/8}
\le C'\max_{0\le i<N_0}\overline\varpi_{I_i,16}^2
\le C'L\varepsilon_\delta.
\end{equation}

\subsubsection{Comparing the Residuals along Two Histories}\label{sec:decoupling}\label{sec:LOLO_forward}\label{sec:LOLO_backward}
The leave-one-layer-out errors, and the propagation error of Section~\ref{sec:ProofMain}, are both differences between the residuals of the layers trained along the history of the network and along a second history. We bound such differences for a general second history. The network with a layer removed is the special case \eqref{eqn:LOLO_as_history}.

\paragraph{} Write $\bm z_{r,k}:=(h_{r-1,k},a_{r,k})$ for the history of the discrete model. Suppose that we are given
$\bm Z=(\bm Z_r)_{r=1}^L$, with $\bm Z_r=(\bm Z_{r,k})_{k=0}^K$ and
$\bm Z_{r,k}=(H^*_{r-1,k},A^*_{r,k})\in\mathbb R^{2D}$, such that
\[
\max_{1\le r\le L}\max_{0\le k\le K}\|\bm Z_{r,k}\|_{\bar2}\le C_B ,
\qquad
\|\bm Z_{r,k}\|_{\bar2}:=\|H^*_{r-1,k}\|_{\bar2}\vee\|A^*_{r,k}\|_{\bar2},
\]
that may depend arbitrarily on the parameters. A star denotes the quantities of layer $r$ trained, from the same initial parameters $\Phi^r_0,\Psi^r_0$, along $\bm Z_r$ in place of $\bm z_r$:
\[
\Phi^{*,r}_k:=\Phi^r_k(\bm Z_r),\qquad\Psi^{*,r}_k:=\Psi^r_k(\bm Z_r),\qquad p^{*,r}_k:=P^r_k(\bm Z_r),\qquad q^{*,r}_k:=Q^r_k(\bm Z_r).
\]
For either history, write
\[
\bm{f}^r_{k} :=  \bm{\rho}'(p^r_{k})\odot q^r_{k}, \qquad \bm{g}^{r}_{k} :=  \bm{\rho}'(p^r_{k})\odot \mathrm{cl}(q^r_{k}),
\]
and $\bm f^{*,r}_{k}=\bm f^r_k(\bm Z_r)$, $\bm g^{*,r}_{k}=\bm g^r_k(\bm Z_r)$ for the same expressions in $p^{*,r}_k,q^{*,r}_k$. The bounds \eqref{eqn:op_growth} and \eqref{eqn:feature_op} only use that the history along which layer $r$ is trained is bounded by $C_B$, so they give for every $r$ and every $j\le K$,
\begin{equation}\label{eqn:feature_op_general}
\|p^{*,r}_j\|_{\bar2}\vee\|q^{*,r}_j\|_{\bar2}\vee\|\bm\rho(p^{*,r}_j)\|_{\bar2}\vee\|\bm f^{*,r}_j\|_{\bar2}\le C'\varpi^r_2,
\qquad
\|\bm g^{*,r}_j\|_{\bar 2}\le\Lambda_\rho c .
\end{equation}

\paragraph{Decoupling errors.} For a set of layers $\mathcal I\subseteq\{1,\ldots,L\}$ and a step $k$ for brevity, define
\begin{align*}
\mathcal E^{H}_{\mathcal I,k}(\bm Z)
&:=\frac1{LM}\sum_{r\in\mathcal I}\Big(\Phi^r_{k}\,\bm\rho(p^r_{k})-\Phi^{*,r}_{k}\,\bm\rho(p^{*,r}_{k})\Big),
&
\mathcal E^{A}_{\mathcal I,k}(\bm Z)
&:=\frac1{LM}\sum_{r\in\mathcal I}\Big(\Psi^r_{k}\,\bm f^r_{k}-\Psi^{*,r}_k\,\bm f^{*,r}_{k}\Big),
\end{align*}
and write $\mathcal E^{H}_{l,k}(\bm Z):=\mathcal E^{H}_{[1,l],k}(\bm Z)$ and $\mathcal E^{A}_{l,k}(\bm Z):=\mathcal E^{A}_{[l+1,L],k}(\bm Z)$.
% Expanding the numerical schemes, we find 
% \[
% \mathcal E^{H}_{l,k}(\bm Z)=h_{l,j}-h_{0,j}-\frac1{LM}\sum_{r=1}^{l}\Phi^{*,r}_{j}\,\bm\rho(p^{*,r}_{j}),
% \qquad
% \mathcal E^{A}_{l,k}(\bm Z)=a_{l,j}-a_{L,j}-\frac1{LM}\sum_{r=l+1}^{L}\Psi^{*,r}_{j}\,\bm f^{*,r}_{j}.
% \]

\paragraph{Distances.} The distances between the two histories and hidden states at layer $r$ are denoted
\[
\Delta_{r,k}(\bm Z):=\max_{0\le k'\le k}\|\bm Z_{r,k'}-\bm z_{r,k'}\|_{\bar2},
\qquad
\Delta^H_{r,k}(\bm Z):=\max_{0\le k'\le k}\|H^*_{r-1,k'}-h_{r-1,k'}\|_{\bar2}.
\]
For a set $\mathcal I$ contained in a block, define the $\ell^8$-distances, normalised by the block length $\mathfrak m$,
\[
d^{\mathcal I}_k(\bm Z):=\Big(\frac1{\mathfrak m}\sum_{r\in\mathcal I}
\Delta_{r,k}(\bm Z)^8\Big)^{1/8},
\qquad
d^{H,\mathcal I}_k(\bm Z):=\Big(\frac1{\mathfrak m}\sum_{r\in\mathcal I}
\Delta^H_{r,k}(\bm Z)^8\Big)^{1/8},
\]
with $d^{\mathcal I}_{-1}(\bm Z):=0$. For the blocks, write
\[
d^{(j)}_k(\bm Z):=d^{I_j}_k(\bm Z),\qquad d_k(\bm Z):=\max_{0\le j<N_0}d^{(j)}_k(\bm Z),
\]
and similarly use superscript $H$ to denote the same quantities restricted to the hidden state errors.

\paragraph{The diagonal.} The adjoint estimate below involves the distance between $\bm z_r$ and the comparison history $\widetilde{\bm Z}_r$, that is, the \emph{diagonal} leave-one-layer-out error. For $p\ge1$ and $l\le l'$ in a block, define
\begin{align}
\label{eqn:mathcal D}
\mathcal D_{k,p}^{[l,l']}
:=
\left(
\frac1{\mathfrak m}
\sum_{r=l}^{l'}
\big(\Delta^{(-r)}_{r,\wedge k}\big)^p
\right)^{\frac1p},
\quad
\Delta^{(-r)}_{r,\wedge k}=\max_{k'\le k}
\|h_{r-1,k'}-h_{r-1,k'}^{(-r)}\|_{\bar2}
\vee
\|a_{r,k'}-a_{r,k'}^{(-r)}\|_{\bar2},
\end{align}
with $\mathcal D^{\emptyset}_{k,p}:=0$. The normalisation is fixed at $1/\mathfrak m$, so $\mathcal D_{k,p}^{[l,l']}$ is non-increasing in $l$ and non-decreasing in $l'$. Since the diagonal error is at most the error over all layers with layer $r$ removed, for every block $I_j$ and $l\in I_j$,
\[
\mathcal D_{k,p}^{[l,l_{j+1}]}
\le
\left(
\frac1{\mathfrak m}
\sum_{r\in I_j}
\|\Delta^{(-r)}_{\wedge k}\|_\infty^p
\right)^{1/p}
\]

\begin{proposition}[Residual comparison]\label{prop:comparison}
Suppose the assumptions of Lemma~\ref{lemma:apriori} hold, and assume \[L\ge N_0, \qquad D\le c_0M, \qquad \delta\ge\exp(-c_1(M^{1/3}\wedge D)).\] Then, there exist constants $C_{\mathrm{fwd}}$ and $\mathfrak C$, depending only on the constants $C'$ of Section~\ref{sec:LOLO_notation}, such that, on $\mathcal E_\delta\cap\mathcal E^{\mathrm{op}}_\delta$, simultaneously for every history $\bm Z$, every $0\le k\le K$, every block $I_i$ and every interval $\mathcal I=[l,l']\subseteq I_i$,
\begin{align}
\|\mathcal E^{H}_{\mathcal I,k}(\bm Z)\|_{\bar2}
&\le C_{\mathrm{fwd}}\,\hat{\mathfrak m}\,
\Big(d^{H,\mathcal I}_k(\bm Z)+\mathbbm{1}_{\{k>0\}}d^{\mathcal I}_{k-1}(\bm Z)\Big),
&\hat{\mathfrak m}&:=\varepsilon\,(1+\lambda_N),\label{eqn:comparison_H}\\
\|\mathcal E^{A}_{\mathcal I,k}(\bm Z)\|_{\bar2}
&\le \mathcal U^{\mathcal I}_k\,d^{\mathcal I}_k(\bm Z),\label{eqn:comparison_A}
\end{align}
where
\begin{equation}\label{eqn:U_bound}
\mathcal U^{\mathcal I}_k:=\mathfrak C\,(\varepsilon+\sqrt{\varepsilon\zeta})\big(1+L\zeta\,\mathcal D^{\mathcal I}_{k,8}\big)\,\Xi_N,
\end{equation}
with $\varepsilon=\mathfrak m/L$, $\zeta=D/(LM^{2/3})$, and $\lambda_N$ and $\Xi_N$ as in Section~\ref{sec:LOLO_notation}. The number of blocks enters both coefficients only through $\lambda_N$, which accounts for the events holding simultaneously over all the blocks.
\end{proposition}

\begin{proof}
Fix a block $I_i$ and an interval $\mathcal I\subseteq I_i$, and write $\Delta_r:=\Delta_{r,k}(\bm Z)$. We use three tools.
\begin{itemize}
    \item  \emph{Lemma~\ref{lemma:prop_error_prelim}}, for the block $I_i$, so that $n=|I_i|\le\mathfrak m$, with $(\bm Z^0,\bm Z^1)=(\bm z,\bm Z)$, the subset $\mathcal J=\mathcal I$ and the comparison histories $\widetilde{\bm Z}_r$. Its events are part of $\mathcal E_\delta$, and on them it holds for all pairs of histories bounded by $C_B$. On $\mathcal E^{\mathrm{op}}_\delta$, the difference $\widetilde\Delta$ between $\bm Z^0=\bm z$ and $\widetilde{\bm Z}$ is the diagonal error of \eqref{eqn:mathcal D}, so that $\|(\widetilde\Delta_r)_{r\in\mathcal I}\|_{8}=\mathfrak m^{1/8}\mathcal D^{\mathcal I}_{k,8}$.

\item \emph{H\"older's inequality over the layers.} For $1\le s\le 8$,
\begin{equation}\label{eqn:Holder_block}
\|(\Delta_r)_{r\in\mathcal I}\|_{s}\le|\mathcal I|^{\frac1s-\frac18}\Big(\sum_{r\in\mathcal I}\Delta_r^8\Big)^{1/8}\le\mathfrak m^{\frac1s}\,d^{\mathcal I}_k(\bm Z).
\end{equation}
Hence the factors $\frac{\sqrt n}{L}n^{-1/8}\|\Delta\|_{8}$ and $\frac{\sqrt n}{L}n^{-1/4}\|\Delta\|_{4}$ of Lemma~\ref{lemma:prop_error_prelim} are at most $\frac{\sqrt{\mathfrak m}}{L}d^{\mathcal I}_k(\bm Z)$, and
\[
\frac{\sqrt n}{L}\,n^{-1/8}\|\widetilde\Delta\|_{\ell^8}\,n^{-1/8}\|\Delta\|_{\ell^8}\le\frac{\sqrt{\mathfrak m}}L\,\mathcal D^{\mathcal I}_{k,8}\,d^{\mathcal I}_k(\bm Z).
\]
\item \emph{ Weighted sums.} H\"older's inequality with exponents $(\frac87,8)$, and Jensen's inequality for the weights $\frac1{\mathfrak m}$ on $I_i$, give
\begin{align}
\sum_{r\in\mathcal I}(\varpi^r_2)^2\Delta_r
&\le\Big(\sum_{r\in I_i}(\varpi^r_2)^{16/7}\Big)^{\frac78}\Big(\sum_{r\in\mathcal I}\Delta_r^8\Big)^{\frac{1}{8}}\le\mathfrak m\overline\varpi_{I_i,16}^2d^{\mathcal I}_k(\bm Z)
\le C'\mathfrak m\,L\varepsilon_\delta\,d^{\mathcal I}_k(\bm Z),\label{eqn:Holder_weighted}
\end{align}
since $\overline\varpi_{I_i,16}^2\le C'L\varepsilon_\delta$ on $\mathcal E_\delta$. The same holds with $\Delta_{r,k}$ replaced by $\Delta^H_{r,k}$ or by $\Delta_{r,k-1}$, and $d^{\mathcal I}_k$ by $d^{H,\mathcal I}_k$ or $d^{\mathcal I}_{k-1}$.
\end{itemize}
\paragraph{The hidden states.} For each layer $r\in\mathcal I$, add and subtract $\Phi^r_k\bm\rho(p^{*,r}_k)$, and expand $\Phi^r_k-\Phi^{*,r}_k$ by the updates of the two layers, to obtain
\begin{align*}
\frac1M\Big(\Phi^r_k\bm\rho(p^r_k)-\Phi^{*,r}_k\bm\rho(p^{*,r}_k)\Big)
={}&\frac1M\Phi^r_k\big(\bm\rho(p^r_k)-\bm\rho(p^{*,r}_k)\big)
-\eta\sum_{k'<k}a_{r,k'}\big\langle\bm\rho(p^r_{k'})-\bm\rho(p^{*,r}_{k'}),\bm\rho(p^{*,r}_k)\big\rangle_{\bar2}\\
&+\eta\sum_{k'<k}\big(A^*_{r,k'}-a_{r,k'}\big)\big\langle\bm\rho(p^{*,r}_{k'}),\bm\rho(p^{*,r}_k)\big\rangle_{\bar2}.
\end{align*}
Since $\rho$ is $\Lambda_\rho$-Lipschitz (Assumption \ref{assm:rho}), we can apply Lemma \ref{lemma:apriori} and \ref{lemma:DeltaP_prelim} to bound the first term by 
\[
\left\|\frac1M\Phi^r_k\big(\bm\rho(p^r_k)-\bm\rho(p^{*,r}_k)\big)\right\|_{\bar 2}\le C'(\varpi^r_2)^2(\Delta^H_{r,k}(\bm Z)+\mathbbm 1_{\{k>0\}}\Delta_{r,k-1}(\bm Z)).
\]
 On $\mathcal E^\mathrm{op}_\delta$, $\|a_{r,k'}\|_{\bar2}\le C_B$ and the $\bar{2}$-norm of $\bm{\rho}(p^*_{r,k})$ can be bounded according to \eqref{eqn:feature_op_general}. Therefore, the second is at most $ C'(\varpi^r_2)^2\Delta_{r,k-1}(\bm Z)$. By the Cauchy--Schwarz inequality, \eqref{eqn:feature_op_general} and linear growth of $\rho$, the third is at most $C'(\varpi^r_2)^2\Delta_{r,k-1}$. Since $\eta K\le C'$ and the sums over $k'$ are empty if $k=0$, summing over $r\in\mathcal I$ gives
\[
\|\mathcal E^H_{\mathcal I,k}(\bm Z)\|_{\bar2}\le\frac{C'}L\sum_{r\in\mathcal I}(\varpi^r_2)^2\Big(\Delta^H_{r,k}(\bm Z)+\mathbbm 1_{\{k>0\}}\Delta_{r,k-1}(\bm Z)\Big).
\]
By \eqref{eqn:Holder_weighted}, with $\Delta^H_{r,k}$ and with $\Delta_{r,k-1}$, which produces an $L\varepsilon_\delta$ term that is simplified by \eqref{eqn:Leps_bound}, we obtain
\[
\|\mathcal E^H_{\mathcal I,k}(\bm Z)\|_{\bar2}\le C'\varepsilon\,L\varepsilon_\delta\Big(d^{H,\mathcal I}_k(\bm Z)+\mathbbm 1_{\{k>0\}}d^{\mathcal I}_{k-1}(\bm Z)\Big)
\le C'\varepsilon(1+\lambda_N)\Big(d^{H,\mathcal I}_k(\bm Z)+\mathbbm 1_{\{k>0\}}d^{\mathcal I}_{k-1}(\bm Z)\Big),
\]
which is \eqref{eqn:comparison_H}.

\paragraph{The adjoint variables.} 
For ease of notation, define
\[
\Delta p^r_{k'}:=p^r_{k'}-p^{*,r}_{k'},\qquad\Delta q^r_{k'}:=q^r_{k'}-q^{*,r}_{k'},
\]
whose $\bar2$-norms are bounded by Lemma~\ref{lemma:DeltaP_prelim}. Expanding $\Psi^r_j=\Psi^r_0-\eta\sum_{k'<j}h_{r-1,k'}(\bm g^r_{k'})^\top$, and similarly along $\bm Z$, gives
\begin{equation}\label{eqn:Decomp}
\mathcal E^{A}_{\mathcal I,k}(\bm Z)=\mathcal E^1+\mathcal E^2+\mathcal E^3+\mathcal E^4,
\end{equation}
where
\begin{align}
\mathcal E^1&=\frac{1}{LM}\sum_{r\in\mathcal I}\Psi^{r}_{0}\left(\bm{f}^r_{j}-\bm{f}^{*,r}_{j}\right),\label{eqn:a-a^*,1}\\\mathcal E^2&= -\frac{\eta}{LM} \sum_{0\le k' < j}\sum_{r\in\mathcal I}h_{r-1,k'}\left(\bm{g}^{r}_{k'}\right)^\top\left(\bm{f}^{r}_{j}-\bm{f}^{*,r}_{j}
   \right),\label{eqn:a-a^*,2}\\\mathcal E^3&= -\frac{\eta}{LM} \sum_{0\le k' < j}\sum_{r\in\mathcal I}h_{r-1,k'}\left(\bm{f}^{*,r}_{j}\right)^\top\left(\bm{g}^{r}_{k'}-\bm{g}^{*,r}_{k'}
   \right),\label{eqn:a-a^*,3}\\\mathcal E^4&=  \frac{\eta}{L} \sum_{0\le k' < j}\sum_{r\in\mathcal I}\left(H^*_{r-1,k'}-h_{r-1,k'}\right) \left\langle \bm{f}^{*,r}_{j},\bm{g}^{*,r}_{k'} \right\rangle_{\bar 2} . \label{eqn:a-a^*,4}
\end{align}
This follows by adding and subtracting $\Psi^r_j\bm f^{*,r}_j$, and splitting the difference of the two updates of the adjoint parameters as
\[
h_{r-1,k'}(\bm g^r_{k'})^\top-H^*_{r-1,k'}(\bm g^{*,r}_{k'})^\top=h_{r-1,k'}\big(\bm g^r_{k'}-\bm g^{*,r}_{k'}\big)^\top+\big(h_{r-1,k'}-H^*_{r-1,k'}\big)(\bm g^{*,r}_{k'})^\top .
\]
The terms $\mathcal E^2$, $\mathcal E^3$ and $\mathcal E^4$ are bounded layer by layer. Fix $r\in\mathcal I$ and $k'<j$, and note that $\|h_{r-1,k'}\|_{\bar2}\le C_B$.
For the term $\mathcal E^2$, decompose the difference in $\bm{f}$ via \[\bm f^r_j-\bm f^{*,r}_j=\bm\rho'(p^r_j)\odot\Delta q^r_j+q^{*,r}_j\odot\big(\bm\rho'(p^r_j)-\bm\rho'(p^{*,r}_j)\big).\]
Since the entries of $\bm g^r_{k'}$ are bounded by $\Lambda_\rho c$,
\[
\big|\big\langle\bm g^r_{k'},\bm f^r_j-\bm f^{*,r}_j\big\rangle_{\bar2}\big|
\le\Lambda_\rho^2c\Big(\|\Delta q^r_j\|_{\bar2}+\|q^{*,r}_j\|_{\bar2}\|\Delta p^r_j\|_{\bar2}\Big)
\le C'(\varpi^r_2)^2\Delta_{r,j},
\]
having applied \eqref{eqn:feature_op_general} and Lemma \ref{lemma:DeltaP_prelim}. For $\mathcal E^3$, each neuron satisfies \\$|g^{rm}_{k'}-g^{*,rm}_{k'}|\le\Lambda_\rho c\,(|\Delta p^{rm}_{k'}|+|\Delta q^{rm}_{k'}|)$, and thus by Cauchy--Schwarz
\[
\big|\big\langle\bm f^{*,r}_j,\bm g^r_{k'}-\bm g^{*,r}_{k'}\big\rangle_{\bar2}\big|
\le\|\bm f^{*,r}_j\|_{\bar2}\,\Lambda_\rho c\big(\|\Delta p^r_{k'}\|_{\bar2}+\|\Delta q^r_{k'}\|_{\bar2}\big)
\le C'(\varpi^r_2)^2\Delta_{r,k'}.
\]
Finally, the difference in hidden states in $\mathcal E^4$ is bounded by $\Delta_{r,j-1}$, while the linear growth of $\rho$ and boundedness of $\mathrm{cl}$ yield
\[|\langle\bm f^{*,r}_j,\bm g^{*,r}_{k'}\rangle_{\bar2}|\le\Lambda_\rho c\,C'\varpi^r_2.\]
 Since $\eta K\le C'$, summing the above estimates over $k'<j$ and $r\in\mathcal I$, and using \eqref{eqn:Holder_weighted} and \eqref{eqn:Leps_bound},
\begin{equation}\label{F:E234}
\|\mathcal E^2\|_{\bar2}+\|\mathcal E^3\|_{\bar2}+\|\mathcal E^4\|_{\bar2}
\le\frac{C'}L\sum_{r\in\mathcal I}(\varpi^r_2)^2\Delta_{r,j}(\bm Z)
\le C'\varepsilon(1+\lambda_N)\,d^{\mathcal I}_j(\bm Z).
\end{equation}
For $\mathcal E^1$, we apply Lemma \ref{lemma:prop_error_prelim} with $f(p,q)=\rho'(p)q$, $\theta^{rm}=\psi^{rm}_0$ and the exponents $(3/2,3)$, together with \eqref{eqn:Holder_block} to obtain
\[
\|\mathcal E^1\|_{\bar2}\le C'\frac{\sqrt{\mathfrak m}}L\Big[C^{I_i}_{2,3}(\Psi_0)\big(\mathcal K_3^2\,\mathcal D^{\mathcal I}_{j,8}+\mathcal K_2\mathcal K_{6,0}\big)+C^{I_i}_{2,2}(\Psi_0)\,\mathcal K_2\Big]d^{\mathcal I}_j(\bm Z).
\]
On $\mathcal E_\delta$, by Lemma \ref{lemma:OpNormLayerAvg}
\[
\frac{\sqrt{\mathfrak m}}L\,C^{I_i}_{2,3}(\Psi_0)\le C'\sqrt\varepsilon\Big(\sqrt\varepsilon+\frac{\sqrt{D+\log(1/\delta')}}{\sqrt L\,M^{1/3}}\Big)\le C'(\varepsilon+\sqrt{\varepsilon\zeta})(1+\lambda_N)^{1/2}.
\]
The only potentially unbounded term in \eqref{eqn:K_bounds} is $L\zeta$. Therefore, we retain $L\zeta\mathcal D^{\mathcal I}_{k,8}$ and absorb the remaining terms into the constant, using $\mathcal D^{\mathcal I}_{k,8}\le 2C_B$, and apply our shape and failure probability constraints \eqref{eqn:S&FPconstraints} to obtain
\[
\mathcal K_3^2\,\mathcal D^{\mathcal I}_{k,8}+\mathcal K_2\mathcal K_{6,0}\le C'(1+\lambda_N)\big(1+L\zeta\,\mathcal D^{\mathcal I}_{k,8}\big),
\qquad
\mathcal K_2^2\le C'(1+\lambda_N),
\]
where $\lambda_N$ is defined in Section~\ref{sec:LOLO_notation}. Hence, combining the above estimates with \\$C^{I_i}_{2,2}(\Psi_0)\le C'\sqrt{\mathfrak m}\,\mathcal K_2$, which follows from the definition \eqref{eqn:C_{p,q}^{l}}, Jensen's inequality and \eqref{eqn:K_bounds} so that $\overline\varpi_{I_i,2}^2\le C'L\varepsilon_\delta$ on $\mathcal E_\delta$, yields
\[
\|\mathcal E^1\|_{\bar{2}} \le C'\left[ (\varepsilon+\sqrt{\varepsilon \zeta})(1+L\zeta \mathcal D_{k,8}^{\mathcal I})\Xi_N\right]d^{\mathcal I}_{k}(\bm{Z}).
\]
The desired result \eqref{eqn:comparison_A} is obtained by adding the above to \eqref{F:E234} and \\applying $\varepsilon(1+\lambda_N)\le(\varepsilon+\sqrt{\varepsilon\zeta})\,\Xi_N$.
\end{proof}

\paragraph{Leave-one-layer-out as a special case.} Fix a removed layer $l^*$, and write $h^*,a^*$ for the hidden states and adjoint variables of the network with layer $l^*$ removed. Consider the history
\begin{equation}
\label{eqn:LOLO_as_history}
\bm Z^{(l^*)}_{r,k}:=\begin{cases}\big(h^{*}_{r-1,k},\,a^{*}_{r,k}\big),&r\ne l^*,\\[2pt] \bm z_{l^*,k},&r=l^*,\end{cases}
\end{equation}
which is bounded by $C_B$ on $\mathcal E^{\mathrm{op}}_\delta$. For $r\ne l^*$, the starred quantities of layer $r$ along $\bm Z^{(l^*)}$ are the parameters and features of layer $r$ in the network with layer $l^*$ removed, and $\Delta_{r,k}(\bm Z^{(l^*)})$, defined in the distances section above, is less than or equal to $\max_{k'\le k}\Delta^{(-l^*)}_{r,k'}$. Since the $\bar 8$-norm of layerwise errors is bounded above by their maximum, for an interval $\mathcal I$ in a block,
\[
d^{\mathcal I}_k(\bm Z^{(l^*)})\le\Delta^{(-l^*)}_{\mathcal I,\wedge k},
\qquad
d^{H,\mathcal I}_k(\bm Z^{(l^*)})\le\Delta^{(-l^*),H}_{\mathcal I,\wedge k}.
\]

\emph{The forward recursion.} Let $l\in I_i$ and $j:=k$. Subtracting the forward passes of the two networks from $l_i$ to $l$,
\[
h_{l,j}-h^*_{l,j}=h_{l_i,j}-h^*_{l_i,j}+\mathcal E^H_{(l_i,l],k}(\bm Z^{(l^*)})+\mathbbm{1}_{\{l_i <  l^* \le l\}}\frac{1}{LM} \Phi^{l^*}_{j} \,\bm{\rho}(p^{l^*}_{j}),
\]
where the last term is the impulse of the removed layer, of norm at most $\iota^H_{l^*}/L$. By \eqref{eqn:comparison_H}, on $\mathcal E_\delta$, for every $l^*$, $k\le K$, $i$ and $l\in I_i$,
\[
\|h_{l,k}-h^*_{l,k}\|_{\bar2}\le\|h_{l_i,k}-h^*_{l_i,k}\|_{\bar2}+\mathbbm 1_{\{l_i<l^*\le l\}}\frac{\iota^H_{l^*}}L+C_{\mathrm{fwd}}\hat{\mathfrak m}\Big(\Delta^{(-l^*),H}_{(l_i,l],\wedge k}+\mathbbm 1_{\{k>0\}}\|\Delta^{(-l^*)}_{\wedge(k-1)}\|_\infty\Big).
\]
The final two terms on the RHS are non-decreasing in $k$, so the same bound holds at every step $k'\le k$. Let \[e_i:=\max_{0\le l\le l_i}\max_{k'\le k}\|h_{l,k'}-h^*_{l,k'}\|_{\bar2},\quad \text{ so that} \quad e_0\le\eta_{\mathrm{in}} K|x|^2\,\mathbbm 1_{\{k>0\}}\|\Delta^{(-l^*)}_{\wedge(k-1)}\|_\infty,\] by \eqref{eqn:boundary_input}, and $\Delta^{(-l^*),H}_{(l_i,l],\wedge k}\le e_{i+1}$.
Taking the maximum over $l\in I_i$ and $k'\le k$ gives
\[
e_{i+1}\le e_i+\mathbbm 1_{\{l^*\in I_i\}}\frac{\iota^H_{l^*}}L+C_{\mathrm{fwd}}\hat{\mathfrak m}\Big(e_{i+1}+\mathbbm 1_{\{k>0\}}\|\Delta^{(-l^*)}_{\wedge(k-1)}\|_\infty\Big).
\]
By the choice \eqref{eqn:N0_choice} of $N_0$ and since $\varepsilon \le 2/N_0$, $C_{\mathrm{fwd}}\hat{\mathfrak m}\le\frac14$, and $e_{i+1}\le 2C_B$ is finite, so that 
\[
e_{i+1}\le\frac43\Big(e_i+\mathbbm 1_{\{l^*\in I_i\}}\frac{\iota^H_{l^*}}L\Big)+\frac13\mathbbm 1_{\{k>0\}}\|\Delta^{(-l^*)}_{\wedge(k-1)}\|_\infty .
\]
Iterating over the $N_0$ blocks, the input error $e_0$ is multiplied by at most $(4/3)^{N_0}$. The impulse occurs in only one block and is similarly multiplied by the same geometric growth factor. Since $e_{N_0}$ bounds all the hidden states $h_0,\ldots,h_L$, we obtain, on $\mathcal E_\delta$, simultaneously for every removed layer $l^*$ and every $k\le K$,
\begin{align}\label{eqn:HiddenState_LOO_Recursion}
    \|\Delta^{(-l^*),H}_{\wedge k}\|_{\infty}&\le \left(\frac 43\right)^{N_0}\left[ \big(1+\eta_{\mathrm{in}} K|x|^2\big)\mathbbm{1}_{\{k>0\}}\| \Delta^{(-l^*)}_{\wedge (k-1)}\|_{\infty} + \frac{\iota^H_{l^*}}{L}\right].
\end{align}

\emph{The backward recursion.} Let $l\in I_i$. Subtracting the backward passes of the two networks from $l_{i+1}$ down to $l-1$,
\[
a_{l-1,j}-a^*_{l-1,j}=a_{l_{i+1},j}-a^*_{l_{i+1},j}+\mathcal E^A_{[l,l_{i+1}],j}(\bm Z^{(l^*)})+\mathbbm{1}_{\{l\le l^*\le l_{i+1}\}}\frac1{LM}\Psi^{l^*}_{j}\bm f^{l^*}_j.
\]
By \eqref{eqn:comparison_A}, and since the resulting bound is non-decreasing in $k$ and non-increasing in $l$, on $\mathcal E_\delta$, simultaneously for every removed layer $l^*$, every $k\le K$, every block $I_i$ and every $l\in I_i$,
\begin{align}
\label{eqn:backward_recurssion_LOO}
\Delta^{(-l^*),A}_{[l-1,l_{i+1}],\wedge k}
\le{}&
\Delta^{(-l^*),A}_{l_{i+1},\wedge k}+
\mathbbm{1}_{\{l\le l^*\le l_{i+1}\}}\frac{\iota^A_{l^*}}{L}+
\mathcal U_{k}^{[l,l_{i+1}]}\,
\Delta^{(-l^*)}_{[l,l_{i+1}],\wedge k}.
\end{align}
Here $\Delta^{(-l^*),A}_{l_{i+1},\wedge k}$ is the error of the adjoint $a_{l_{i+1}}$ at the upper boundary of the block, and the left-hand side covers the adjoints $a_{l-1},\ldots,a_{l_{i+1}}$. The estimate on the diagonal that makes $\mathcal U_k^{[l,l_{i+1}]}\le\frac12$ is established by the following lemma.

\subsubsection{Bounding the Diagonal}\label{sec:LOLO_diag}
\begin{lemma}[Within-block LOLO stability]
\label{lemma:Inductive_Step}
Suppose the hypotheses of Proposition~\ref{prop:comparison} hold, and let
$\varepsilon=\mathfrak m/L$, $\zeta=D/(LM^{2/3})$ and $\Xi_N$ be as
in \eqref{eqn:U_bound}, with $\mathfrak C$ the constant there. Fix
a block $I_i=\{l_i+1,\ldots,l_{i+1}\}$ of the partition
and a training step $0\le k\le K$. Suppose that, for some
$\hat C>0$,
\begin{align}
\label{eqn:Inductive_hypothesis}
&\left(
\frac1{\mathfrak m}
\sum_{r\in I_i}
|\Delta^{(-r),A}_{l_{i+1},\wedge k}|^8
\right)^{1/8}
\vee
\left(
\frac1{\mathfrak m}
\sum_{r\in I_i}
\|\Delta^{(-r),H}_{I_i,\wedge k}\|_\infty^8
\right)^{1/8}
\le\frac{\hat{C}}{L},
\end{align}
and that
\begin{equation}
\label{eqn:block_smallness}
\mathfrak C
(\varepsilon+\sqrt{\varepsilon\zeta})
(1+4\zeta \hat C)\Xi_N
\le\frac12 .
\end{equation}
Then, on $\mathcal E_\delta\cap\mathcal E^{\mathrm{op}}_\delta$, for every $r,l\in I_i$,
\begin{align}
\label{eqn:within_block_adjoint}
\mathcal D_{k,8}^{[l,l_{i+1}]}
\le\frac{4\hat C}{L},\qquad \|\Delta^{(-r),A}_{[r,l_{i+1}],\wedge k}
\|_\infty
\le
2|\Delta^{(-r),A}_{l_{i+1},\wedge k}|
+
\|\Delta^{(-r),H}_{I_i,\wedge k}\|_\infty.
\end{align}
\end{lemma}

\begin{proof}
For fixed $r\in I_i$ and $r<l\le l_{i+1}$, the impulse term
in \eqref{eqn:backward_recurssion_LOO} vanishes, and
$\mathcal D_{k,8}^{[l,l_{i+1}]}\le\mathcal D_{k,8}^{[r+1,l_{i+1}]}$.
Hence \eqref{eqn:backward_recurssion_LOO}, together with
\eqref{eqn:U_bound}, gives
\begin{align*}
\|\Delta^{(-r),A}_{[l-1,l_{i+1}],\wedge k}
\|_\infty
\le{}&
|\Delta^{(-r),A}_{l_{i+1},\wedge k}|+\alpha_r\left(
\|\Delta^{(-r),A}_{[l,l_{i+1}],\wedge k}\|_\infty
+\|\Delta^{(-r),H}_{I_i,\wedge k}\|_\infty
\right),
\end{align*}
where
\[
\alpha_r
=
\mathfrak C(\varepsilon+\sqrt{\varepsilon\zeta})
\left(
1+L\zeta\mathcal D_{k,8}^{[r+1,l_{i+1}]}
\right)\Xi_N.
\]

Since the normalisation in \eqref{eqn:mathcal D} is
fixed at $1/\mathfrak m$, independently of the range of
summation, $\mathcal D_{k,8}^{[l,l_{i+1}]}$ is non-increasing
in $l$. As a result, $\alpha_r$ is non-decreasing as $r$
decreases. We prove the diagonal estimate by backward induction over $l$, from $l=l_{i+1}$ down to $l=l_i+1$. Suppose that, for some $l_i<l\le l_{i+1}$,
\[
\mathcal D_{k,8}^{[l+1,l_{i+1}]}\le\frac{4\hat C}{L},
\]
which holds initially at $l=l_{i+1}$ by the convention $\mathcal D_{k,8}^{\emptyset}=0$. Then, by assumption of the Lemma
\[
\alpha_l
\le
\mathfrak C(\varepsilon+\sqrt{\varepsilon\zeta})
(1+4\zeta\hat  C)\Xi_N
\le\frac12,
\]
and so, by the monotonicity above, $\alpha_r\le\frac12$
for every $l\le r\le l_{i+1}$.
As a result, the geometric sums
\[
T_n^r:=\sum_{s=0}^n\alpha_r^s
\]
satisfy $T_n^r\le T_n^l\le2$ for every $n\ge0$ and
$l\le r\le l_{i+1}$.
For each such $r$, iterating the backward recursion
from $l_{i+1}$ down to $r$ gives
\begin{align}\label{eqn:Intermediate}
\|\Delta^{(-r),A}_{[r,l_{i+1}],\wedge k}
\|_\infty
\le{}&
T_{l_{i+1}-r}^r
|\Delta^{(-r),A}_{l_{i+1},\wedge k}|+(T_{l_{i+1}-r}^r-1)
\|\Delta^{(-r),H}_{I_i,\wedge k}\|_\infty\notag\\
\le{}&
2|\Delta^{(-r),A}_{l_{i+1},\wedge k}|+
\|\Delta^{(-r),H}_{I_i,\wedge k}\|_\infty.
\end{align}
Taking the normalised $\ell^8$-average over
$r\in[l,l_{i+1}]$, Minkowski's inequality and
\eqref{eqn:Inductive_hypothesis} yield
\[
\mathcal D_{k,8}^{[l,l_{i+1}]}
\le
\frac{2\hat C}{L}+\frac{\hat C}{L}
=\frac{3\hat C}{L}
\le\frac{4\hat C}{L}.
\]
This closes the backward induction, and thus proves
\eqref{eqn:within_block_adjoint}, where the second claim is \eqref{eqn:Intermediate}.
\end{proof}

\begin{corollary}\label{cor:Inductive_Step}
Suppose the hypotheses of
Lemma~\ref{lemma:Inductive_Step} hold, for the block $I_i$ of the partition of Section~\ref{sec:LOLO_notation}.
Then, on $\mathcal E_\delta\cap\mathcal E^{\mathrm{op}}_\delta$, for every $1\le l^*\le L$,
\begin{align*}
\|\Delta^{(-l^*),A}_{[l_i,l_{i+1}],\wedge k}
\|_\infty
\le {}&
2|\Delta^{(-l^*),A}_{l_{i+1},\wedge k}|
+
\|\Delta^{(-l^*),H}_{[l_i+1,l_{i+1}],\wedge k}
\|_\infty
+2\cdot\mathbbm{1}_{\{l_i<l^*\le l_{i+1}\}}\frac{\iota^A_{l^*}}{L}.
\end{align*}
\end{corollary}

\begin{proof}
For notational ease write
\[
H_*=
\|\Delta^{(-l^*),H}_{[l_i+1,l_{i+1}],\wedge k}
\|_\infty,
\qquad
A_*=\|\Delta^{(-l^*),A}_{[l_i,l_{i+1}],\wedge k}\|_\infty,
\]
where $A_*\le 2C_B$ is finite. By
Lemma~\ref{lemma:Inductive_Step},
$\mathcal D_{k,8}^{[l_i+1,l_{i+1}]}\le 4\hat CL^{-1}$, and since
$\mathcal D_{k,8}^{[l,l_{i+1}]}$ is non-increasing in $l$,
\eqref{eqn:U_bound} and \eqref{eqn:block_smallness} give
$\mathcal U_k^{[l,l_{i+1}]}\le\frac12$ for every
$l_i<l\le l_{i+1}$. Furthermore, the impulse is only felt once during the backward recursion. Therefore, by bounding
$\|\Delta^{(-l^*)}_{[l,l_{i+1}],\wedge k}\|_\infty\le A_*+H_*$, the recursion \eqref{eqn:backward_recurssion_LOO} for every $l_i<l\le l_{i+1}$ becomes
\[
|\Delta^{(-l^*),A}_{l-1,\wedge k}|
\le
|\Delta^{(-l^*),A}_{l_{i+1},\wedge k}|+\mathbbm{1}_{\{l_i<l^*\le l_{i+1}\}}\frac{\iota^A_{l^*}}{L}
+\frac12\left(A_*+H_*\right).
\]
Taking the
maximum over the layers of the block and rearranging, which is
possible since $A_*$ is finite, gives
\[
A_*\le 2|\Delta^{(-l^*),A}_{l_{i+1},\wedge k}|+H_*+2\cdot\mathbbm{1}_{\{l_i<l^*\le l_{i+1}\}}
\frac{\iota^A_{l^*}}{L}.
\]
\end{proof}

\subsubsection{Induction over the Blocks and Training Steps}\label{sec:LOLO_induction}
\paragraph{Choice of the number of blocks.} By \eqref{eqn:U_bound}, the stability coefficient in Lemma~\ref{lemma:Inductive_Step} is at most $\mathfrak C(\varepsilon+\sqrt{\varepsilon\zeta})(1+4\zeta \hat C)\Xi_N$, where $\hat C$ is the constant of \eqref{eqn:Inductive_hypothesis}. Since $M\ge1$, $\Xi_N$ is ounded by $(1+\log(c_KN_0))^{3/2}$, where $c_K$ is the number of events per block of Section~\ref{sec:LOLO_notation}, which depends only on $K$. We choose $N_0$ sufficiently large that
\begin{equation}
\label{eqn:N0_choice}
\frac{2\mathfrak C}{N_0}\big(1+\log(c_KN_0)\big)^{3/2}\le\frac14
\qquad\text{and}\qquad
\frac{2C_{\mathrm{fwd}}}{N_0}\big(1+\log(c_KN_0)\big)\le\frac14 .
\end{equation}
The first condition gives the smallness condition \eqref{eqn:block_smallness} of the backward pass below. The second gives $C_{\mathrm{fwd}}\hat{\mathfrak m}\le\frac14$ in the forward pass of Section~\ref{sec:decoupling}, which uses the same partition, since $\hat{\mathfrak m}=\varepsilon(1+\lambda_N)$, $\varepsilon\le2/N_0$ and $\lambda_N\le\log(c_KN_0)$. The constants $\mathfrak C$, $C_{\mathrm{fwd}}$ and $c_K$ depend only on $K$ and the constants $C'$ of Section~\ref{sec:LOLO_notation}, and not on $N_0$, since all the dependence on $N_0$ has been kept explicit through $\lambda_N$. Since the left-hand sides of \eqref{eqn:N0_choice} tend to zero as $N_0\to\infty$, such an $N_0$ exists. We now fix $N_0$. It depends neither on the training step nor on $\zeta$, and, since $L\ge N_0$, the partition satisfies $\varepsilon\le2/N_0$.

\paragraph{Notation.} We work on $\mathcal E_\delta\cap\mathcal E^{\mathrm{op}}_\delta$. For a removed layer $r$ and a step $k$, write
\[
\mathsf e_{r,k}:=\|\Delta^{(-r)}_{\wedge k}\|_\infty,\qquad
\mathsf h_{r,k}:=\|\Delta^{(-r),H}_{\wedge k}\|_\infty,\qquad
\mathsf e_{r,-1}:=0,
\]
and $\iota_r=\iota^H_r\vee\iota^A_r$ for the impulse \eqref{eqn:impulse_def}. Let
\[
\hat \gamma:=\Big(\frac83\Big)^{N_0}(2+C_\partial)\big(2+\eta_{\mathrm{in}}K|x|^2+C_\partial\tilde\eta_{\mathrm{out}}K\big),
\qquad
b_k:=\hat \gamma(1+b_{k-1}),\quad b_{-1}:=0,
\]
so that $b_k\le (k+1)\hat \gamma^{k+1}$, since $\hat \gamma\ge1$. These constants depend only on $N_0$ and the constants of Section~\ref{sec:LOLO_notation}. Let $\Gamma:=\mathfrak C(1+\log(c_KN_0))^{3/2}$ and $C'_\iota$ be the constant of
\begin{equation}\label{F:iota_avg}
\Big(\frac1{\mathfrak m}\sum_{r\in I_j}\iota_r^8\Big)^{1/8}\le C'L\varepsilon_\delta\le C'_\iota\big(1+\log(c_KN_0)\big),\qquad0\le j<N_0,
\end{equation}
which follows from \eqref{eqn:impulse_avg} and \eqref{eqn:Leps_bound}. Set
\[
\overline C_{\mathrm{bd}}:=C'_\iota\big(1+\log(c_KN_0)\big)\,b_K,
\qquad
\zeta_*:=\min\Big\{\frac1{8\overline C_{\mathrm{bd}}},\;\frac{N_0}{288\,\Gamma^2}\Big\}.
\]
Since $\varepsilon \le 2N_0^{-1}$, $\zeta_*$ satisfies
\[
\mathfrak C
(\varepsilon+\sqrt{\varepsilon\zeta_*})
(1+4\zeta_* \overline{C}_{\mathrm{bd}})\Xi_N
\le \frac{3\mathfrak{C}}{2}\left(\frac{2}{N_0}+\frac{1}{12\Gamma}\right)(1+\log(c_K N_0))^{\frac{3}{2}} \le \frac{1}{2},
\]
where the second inequality applies the constraint on $N_0$ \eqref{eqn:N0_choice}. Since the LHS of the smallness condition on $\zeta$ \eqref{eqn:block_smallness} increases in $\zeta$, it is satisfied for every $\zeta\le \zeta_*$.
Hence the restriction on the shape is
\begin{equation}
\label{eqn:Final_shape_constraints}
D\le c_0M,
\qquad
\frac{D}{LM^{2/3}}\le\zeta_*,
\end{equation}
where $\zeta_*$ depends only on $K$ and the constants in the assumptions, not on $L$, $M$, $D$ or $\delta$.
\paragraph{Two Pointwise Estimates.} Fix a step $k$ and a removed layer $r$. Then, \eqref{eqn:HiddenState_LOO_Recursion} reads as
\begin{equation}\label{F:fwd}
\mathsf h_{r,k}\le\Big(\frac43\Big)^{N_0}\Big[\big(1+\eta_{\mathrm{in}}K|x|^2\big)\mathsf e_{r,k-1}+\frac{\iota_r}L\Big].
\end{equation}
Next, suppose that the hypotheses of Lemma~\ref{lemma:Inductive_Step} hold for every block $I_{i'}$ with $i'\ge i$, at the step $k$. Set \[\mathsf{a}^r_{i'}:=\Delta^{(-r),A}_{[l_{i'},l_{i'+1}],\wedge k} \quad  \text{for} \quad  i'<N_0, \quad \mathsf{a}^r_{N_0}:=\Delta^{(-r),A}_{L,\wedge k}.\] Since $|\Delta^{(-r),A}_{l_{i'+1},\wedge k}|\le \mathsf{a}^r_{i'+1}$ by definition, Corollary~\ref{cor:Inductive_Step} gives
\[
\mathsf{a}^r_{i'}\le2\mathsf{a}^r_{i'+1}+\mathsf h_{r,k}+2\cdot\mathbbm 1_{\{r\in I_{i'}\}}\frac{\iota_r}L,\qquad i\le i'<N_0 .
\]
The impulse enters in at most one block, so iterating from $i'=N_0-1$ gives \[\mathsf{a}^r_{i'}\le2^{N_0}(\mathsf{a}^r_{N_0}+\mathsf h_{r,k}+\mathbbm{1}_{r>l_i}\iota_r/L).\] The terminal condition $\mathsf{a}^r_{N_0}$ is bounded by Lemma~\ref{lemma:boundary} to give
\begin{equation}\label{F:bwd}
\max_{i\le i'\le N_0}\mathsf{a}^r_{i'}\le2^{N_0}\Big[(1+C_\partial)\mathsf h_{r,k}+C_\partial\tilde\eta_{\mathrm{out}}K\,\mathsf e_{r,k-1}+\mathbbm{1}_{r>l_i}\frac{\iota_r}L\Big].
\end{equation}
If \eqref{F:bwd} holds with $i=0$, the ranges $[l_{i'},l_{i'+1}]$ cover all the adjoint variables, and so $\mathsf e_{r,k}$ is bounded by $\mathsf h_{r,k}$ together with $\max_{i'}\mathsf{a}^r_{i'}$. Substituting \eqref{F:fwd} and collecting the coefficients, each of which is at most $\hat \gamma$,
\begin{equation}\label{F:scalar}
\mathsf e_{r,k}\le \hat \gamma\Big(\mathsf e_{r,k-1}+\frac{\iota_r}L\Big).
\end{equation}
The same computation shows that, if \eqref{F:bwd} holds for $i+1$, then
\begin{equation}\label{F:boundary}
|\Delta^{(-r),A}_{l_{i+1},\wedge k}|\vee\mathsf h_{r,k}\le \hat \gamma\Big(\mathsf e_{r,k-1}+\frac{\iota_r}L\Big),
\end{equation}
since $|\Delta^{(-r),A}_{l_{i+1},\wedge k}|\le\mathsf{a}^r_{i+1}$. For $i=N_0-1$, \eqref{F:boundary} holds directly from Lemma~\ref{lemma:boundary} and \eqref{F:fwd}.

\paragraph{The Claim.} Assume $\zeta\le\zeta_*$. Then, for every $0\le k\le K$ and every removed layer $r$,
\begin{equation}\label{F:claim}
\mathsf e_{r,k}\le b_k\,\frac{\iota_r}L .
\end{equation}

\emph{Proof.} By induction over $k$. Assume \eqref{F:claim} at the step $k-1$. This assumption is vacuous at $k=0$, since $\mathsf e_{r,-1}=0=b_{-1}$. Then the right-hand side of \eqref{F:boundary} is at most $\hat \gamma(1+b_{k-1})\iota_r L^{-1}$, and $b_k = \hat \gamma(1+b_{k-1})$. We verify the hypotheses of Lemma~\ref{lemma:Inductive_Step} for each block $I_{i}$ in turn, at step $k$. Suppose they hold for the blocks above $I_i$. Then the previous argument yields that \eqref{F:boundary} holds for every removed layer $r\in I_i$, and by the inductive hypothesis
\[
|\Delta^{(-r),A}_{l_{i+1},\wedge k}|\vee\mathsf h_{r,k}\le b_k\frac{\iota_r}L.
\]
Taking the normalised $\ell^8$-average over $r\in I_i$ and using \eqref{F:iota_avg},
\[
\Big(\frac1{\mathfrak m}\sum_{r\in I_i}|\Delta^{(-r),A}_{l_{i+1},\wedge k}|^8\Big)^{1/8}\vee\Big(\frac1{\mathfrak m}\sum_{r\in I_i}\mathsf h_{r,k}^8\Big)^{1/8}
\le\frac{b_k}L\Big(\frac1{\mathfrak m}\sum_{r\in I_i}\iota_r^8\Big)^{1/8}
\le\frac{\overline C_{\mathrm{bd}}}L,
\]
using $b_k\le b_K$ as $\hat \gamma\ge 1$. This is \eqref{eqn:Inductive_hypothesis} with $\hat C=\overline C_{\mathrm{bd}}$. Since $\zeta\le\zeta_*$, \eqref{eqn:block_smallness} holds as well, and the hypotheses of Lemma~\ref{lemma:Inductive_Step} hold for $I_i$. As a result, the assumptions of Lemma~\ref{lemma:Inductive_Step} hold in every block, so \eqref{F:bwd} holds with $i=0$, and \eqref{F:scalar} with \eqref{F:claim} at the step $k-1$ gives \eqref{F:claim} at the step $k$. \hfill$\BlackBox$

\paragraph{Conclusion of Proposition \ref{lemma:MainLemma}.} Taking the normalised $\ell^8$-average of \eqref{F:claim} at $k=K$ over each block, and using \eqref{eqn:impulse_avg},
\begin{equation}\label{eqn:Final_LOLO_estimate}
\max_{0\le j<N_0}\Big(\frac1{\mathfrak m}\sum_{r\in I_j}\|\Delta^{(-r)}_{\wedge K}\|_\infty^8\Big)^{1/8}\le\frac{b_K}L\,C'L\varepsilon_\delta\le C_K\Big(\frac1L+\frac{D+\log(2N_0/\delta)}{LM}\Big),
\end{equation}
where the last inequality is \eqref{eqn:Leps_bound} with $C_K:=C'b_K$. As a by-product, Lemma~\ref{lemma:Inductive_Step} gives the diagonal estimate $\mathcal D^{[l,l_{i+1}]}_{k,8}\le4\overline C_{\mathrm{bd}}/L$ in every block and at every step.

\paragraph{No circularity.} The constants are fixed in the order $N_0\to\hat \gamma\to b_K\to\overline C_{\mathrm{bd}}\to\zeta_*$, and none of them depends on $\zeta$. The constant $c_K$ appearing in $\log(c_KN_0)$, in $\Gamma$, \eqref{F:iota_avg} and $\overline C_{\mathrm{bd}}$, is the number of events per block of Section~\ref{sec:LOLO_notation}, which depends only on $K$. It is not the growth constant $b_K$. Hence the choice \eqref{eqn:N0_choice} of $N_0$ involves only $\mathfrak C$, $C_{\mathrm{fwd}}$ and $c_K$, which are independent of $N_0$, of $b_K$ and of $\zeta$. The constants $\hat \gamma$ and $b_K$ grow with $N_0$, for example through $(8/3)^{N_0}$, but they are chosen after $N_0$ and only enter the restriction on $\zeta$, through $\overline C_{\mathrm{bd}}$ and $\zeta_*$. Finally, the claim \eqref{F:claim} at the step $k$ is used only at the later steps, and the smallness condition at the step $k$ is verified with the deterministic constant $\overline C_{\mathrm{bd}}$, which bounds all the steps at once. So no estimate is used before it is proved.

\subsubsection{The Decoupling Error}\label{sec:decoupling_cor}
With the diagonal estimate, Proposition~\ref{prop:comparison} bounds the decoupling errors of an arbitrary history by its distance to the history of the network.

\begin{corollary}[Decoupling Error]
\label{cor:MainLemma}
Suppose the assumptions of Lemma~\ref{lemma:apriori} hold, and assume $L\ge N_0$, $\delta\ge\exp(-c_1M^{1/3})$ and the shape
constraints \eqref{eqn:Final_shape_constraints}. Then, on $\mathcal E_\delta\cap\mathcal E^{\mathrm{op}}_\delta$,
simultaneously for every history $\bm Z$, every
$0\le k\le K$ and every $0\le l\le L$,
\begin{align}
\|\mathcal E^{H}_{l,k}(\bm Z)\|_{\bar2}
&\le \frac14\sum_{j:\,l_j<l}
\left(d^{H,(j)}_k(\bm Z)+\mathbbm{1}_{\{k>0\}}d^{(j)}_{k-1}(\bm Z)\right),\label{eqn:decoupling_blocks_H}\\
\|\mathcal E^{A}_{l,k}(\bm Z)\|_{\bar2}
&\le \frac12\sum_{j:\,l_{j+1}>l}d^{(j)}_k(\bm Z).\label{eqn:decoupling_blocks_A}
\end{align}
In particular, with $C_{\mathrm{dec}}:=N_0/2$,
\begin{align*}
\max_{0\le l\le L}\|\mathcal E^{H}_{l,k}(\bm Z)\|_{\bar2}
&\le C_{\mathrm{dec}}
\left(d^H_k(\bm Z)+\mathbbm 1_{\{k>0\}}d_{k-1}(\bm Z)\right),
&
\max_{0\le l\le L}\|\mathcal E^{A}_{l,k}(\bm Z)\|_{\bar2}
&\le C_{\mathrm{dec}}\,d_k(\bm Z).
\end{align*}
\end{corollary}

\begin{proof}
Splitting the layers $[1,l]$ and $[l+1,L]$ into their intersections with the blocks,
\[
\mathcal E^{H}_{l,k}(\bm Z)=\sum_{j:\,l_j<l}\mathcal E^{H}_{(l_j,\,l\wedge l_{j+1}],k}(\bm Z),
\qquad
\mathcal E^{A}_{l,k}(\bm Z)=\sum_{j:\,l_{j+1}>l}\mathcal E^{A}_{[l'_j,\,l_{j+1}],k}(\bm Z),
\quad l'_j:=(l+1)\vee(l_j+1).
\]
Each piece is an interval in a block, and its distances are at most those of the whole block. For the hidden states, \eqref{eqn:comparison_H} and $C_{\mathrm{fwd}}\hat{\mathfrak m}\le\frac14$, by the choice \eqref{eqn:N0_choice} of $N_0$, give \eqref{eqn:decoupling_blocks_H}. For the adjoint variables, the diagonal estimate of Section~\ref{sec:LOLO_induction} at the step $k$, which holds on $\mathcal E_\delta$ under \eqref{eqn:Final_shape_constraints}, gives
$\mathcal D^{[l'_j,l_{j+1}]}_{k,8}\le 4\overline C_{\mathrm{bd}}/L$, so that, by \eqref{eqn:U_bound},
\[
\mathcal U^{[l'_j,l_{j+1}]}_k
\le
\mathfrak C(\varepsilon+\sqrt{\varepsilon\zeta})
\left(1+4\zeta\overline C_{\mathrm{bd}}\right)\Xi_N .
\]
This is the left-hand side of the smallness condition \eqref{eqn:block_smallness} with $\hat C=\overline C_{\mathrm{bd}}$, and so is at most $\frac12$ by the choice of $\zeta_*$ in Section~\ref{sec:LOLO_induction}. Hence \eqref{eqn:comparison_A} gives \eqref{eqn:decoupling_blocks_A}. Finally, each sum has at most $N_0$ terms, each bounded by the corresponding maximum over $j$.
\end{proof}

\subsection{The Bias of the Residuals}\label{sec:bias}
Throughout this subsection we fix a layer and a neuron, and drop
their indices. Let $\theta_0=(\psi_0,\phi_0)\in\mathbb R^{2D}$
denote the initial parameters of the neuron, whose law satisfies Assumption~\ref{assm:Stein}. Given two deterministic histories
$\bm Z^0,\bm Z^1$ , such that $\| \bm{Z}^i\|_{\infty,\bar 2}\le C_B$ for $i=0,1$, define
\[
\Delta H_k:=H^1_k-H^0_k,\qquad \Delta A_k:=A^1_k-A^0_k,\qquad
\bm Z^s:=\bm Z^0+s(\bm Z^1-\bm Z^0),\quad s\in[0,1],
\]
and use the shorthand $P_{k,s}:=P_k(\bm Z^s)$, $Q_{k,s}:=Q_k(\bm Z^s)$. The interpolated histories $\bm Z^s$ are also bounded by $C_B$.
We measure the difference between the histories through the
one-dimensional projections, for $w\in\mathbb R^D$ and $0\le k\le K$,
\[
\mathfrak d^H_k(w):=|\langle w,\Delta H_{k}\rangle_{\bar2}|+\eta\sum_{k'<k}|\langle w,\Delta H_{k'}\rangle_{\bar2}|,
\quad
\mathfrak d^A_k(w):=|\langle w,\Delta A_{k}\rangle_{\bar2}|+\eta\sum_{k'<k}|\langle w,\Delta A_{k'}\rangle_{\bar2}|,
\]
through the projections onto the directions generated by the history
$\bm Z^0$,
\[
\mathfrak d_k(\bm Z^0):=\max_{0\le k''\le K} \max_{k'\le k}\Big(|\langle H^0_{k''},\Delta H_{k'}\rangle_{\bar2}|\vee|\langle A^0_{k''},\Delta A_{k'}\rangle_{\bar2}|\Big),
\]
and through the RMS error
$\mathfrak e_k:=\max_{k'\le k}\|\Delta H_{k'}\|_{\bar2}\vee\|\Delta A_{k'}\|_{\bar2}$.
Finally, for $u=(u_1,u_2)\in\mathbb R^{2D}$, define
\begin{equation}
\label{eqn:coordinatewise_error}
\mathfrak v_k(u):=\|u_1\odot\Delta H_{k}\|_{\bar2}+\|u_2\odot\Delta A_{k}\|_{\bar2}+\eta\sum_{k'<k}\big(\|u_1\odot\Delta H_{k'}\|_{\bar2}+\|u_2\odot\Delta A_{k'}\|_{\bar2}\big).
\end{equation}
We retain the sums over earlier steps in $\mathfrak d^H_k(w)$ and $\mathfrak d^A_k(w)$ so that, when averaging over directions in Section~\ref{sec:ProofThmMain}, the sums can be interchanged. This would not be possible with maxima over the steps. By contrast, $\mathfrak d_k(\bm Z^0)$ and $\mathfrak e_k$ are independent of $w$, so taking maxima is harmless.
For the refinement in part~3 below, and in Lemma~\ref{lemma:PQ_Lip_psi2}, we also use the versions in which the adjoint parts are taken at the step $k-1$:
\begin{gather*}
\mathfrak d^{A,-}_k(w):=\eta\sum_{k'<k}|\langle w,\Delta A_{k'}\rangle_{\bar2}|,
\qquad\mathfrak e^-_k:=\max_{k'\le k}\|\Delta H_{k'}\|_{\bar2}\vee\max_{k'\le k-1}\|\Delta A_{k'}\|_{\bar2},\\
\mathfrak d^-_k(\bm Z^0):=\max_{0\le k''\le K}\Big(\max_{k'\le k}|\langle H^0_{k''},\Delta H_{k'}\rangle_{\bar2}|\vee\max_{k'<k}|\langle A^0_{k''},\Delta A_{k'}\rangle_{\bar2}|\Big),
\\
\mathfrak v^-_k(u):=\|u_1\odot\Delta H_{k}\|_{\bar2}+\eta\sum_{k'<k}\big(\|u_1\odot\Delta H_{k'}\|_{\bar2}+\|u_2\odot\Delta A_{k'}\|_{\bar2}\big).
\end{gather*}

\begin{lemma}[Bias of the residuals]
\label{lemma:Lip_in_E} For some constants $c_f,n_f>0$,
let $f\in C^2(\mathbb R^2)$ satisfy
\[
|\nabla f(p,q)|+\|\nabla^2f(p,q)\|_{\mathrm{op}}\le c_f(1+|p|+|q|)^{n_f},
\]
and let $\bm Z^0,\bm Z^1$ be deterministic histories bounded by
$C_B$. Under Assumptions~\ref{assm:SteinMLU}, \ref{assm:rho} and \ref{assm:clipp}, there exists $C_{\E}>0$, depending only on
$c_f,n_f,C_B,K,\eta K,\Lambda_\rho,c,\sigma, \mu$ and $\mathbb V$,
such that for every $0\le k\le K$ the following hold.
\begin{enumerate}
\item Let $R\ge1$, $u_i\in\mathbb R^{D}$ with
$\|u_i\|_{\bar2}\le R$ for $i=1,2$, and let
$X:=\langle u_1,\psi_0\rangle_{\bar2}+\langle u_2,\phi_0\rangle_{\bar2}$. Then
\begin{align*}
\Big|\mathbb E\Big[X\big(f(P_{k,1},Q_{k,1})-f(P_{k,0},Q_{k,0})\big)\Big]\Big|
\le{}& C_{\E}\Big[\mathfrak d^H_k(u_1)+\mathfrak d^A_k(u_2)+\sqrt{\tfrac{\mathbb V}{D}}\,\mathfrak v_k(u)\Big]\\
&+R\,C_{\E}\Big[\mathfrak d_k(\bm Z^0)+\mathfrak e_k^2+\sqrt{\tfrac{\mathbb V}{D}}\,\|\bm Z^0\|_{\bar\infty}\,\mathfrak e_k\Big].
\end{align*}
\item Let $g\in C^2(\mathbb R^2)$ satisfy the same growth
condition as $f$, with constants $c_f,n_f$ enlarged if necessary, as well as $|g(0,0)|\le c_f$. Then, for any $0\le\mathfrak s\le K$,  the same
bound holds for $X:=g(P_{\mathfrak s,0},Q_{\mathfrak s,0})$, taking $R=1$ and without the first bracket.
\item If $f$ does not depend on $q$, then the adjoint parts may all be taken at the step $k-1$. That is, parts~1 and~2 hold with $\mathfrak d^A_k(u_2)$, $\mathfrak d_k(\bm Z^0)$, $\mathfrak e_k$ and $\mathfrak v_k(u)$ replaced by $\mathfrak d^{A,-}_k(u_2)$, $\mathfrak d^-_k(\bm Z^0)$, $\mathfrak e^-_k$ and $\mathfrak v^-_k(u)$, respectively.
\end{enumerate}
\end{lemma}

\begin{proof}
All constants $C$ below depend only on the quantities listed in
the statement. By Lemma~\ref{lemma:PQ_psi_2}, $P_{k,s},Q_{k,s}$
are sub-Gaussian uniformly in $s\in[0,1]$ and $k\le K$, and $X$ has finite $p$-th moment for every $p\ge 1$ by Assumption~\ref{assm:SteinMLU}, in part~1, or by the growth condition on $g$, in part~2. Hence, by the growth conditions on $f$, for every $q\ge1$,
\[
\|\nabla f_s\|_{L^q}+\|\nabla^2f_s\|_{L^q}+\|X\|_{L^q}\le C_q,
\qquad \nabla f_s:=\nabla f(P_{k,s},Q_{k,s}),
\]
with $C_q$ depending on $q$ and on the quantities in the statement.
By Lemma~\ref{lemma:BoundMathcalR}, the series over $k'\le k$ of response matrices $\mathcal R_{k',k}(\bm Z^s)$ are a.s.\ bounded by
$(1+\eta C_{\mathcal R})^K\le e^{\eta KC_{\mathcal R}}$. In part~1 we may assume $R=1$ as both sides' dependence on $u$, through $X$ and the first bracket, scales linearly with $u$.

\emph{Step 1: Interpolation.} By the fundamental theorem of
calculus and the linear response formula
\eqref{eqn:VolterraResponse}, applied to the interpolation $\bm Z^s$, for which $\partial_sH^s_{k}=\Delta H_{k}$ and $\partial_sA^s_k=\Delta A_k$,
\begin{align}
\begin{pmatrix}P_{k,1}-P_{k,0}\\ Q_{k,1}-Q_{k,0}\end{pmatrix}
&=\int_0^1\sum_{k'\le k}\mathcal R_{k',k}(\bm Z^s)\,
\mathcal B_{k'}(\bm Z^s)\,ds,\label{eqn:bias_interpolation}\\
f_1-f_0:=f(P_{k,1},Q_{k,1})-f(P_{k,0},Q_{k,0})
&=\int_0^1\sum_{k'\le k}\nabla f_s^\top\mathcal R_{k',k}(\bm Z^s)\,
\mathcal B_{k'}(\bm Z^s)\,ds,\notag
\end{align}
where
\[
\mathcal B_{k'}(\bm Z^s)=
\begin{pmatrix}\langle\psi_0,\Delta H_{k'}\rangle_{\bar2}\\
\langle\phi_0,\Delta A_{k'}\rangle_{\bar2}\end{pmatrix}
-\eta\sum_{k''<k'}
\begin{pmatrix}\rho'(P_{k'',s})\mathrm{cl}(Q_{k'',s})\\ \rho(P_{k'',s})\end{pmatrix}
\odot
\begin{pmatrix}\partial_s\Gamma^{H^s}_{k'',k'}\\ \partial_s\Gamma^{A^s}_{k'',k'}\end{pmatrix}.
\]
By Fubini's theorem, which applies by the moment bounds above, we may exchange the expectation with the integral in $s$. Accordingly, we split the expectation of
$X(f_1-f_0)$ into two parts $T_1+T_2$,
where $T_1$ contains the first vector in $\mathcal B$ and $T_2$
the remaining sum.

\emph{Step 2: The Gram Terms $T_2$.} Since the histories are
deterministic,
\[
\partial_s\Gamma^{H^s}_{k'',k'}
=\langle H^0_{k''},\Delta H_{k'}\rangle_{\bar2}
+\langle\Delta H_{k''},H^0_{k'}\rangle_{\bar2}
+2s\langle\Delta H_{k''},\Delta H_{k'}\rangle_{\bar2},
\]
so that $|\partial_s\Gamma^{H^s}_{k'',k'}|\le2\mathfrak d_k(\bm Z^0)+2\mathfrak e_k^2$ for $k'',k'\le k$,
and likewise for $A$. The remaining factors of $T_2$ are $X$, $\nabla f_s$, $\mathcal R$, $\rho'(P)\mathrm{cl}(Q)$ and $\rho(P)$, all of which have bounded moments of every order. Thus, by H\"older's inequality,
\[
|T_2|\le C\big(\mathfrak d_k(\bm Z^0)+\mathfrak e_k^2\big).
\]

\emph{Step 3: Stein's Identity.}
Consider the part of $T_1$ involving $\psi_0$, where the part involving $\phi_0$ follows identically,
with $(H,u_1,\psi)$ replaced by $(A,u_2,\phi)$. For fixed $s$, write \[W_{k'}:=\nabla f_s^\top(\mathcal R_{k',k}(\bm Z^s))_{\cdot1},\]
to express this part as
\[
\sum_{k'\le k}\mathbb E\big[XW_{k'}\langle\psi_0,\Delta H_{k'}\rangle_{\bar2}\big]
% =\frac1D\sum_{k'\le k}\sum_{i=1}^D\Delta H^i_{k'}\,\mathbb E\big[\psi_0^iXW_{k'}\big]
=\sum_{k'\le k}\sum_{i=1}^D\Delta H^i_{k'}\,
\mathbb E\big[\kappa(\psi_0^i)\big(\partial_{\psi^i}X\,W_{k'}+X\,\partial_{\psi^i}W_{k'}\big)\big].
\]
The equality results from Stein's identity (Assumption~\ref{assm:SteinMLU}) in the coordinate $\psi_0^i$, with the other
coordinates fixed. Here, the integrability conditions hold since $\mathbb E[\kappa(\psi^i_0)^2]=\mu^2+\mathbb V$ and all the other factors have moments of every order.
Now, we expand the two derivatives. The first is $\partial_{\psi^i}X=u_{1,i}/D$. The second, by \eqref{eqn:PQ_Deriv_theta} along $\bm Z^s$ and by \eqref{eqn:MathcalRDerivs_form} of Lemma~\ref{lemma:MathcalRDerivs}, applied to the deterministic history $\bm Z^s$, is
\[
\partial_{\psi^i}W_{k'}=\frac1D\sum_{k''\le k}H^{s,i}_{k''}\,V_{k',k''},
\qquad
V_{k',k''}:=(\mathcal R_{k'',k})_{\cdot1}^\top\,\nabla^2f_s\,(\mathcal R_{k',k})_{\cdot1}+\mathbbm 1_{\{k''<k\}}\nabla f_s^\top\big(\mathcal N^{\psi}_{k''}\big)_{\cdot1},
\]
where $\mathcal N^\psi_{k''}$ are the matrices of Lemma~\ref{lemma:MathcalRDerivs} for the pair $(k',k)$, which vanish if $k'=k$. Hence the part of $T_1$ involving $\psi_0$ equals
\begin{equation}
\label{eqn:Stein_expression}
\frac1D\sum_{k'\le k}\sum_{i=1}^Du_{1,i}\,\Delta H^i_{k'}\,\mathbb E\big[\kappa(\psi_0^i)\,W_{k'}\big]
+\frac1D\sum_{k',k''\le k}\sum_{i=1}^DH^{s,i}_{k''}\,\Delta H^i_{k'}\,\mathbb E\big[\kappa(\psi_0^i)\,X\,V_{k',k''}\big],
\end{equation}
where $W_{k'}$, $X$ and $V_{k',k''}$ are the same random variables for every $i$, although they still depend on $\psi_0^i$. By Lemma~\ref{lemma:BoundMathcalR}, $\mathcal R_{k,k}=I_2$ and $\|\mathcal R_{k',k}\|_{\mathrm{op}}\le\eta C_{\mathcal R}e^{\eta K C_{\mathcal R}}$ for $k'<k$, and, by \eqref{eqn:MathcalRDerivs_bound},
\[
\sum_{k'<k}\sum_{k''<k}\|\mathcal N^\psi_{k''}\|_{\mathrm{op}}\le\sqrt2\,\eta K\,C'(1+\eta C')^{2K} .
\]
Together with the moment bounds on $X$, $\nabla f_s$ and $\nabla^2f_s$ and H\"older's inequality, this gives
\begin{equation}
\label{eqn:Stein_weights}
\|W_k\|_{L^2}\le C,\qquad \|W_{k'}\|_{L^2}\le C\eta\quad(k'<k),\qquad \sum_{k',k''\le k}\|X\,V_{k',k''}\|_{L^2}\le C .
\end{equation}

\emph{Step 4: Mean and Fluctuation.} We split
$\kappa(\psi_0^i)=\mu+(\kappa(\psi_0^i)-\mu)$ in \eqref{eqn:Stein_expression}. By \eqref{eqn:Stein_weights} and $\eta k\le\eta K$, the mean part is at most $\mu$ times
\[
\sum_{k'\le k}|\langle u_1,\Delta H_{k'}\rangle_{\bar2}|\,\|W_{k'}\|_{L^2}
+\sum_{k',k''\le k}|\langle H^s_{k''},\Delta H_{k'}\rangle_{\bar2}|\,\|X\,V_{k',k''}\|_{L^2}
\le C\Big(\mathfrak d^H_k(u_1)+\mathfrak d_k(\bm Z^0)+\mathfrak e_k^2\Big),
\]
since $H^s_{k''}=H^0_{k''}+s\Delta H_{k''}$. For the fluctuation part, let
\[
S_{k'}(x):=\sum_{i=1}^D\big(\kappa(\psi_0^i)-\mu\big)x_i\,\Delta H^i_{k'},\qquad x\in\R^D .
\]
The variables $\kappa(\psi_0^i)-\mu$ are i.i.d., centred with variance $\mathbb V$, and so
\[
\|S_{k'}(x)\|_{L^2}=\sqrt{\mathbb VD}\,\|x\odot\Delta H_{k'}\|_{\bar2} .
\]
The Cauchy--Schwarz inequality in each expectation and \eqref{eqn:Stein_weights} give
\begin{align*}
\Big|\frac1D\sum_{k'\le k}\mathbb E\big[W_{k'}S_{k'}(u_1)\big]\Big|
&\le C\sqrt{\tfrac{\mathbb V}D}\,\Big(\|u_1\odot\Delta H_k\|_{\bar2}+\eta\sum_{k'<k}\|u_1\odot\Delta H_{k'}\|_{\bar2}\Big),\\
\Big|\frac1D\sum_{k',k''\le k}\mathbb E\big[X\,V_{k',k''}\,S_{k'}(H^s_{k''})\big]\Big|
&\le C\sqrt{\tfrac{\mathbb V}D}\,\max_{k',k''\le k}\|H^s_{k''}\odot\Delta H_{k'}\|_{\bar2} .
\end{align*}
The first line, added to the analogous $\phi_0$-part in terms of $u_2$ and $\Delta A_{k'}$, is precisely $C\sqrt{\mathbb V/D}\,\mathfrak v_k(u)$. For the second line,
\[
\|H^s_{k''}\odot\Delta H_{k'}\|_{\bar2}\le\|H^0_{k''}\|_{\bar\infty}\|\Delta H_{k'}\|_{\bar2}+\|\Delta H_{k''}\|_2\,\|\Delta H_{k'}\|_{\bar2}\le\|\bm Z^0\|_{\bar\infty}\,\mathfrak e_k+\sqrt D\,\mathfrak e_k^2,
\]
and analogously for the $\phi_0$-part. Altogether,
\[
|T_1|\le C\Big[\mathfrak d^H_k(u_1)+\mathfrak d^A_k(u_2)+\mathfrak d_k(\bm Z^0)+\mathfrak e_k^2
+\sqrt{\tfrac{\mathbb V}{D}}\big(\mathfrak v_k(u)+\|\bm Z^0\|_{\bar\infty}\mathfrak e_k\big)\Big],
\]
which, together with Step~2, proves 1.

\emph{Step 5:} Let $X=g(P_{\mathfrak s,0},Q_{\mathfrak s,0})$.
The only property of $X=\langle u,\theta_0\rangle$ used above,
other than its moments, is the form of its derivative in Step~3.
Now, by \eqref{eqn:PQ_Deriv_theta} along $\bm Z^0$,
\[
\partial_{\psi^i}X=\sum_{k''\le\mathfrak s}\nabla g^\top(\mathcal R_{k'',\mathfrak s}(\bm Z^0))_{\cdot1}H^{0,i}_{k''}/D,
\]
which is of the form of Step~3 with $x=H^0_{k''}$, and similarly for $\partial_{\phi^i}X$ with $x=A^0_{k''}$. Hence, the
terms $\mathfrak d^H_k(u_1)$, $\mathfrak d^A_k(u_2)$ and $\mathfrak v_k(u)$ are
replaced by terms already contained in $\mathfrak d_k(\bm Z^0)$ and
$\|\bm Z^0\|_{\bar\infty}\mathfrak e_k$, which proves 2.

\emph{Step 6: Proof of 3.} In Steps~1--4 for the pair $(\bm Z^0,\bm Z^1)$, the difference $\Delta A_k$ at the last step only enters through terms that vanish when $\nabla f=(\partial_pf,0)$. To make this precise without tracking these terms, let $\bm Z^{1,-}$ be the history which coincides with $\bm Z^1$ except that \\$A^{1,-}_{k'}:=A^0_{k'}$ for $k'\ge k$. It is deterministic and bounded by $C_B$, and $P_k(\bm Z^{1,-})=P_k(\bm Z^1)$, since $P_k$ does not depend on $(A_{k'})_{k'\ge k}$. If $f$ does not depend on $q$, then \[f(P_{k,1},Q_{k,1})=f(P_k(\bm Z^{1,-}),Q_k(\bm{Z}^{1,-})),\] while $X$ is unchanged, as $X$ only depends on $\bm Z^0$ in part~2. Applying parts~1 and~2 to the pair $(\bm Z^0,\bm Z^{1,-})$, the quantities $\mathfrak d^A_k(u_2)$, $\mathfrak d_k(\bm Z^0)$, $\mathfrak e_k$ and $\mathfrak v_k(u)$ of this pair are exactly $\mathfrak d^{A,-}_k(u_2)$, $\mathfrak d^-_k(\bm Z^0)$, $\mathfrak e^-_k$ and $\mathfrak v^-_k(u)$, while $\mathfrak d^H_k(u_1)$ and $\|\bm Z^0\|_{\bar\infty}$ are unchanged. This proves 3.
\end{proof}

\begin{remark}[Conditional form]\label{rem:Lip_in_E_conditional}
Lemma~\ref{lemma:Lip_in_E} is applied conditionally. Let $\mathcal G$ be a $\sigma$-algebra independent of $\theta_0$, let $\bm Z^0,\bm Z^1$ be $\mathcal G$-measurable random histories bounded by $C_B$, and let $u$ be $\mathcal G$-measurable with $\|u_1\|_{\bar2}\vee\|u_2\|_{\bar2}\le R$. Since $\theta_0$ is independent of $\mathcal G$, the conditional expectation given $\mathcal G$ of a jointly measurable and integrable function of $\theta_0$ and of $\mathcal G$-measurable variables is obtained by computing the expectation over $\theta_0$ with the latter frozen. Hence parts~1--3 hold almost surely for the conditional expectations, and so does Lemma~\ref{lemma:PQ_Lip_psi2} for the conditional $\psi_2$-norms, given $\mathcal G$, with the right-hand sides computed from the $\mathcal G$-measurable histories and directions. In Section~\ref{sec:ProofThmMain}, $\mathcal G=\mathcal F_{-r}$, which by Appendix~\ref{sec:Notation} contains $\mathcal F_W$, $\bm Z^1=\widetilde{\bm Z}_r$, which is $\mathcal F_{-r}$-measurable, $\bm Z^0=\bm Z^\infty_r$, which is $\mathcal F_W$-measurable and bounded by $C_{\mathrm{lim}}\le C_B$ on the $\mathcal F_W$-measurable event $\mathcal E^W_\delta$, where the conclusions are used, and $u$ is built from $\mathcal F_W$-measurable directions.
\end{remark}

\subsection{Fluctuations of the Leave-One-Out Residuals}\label{sec:fluct}
Let us use the notation of Section~\ref{sec:bias}.

\begin{lemma}[Lipschitz dependence of the features on the history]\label{lemma:PQ_Lip_psi2}
Suppose Assumptions~\ref{assm:SteinMLU}, \ref{assm:rho} and~\ref{assm:clipp} hold, and let $\bm Z^0,\bm Z^1$ be deterministic histories bounded by $C_B$. There is $C_P$, depending only on $C_B,\eta K,\Lambda_\rho,c$ and $\sigma$, such that, for every $0\le k\le K$,
\[
\|P_{k,1}-P_{k,0}\|_{\psi_2}\vee\|Q_{k,1}-Q_{k,0}\|_{\psi_2}\le C_P\,\mathfrak e_k,
\qquad
\|P_{k,1}-P_{k,0}\|_{\psi_2}\le C_P\,\mathfrak e^-_k .
\]
\end{lemma}
\begin{proof}
By the first identity in \eqref{eqn:bias_interpolation} and Lemma~\ref{lemma:BoundMathcalR},
\[
|P_{k,1}-P_{k,0}|\vee|Q_{k,1}-Q_{k,0}|\le  \int_0^1\Big(|\mathcal B_k(\bm{Z}^s)|+\sum_{k'< k}\eta C_{\mathcal R}(1+\eta C_{\mathcal R})^{k-{k'}-1}|\mathcal B_{k'}(\bm Z^s)|\Big)\,ds .
\]
Since $\bm Z^i$ are assumed to be bounded by $C_B$, $|\partial_s\Gamma^{H^s}_{k'',k'}|\vee|\partial_s\Gamma^{A^s}_{k'',k'}|$, for $k''<k'$, is bounded by $ 2C_B\mathfrak{e}_{k'}$. Hence, by Assumptions~\ref{assm:rho} and~\ref{assm:clipp},
\[
|\mathcal B_{k'}(\bm Z^s)|\le|\langle\psi_0,\Delta H_{k'}\rangle_{\bar2}|+|\langle\phi_0,\Delta A_{k'}\rangle_{\bar2}|+2\eta\Lambda_\rho C_B(c\vee 1)\,\mathfrak e_{k'}\sum_{k''<k'}\big(2+|Q_{k'',s}|+|P_{k'',s}|\big).
\]
By Assumption~\ref{assm:SteinMLU}, $\|\langle\psi_0,x\rangle_{\bar2}\|_{L^q}\le C\sigma\sqrt q\,\|x\|_{\bar2}$ for every $q\ge1$ and deterministic $x\in\mathbb R^D$, and similarly for $\phi_0$. Furthermore, by Lemma~\ref{lemma:PQ_psi_2}, $\|P_{k'',s}\|_{L^q}\vee \|Q_{k'',s}\|_{L^q}\le C\sqrt q$ uniformly in $s$. Therefore, applying Minkowski's integral inequality, simplifying the geometric series using the monotonicity of $\mathfrak{e}_{k'}$ in $k'$, we obtain
\[
\|P_{k,1}-P_{k,0}\|_{L^q}\vee\|Q_{k,1}-Q_{k,0}\|_{L^q}\le C\sqrt q\,\mathfrak e_k
\] 
for every $q\ge1$. This is equivalent to the first claim \cite[Proposition~2.6.1]{Vershynin_2026}. 

\paragraph{}For the second bound, let $\bm Z^{1,-}$ be as in Step~6 of Lemma~\ref{lemma:Lip_in_E}. Since $P_k$ does not depend on $(A_{k'})_{k'\ge k}$, we have $P_k(\bm Z^{1,-})=P_k(\bm Z^1)$. Moreover, $\mathfrak e_k$ for the pair $(\bm Z^0,\bm Z^{1,-})$ equals $\mathfrak e^-_k$ for $(\bm Z^0,\bm Z^1)$. Applying the first bound to $(\bm Z^0,\bm Z^{1,-})$ therefore gives the second result. 
\end{proof}
By Remark~\ref{rem:Lip_in_E_conditional}, the lemma also holds for the conditional $\psi_2$-norms given a $\sigma$-algebra independent of $\theta_0$, with histories measurable with respect to it.

\begin{lemma}[Fluctuation of the leave-one-out residuals]
\label{lemma:Fluct}
Suppose Assumptions~\ref{assm:SteinMLU}, \ref{assm:rho} and~\ref{assm:clipp} hold. For $1\le r\le L$, let $\bm Z^1_r:=\widetilde{\bm Z}_r$ be the comparison history of Section~\ref{sec:LOLO_notation} and $\bm Z^0_r:=\bm Z^\infty_r$, and let $v\in\mathbb R^D$ be $\mathcal F_W$-measurable. Let $\mathcal E$ be an $\mathcal F_W$-measurable event on which the histories $\bm Z^\infty_r$, $1\le r\le L$, are bounded by $C_B$, for instance $\mathcal E=\mathcal E^W_{\delta/2}$, by Lemma~\ref{lemma:C1} and $C_{\mathrm{lim}}\le C_B$. For $0\le j\le K$, set
\begin{align*}
\Delta_r(j)&:=\max_{k'\le j}\|\bm Z^1_{r,k'}-\bm Z^0_{r,k'}\|_{\bar2},\qquad \bar\Delta(j):=\left(\frac1L\sum_{r=1}^L\Delta_r(j)^2\right)^{1/2},
\\
X^H_r(v;j)&:=\frac1M\sum_{m=1}^{M}\Big\langle v,\phi^{rm}_j(\bm Z^1_r)\rho\big(P^{rm}_j(\bm Z^1_r)\big)
-\phi^{rm}_j(\bm Z^0_r)\rho\big(P^{rm}_j(\bm Z^0_r)\big)\Big\rangle_{\bar2},
\\
X^{H,\mathrm{cen}}_r(v;j)&:=X^H_r(v;j)-\mathbb E[X^H_r(v;j)\,|\,\mathcal F_{-r}],
\qquad
S^H_j(v):=\max_{1\le l\le L}\max_{k'\le j}\Big|\frac1L\sum_{r=1}^{l}X^{H,\mathrm{cen}}_r(v;k')\Big| .
\end{align*}
Then, there exists $C_{\mathrm{conc}}$, depending only on $\sigma,\eta K,C_B,\Lambda_\rho,K$ and $c$, such that, for every $\delta\in(0,1]$, the following holds with conditional probability at least $1-\delta$ given $\mathcal F_W$, almost surely on $\mathcal E$, simultaneously for all $0\le j\le K$,
\begin{equation}
\label{eqn:Fluct_tail}
S^H_j(v)
\le C_{\mathrm{conc}}\|v\|_{\bar2}\Big(\bar\Delta(j)+\frac1L\Big)\,
\mathcal T^{M}_{\delta,3/2},
\end{equation}
with $\mathcal T^{M}_{\delta,3/2}$ the rate of Appendix~\ref{sec:Notation}. Moreover:
\begin{enumerate}
\item For every $\delta\in(0,1]$, the following holds with conditional probability at least $1-\delta$ given $\mathcal F_W$, almost surely on $\mathcal E$,
\[
\mathcal Z(v):=\max_{0\le j\le K}\frac{S^H_j(v)}{\|v\|_{\bar2}\big(\bar\Delta(j)+\frac1L\big)}
\le a_1+a_2\sqrt{\log(e/\delta)}+a_3\log(e/\delta)^{3/2},
\]
where
\[
a_2:=\frac{C_{\mathrm{conc}}}{\sqrt M},\qquad a_3:=\frac{C_{\mathrm{conc}}}M,\qquad a_1:=a_2\sqrt{\log(e(K+1))}+a_3\log(e(K+1))^{3/2},
\]
which is the form of tail required by Lemma~\ref{lemma:LogRemoval_mixed}.
\item Since $\mathcal E\in\mathcal F_W$, both bounds also hold with probability at least $1-\delta$ on $\mathcal E$. That is, the probability that $\mathcal E$ holds and the bound fails is at most $\delta$.
\end{enumerate}
The same holds for the adjoint variables, with $\phi\rho(P)$ replaced by
$\psi\rho'(P)Q$, where the sums are instead over layers $L\ge r>l$.
\end{lemma}

\begin{proof}
Both sides of \eqref{eqn:Fluct_tail} are $1$-homogeneous in $v$, so we assume $\|v\|_{\bar2}=1$. Throughout, $C$ denotes a constant depending only on the quantities listed in the statement, which may change from line to line. Furthermore, for ease of notation, all $\psi_\alpha$-norms below are conditional on $\mathcal F_{-r}$, for appropriate $r$.

\emph{Step 1: Conditioning.} The history $\widetilde{\bm Z}_r$ is $\mathcal F_{-r}$-measurable and bounded by $C_B$, by construction. On the $\mathcal F_W$-measurable event $\mathcal E$, the limiting history $\bm Z^\infty_r$ is bounded by $C_B$. Both $\bm Z^\infty_r$ and $v$ are $\mathcal F_W$-measurable, hence $\mathcal F_{-r}$-measurable. Hence, conditionally on $\mathcal F_{-r}$, both histories and $v$ are fixed, the $M$ summands $\xi_{rm}$ of $X^H_r(v;j)$ are i.i.d., and Lemmas~\ref{lemma:PQ_psi_2} and~\ref{lemma:PQ_Lip_psi2} apply as in Remark~\ref{rem:Lip_in_E_conditional}. 

\emph{Step 2: Sub-Weibull Bounds.} Fix $r$ and $j$, drop the indices $r,m$, and write
\[
P^i_k:=P_k(\bm Z^i_r),\qquad Q^i_k:=Q_k(\bm Z^i_r),\qquad (H^i_k,A^i_k):=\bm Z^i_{r,k},\qquad i\in\{0,1\}.
\]
We claim that
\begin{equation}\label{eqn:Fluct_xi}
\|\xi\|_{\psi_{2/3}}\le C\Delta_r(j).
\end{equation}
By Lemma~\ref{lemma:PQ_psi_2}, Assumption \ref{assm:MLU} and Lemma~\ref{lemma:PQ_Lip_psi2}, for $k\le j$,
\begin{align*}
\|P^i_k\|_{\psi_2}\vee\|Q^i_k\|_{\psi_2}\vee\|\langle v,\phi_0\rangle_{\bar2}\|_{\psi_2}\vee\|\langle v,\psi_0\rangle_{\bar2}\|_{\psi_2}&\le C,\\
\|P^1_k-P^0_k\|_{\psi_2}\vee\|Q^1_k-Q^0_k\|_{\psi_2}&\le C\Delta_r(j),
\end{align*}
and, by the Cauchy--Schwarz inequality,
\[
|\langle v,H^1_k-H^0_k\rangle_{\bar2}|\vee|\langle v,A^1_k-A^0_k\rangle_{\bar2}|\le\Delta_r(j),
\qquad
|\langle v,H^0_k\rangle_{\bar2}|\vee|\langle v,A^0_k\rangle_{\bar2}|\le C_B .
\]
For the hidden states, expanding the update of $\phi_j$ along both histories gives
\begin{align*}
\xi
={}&\langle v,\phi_0\rangle_{\bar2}\big(\rho(P^1_j)-\rho(P^0_j)\big)
-\eta\sum_{k'<j}\langle v,A^1_{k'}-A^0_{k'}\rangle_{\bar2}\,\rho(P^1_{k'})\rho(P^1_j)\\
&-\eta\sum_{k'<j}\langle v,A^0_{k'}\rangle_{\bar2}\Big(\rho(P^1_{k'})\big(\rho(P^1_j)-\rho(P^0_j)\big)+\big(\rho(P^1_{k'})-\rho(P^0_{k'})\big)\rho(P^0_j)\Big).
\end{align*}
By Assumption~\ref{assm:rho}, $\rho$ is Lipschitz with linear growth. Therefore, each term is a bounded deterministic factor times a product of two sub-Gaussian factors, and one factor of each term is of order $\Delta_r(j)$. Since a product of two sub-Gaussian variables has $\psi_1$-norm at most a constant times the product of their $\psi_2$-norms \cite[Proposition~D.2]{kuchibhotla2022moving} and that the $\psi_{2/3}$-quasi-norm is dominated by the $\psi_1$-norm, claim \eqref{eqn:Fluct_xi} follows. For the adjoint variables, let
\[
G^i:=\rho'(P^i_{k'})\,\mathrm{cl}(Q^i_{k'})\,\rho'(P^i_j)\,Q^i_j .
\]
Expanding the update of $\psi_j$ along both histories gives
\begin{align*}
\xi
={}&\langle v,\psi_0\rangle_{\bar2}\Big(\big(\rho'(P^1_j)-\rho'(P^0_j)\big)Q^1_j+\rho'(P^0_j)\big(Q^1_j-Q^0_j\big)\Big)\\
&-\eta\sum_{k'<j}\Big(\langle v,H^1_{k'}-H^0_{k'}\rangle_{\bar2}\,G^1+\langle v,H^0_{k'}\rangle_{\bar2}\big(G^1-G^0\big)\Big).
\end{align*}
By Assumptions~\ref{assm:rho} and \ref{assm:clipp}, $\rho'$ and $\mathrm{cl}$ are bounded and Lipschitz, so that
\[
|G^1|\le C|Q^1_j|,
\qquad
|G^1-G^0|\le C\big(|P^1_{k'}-P^0_{k'}|+|Q^1_{k'}-Q^0_{k'}|+|P^1_j-P^0_j|\big)|Q^1_j|+C|Q^1_j-Q^0_j| .
\]
The first term of $\xi$ contains a product of three sub-Gaussian factors, one of which is of order $\Delta_r(j)$, and so the $\psi_{\frac23}$-quasi-norm of $\xi$ is of order $\Delta_r(j)$ \cite[Proposition D.2]{kuchibhotla2022moving}. Then, claim \eqref{eqn:Fluct_xi} for the adjoint variable residual follows by the same argument as that for the hidden states.

\emph{Step 3: Normalisation} Let
\[
\Delta^+_r(j):=\Delta_r(j)+\frac1L,
\]
which is positive and $\mathcal F_{-r}$-measurable. By \eqref{eqn:Fluct_xi}, conditionally on $\mathcal F_{-r}$, the variable \\$X^{H,\mathrm{cen}}_r(v;j)/\Delta^+_r(j)$ is an average of $M$ i.i.d.\ centred variables with $\psi_{2/3}$-norm at most $C$. The concentration inequality for sums of independent sub-Weibull variables \cite[Theorem~3.1]{kuchibhotla2022moving}, with $t=\log(e/\delta')$, gives, for every $\delta'\in(0,1]$, with probability at least $1-\delta'$ when conditioned on $\mathcal F_{-r}$,
\[
\frac{|X^{H,\mathrm{cen}}_r(v;j)|}{\Delta^+_r(j)}\le C\bigg(\sqrt{\frac{\log(e/\delta')}M}+\frac{\log(e/\delta')^{3/2}}M\bigg).
\]
The RHS is deterministic, so the bound also holds conditionally on $\mathcal F_W$, on $\mathcal E$.

\emph{Step 4: Maximal Bounds.} Let
\[
\Upsilon_r:=\max_{0\le j\le K}\frac{|X^{H,\mathrm{cen}}_r(v;j)|}{\Delta^+_r(j)},
\qquad
b:=\sqrt{\frac{\log(e(K+1))}M}+\frac{\log(e(K+1))^{3/2}}M .
\]
A union bound over the $K+1$ steps, with $\delta'=\delta/(K+1)$, gives, for each $r$ and every $\delta\in(0,1]$, with probability at least $1-\delta$,
\[
\Upsilon_r\le C\bigg(b+\sqrt{\frac{\log(e/\delta)}M}+\frac{\log(e/\delta)^{3/2}}M\bigg),
\]
having separated the contributions of $K$ from $\delta$. Lemma~\ref{lemma:LogRemoval_mixed}, with $p=2$, uniform weights and $\beta=\frac32$, only uses these marginal tails, and no independence across the layers, and so we apply it under the conditional law given $\mathcal F_W$, on $\mathcal E$ (Remark~\ref{rem:LogRemoval_weights}). It gives, with conditional probability at least $1-\delta$,
\begin{equation}
\label{eqn:Fluct_Y}
\Big(\frac1L\sum_{r=1}^L\Upsilon_r^2\Big)^{1/2}\le C\bigg(b+\sqrt{\frac{\log(e/\delta)}M}+\frac{\log(e/\delta)^{3/2}}M\bigg)\le C\,\mathcal T^M_{\delta,3/2},
\end{equation}
where the last step uses
\[
\log(e(K+1))\vee\log(e/\delta)\le\mathfrak l_\delta .
\]

\emph{Step 5: Conclusion.} Since $\Delta_r(\cdot)$ is non-decreasing, $|X^{\mathrm{cen}}_r(v;k')|\le\Delta^+_r(j)\,\Upsilon_r$ for $k'\le j$. By the Cauchy--Schwarz and Minkowski inequalities, for every $0\le j\le K$,
\[
S_j(v)
\le\frac1L\sum_{r=1}^L\Delta^+_r(j)\,\Upsilon_r
\le\Big(\bar\Delta(j)+\frac1L\Big)\Big(\frac1L\sum_{r=1}^L\Upsilon_r^2\Big)^{1/2}.
\]
With \eqref{eqn:Fluct_Y}, this proves \eqref{eqn:Fluct_tail}, and the first inequality of \eqref{eqn:Fluct_Y} gives part~1. Part~2 follows by taking the expectation of the conditional failure-probability bound on the $\mathcal F_W$-measurable event $\mathcal E$. The adjoint case is identical, with the sums over $r>l$.
\end{proof}

\section{Monte Carlo Error}\label{sec:MC}
For $f:\R^2\to\R$ with $|f(p,q)|\le c_f(1+|p|+|q|)$ and $\theta\in\{\phi,\psi\}$, define the centred partial sums
\begin{equation}
\label{eqn:MC_sum}
S^{\theta,f}_{l,k}:=\frac1{LM}\sum_{r=1}^l\sum_{m=1}^M\Big(\hat\theta^{rm}_k\,f\big(\hat P^{rm}_k,\hat Q^{rm}_k\big)-\mathbb E\big[\hat\theta^{rm}_k\,f\big(\hat P^{rm}_k,\hat Q^{rm}_k\big)\,\big|\,\mathcal F_W\big]\Big)\in\R^D,
\end{equation}
for $0\le l\le L,\ 0\le k\le K.$
Here, as in Section~\ref{sec:LimitingModel}, the hats denote the parameters and features of the neurons of layer $r$ trained along the limiting history $\bm Z^\infty_r$, which is $\mathcal F_W$-measurable. The finite width Monte Carlo errors of Section~\ref{sec:ProofMain} are, at a deterministic step $k$,
\begin{equation}
\label{eqn:MC_as_sums}
\mathcal E^H_{\mathrm{m.c.},l,k}=S^{\phi,\rho}_{l,k},
\qquad
\mathcal E^A_{\mathrm{m.c.},l,k}=S^{\psi,f}_{L,k}-S^{\psi,f}_{l,k}\quad\text{with }f(p,q)=\rho'(p)q .
\end{equation}

\begin{lemma}[Monte Carlo error]
\label{lemma:MC}
Under Assumptions~\ref{assm:MLU}, \ref{assm:rho} and~\ref{assm:clipp}, let $B>0$ and let $\mathcal E$ be an $\mathcal F_W$-measurable event on which the histories $\bm Z^\infty_r$, $1\le r\le L$, are bounded by $B$. Then, there exists $C$, depending only on $\sigma,K,\eta K,B,\Lambda_\rho,c$ and $c_f$, such that, for every $\mathcal F_W$-measurable $v\in\R^D$ and every $\delta\in(0,1]$, the following holds with conditional probability at least $1-\delta$ given $\mathcal F_W$, almost surely on $\mathcal E$
\begin{equation}
\max_{l\le L}\max_{k\le K}\big|\langle v,S^{\theta,f}_{l,k}\rangle_{\bar2}\big|
\le C\|v\|_{\bar2}\,\mathcal T^{LM}_{\delta,1}.\label{eqn:MC_direction}
\end{equation}
Similarly, for every $\delta\in(0,1]$, the following holds with conditional probability at least $1-\delta$ given $\mathcal F_W$, almost surely on $\mathcal E$
\begin{equation}
\max_{l\le L}\max_{k\le K}\big\|S^{\theta,f}_{l,k}\big\|_{\bar2}
\le C\,\mathcal T^{D,LM}_{\delta,1}.\label{eqn:MC_norm}
\end{equation}
\end{lemma}
\begin{proof}
We work conditionally on $\mathcal F_W$, on $\mathcal E$, so that $v$ and the histories are fixed and bounded by $B$. Fix $k$, and let
\[
\xi_{rm}:=\big\langle v,\hat\theta^{rm}_k\big\rangle_{\bar2}\,f\big(\hat P^{rm}_k,\hat Q^{rm}_k\big)-\mathbb E\big[\big\langle v,\hat\theta^{rm}_k\big\rangle_{\bar2}\,f\big(\hat P^{rm}_k,\hat Q^{rm}_k\big)\big],
\]
thus $\langle v,S^{\theta,f}_{l,k} \rangle_{\bar 2}$ is the average of the $lM$ centred random variables $\xi_{rm}$. Dropping the indices $r,m$, the updates of Section~\ref{sec:Setup} along $\bm Z^\infty_r=(H^\infty_k,A^\infty_k)_k$ give
\[
\langle v,\hat\theta_k\rangle_{\bar2}=\langle v,\theta_0\rangle_{\bar2}-\eta\sum_{k'<k}\langle v,X_{k'}\rangle_{\bar2}\,g\big(\hat P_{k'},\hat Q_{k'}\big),
\qquad
(X,g)=\begin{cases}(A^\infty,\ \rho(p)),&\theta=\phi,\\ (H^\infty,\ \rho'(p)\mathrm{cl}(q)),&\theta=\psi.\end{cases}
\]
The three terms are respectively bounded in $\psi_2$ by Assumption \ref{assm:MLU}, Cauchy--Schwarz and Lemma \ref{lemma:PQ_psi_2}, according to
\begin{align*}
\|\langle v,\theta_0\rangle_{\bar2}\|_{\psi_2}\le C\sigma\|v\|_{\bar2},\qquad
|\langle v,X_{k'}\rangle_{\bar2}|\le B\|v\|_{\bar2},\qquad 
\|\hat P_{k'}\|_{\psi_2}\vee\|\hat Q_{k'}\|_{\psi_2}&\le C.
\end{align*}
By the linear growth of $f$ and $g$, both $f(\hat P_k,\hat Q_k)$ and $g(\hat P_{k'},\hat Q_{k'})$ are bounded in $\psi_2$, and hence $\|\langle v,\hat\theta_k\rangle_{\bar2}\|_{\psi_2}\le C\|v\|_{\bar2}$. A product of two sub-Gaussian variables is sub-exponential \cite[Proposition~D.2]{kuchibhotla2022moving}, and so $\|\xi_{rm}\|_{\psi_1}\le C\|v\|_{\bar2}$. Furthermore, $\xi_{rm}$ are centred and independent across $(r,m)$, so Bernstein's inequality \cite[Theorem~2.9.1]{Vershynin_2026} applies to $ \langle v,S^{\theta,f}_{l,k}\rangle_{\bar 2}$ and yields
\[
\mathbb P\Big(\big|\langle v,S^{\theta,f}_{l,k}\rangle_{\bar2}\big|>t\Big)
\le2\exp\Big(-\,LM\min\Big\{\frac{t^2}{C^2\|v\|_{\bar2}^2},\frac{t}{C\|v\|_{\bar2}}\Big\}\Big),
\]
for every $l\le L$. Following \cite[Lemma A.2]{chizat2025hiddenwidthdeepresnets}, we apply the L\'evy--Ottaviani maximal inequality \cite{ZbigniewS2013} to give
\[
\mathbb P\left(\max_{l\le L}\big|\langle v,S^{\theta,f}_{l,k}\rangle_{\bar2}\big|>3t\right)
\le 3\max_{l\le L}\mathbb P\left(\big|\langle v,S^{\theta,f}_{l,k}\rangle_{\bar2}\big|>t\right),
\]
and absorbing the numerical constants into $C$, we obtain
\[
\mathbb P\Big(\max_{l\le L}
|\langle v,S^{\theta,f}_{l,k}\rangle_{\bar2}|>t\Big)
\le 6\exp\Big(-\,LM\min\Big\{
\frac{t^2}{(3C)^2\|v\|_{\bar2}^2},
\frac{t}{3C\|v\|_{\bar2}}
\Big\}\Big).
\]
With $x:=\log(6(K+1)/\delta)$ and $t=3C\|v\|_{\bar2}\big(\sqrt{x/(LM)}+x/(LM)\big)$, a union bound over the $K+1$ steps gives \eqref{eqn:MC_direction}.

\paragraph{}Finally, the $\bar2$-norm bound follows from the standard $\varepsilon$-net argument of \cite[Lemma~A.2]{chizat2025hiddenwidthdeepresnets}, with a union bound over the $5^D$ points of a $\frac12$-net, which replaces $\mathfrak l_\delta$ by $\mathfrak l_\delta+D\log5$.
\end{proof}

\section{Proof of the Main Theorems}\label{sec:ProofMain}
\paragraph{Probability Bookkeeping.} Each theorem is proved on the intersection of a fixed number $n$ of events, independent of $L,M,D$ and $\delta$, each of which is applied at the level $\delta/n$. Since $\mathfrak l_\delta\ge1$, we have $\mathfrak l_{\delta/n}\le(1+\log n)\,\mathfrak l_\delta$, and similarly for $\mathfrak l^\infty_\delta$, so the factor $1+\log n$ is absorbed into the constants. The hypothesis $\delta\ge\exp(-c_1(D\wedge M^{1/3}))$ gives $\delta/n\ge\exp(-(c_1+\log n)(D\wedge M^{1/3}))$, since $D\wedge M^{1/3}\ge1$, so that every constant is computed with $c_1$ replaced by $(c_1+\log n)$. 

\paragraph{}The events over which we take the intersection are of two kinds. The first kind are the \textit{leave-one-layer-out} event $\mathcal E_{\delta/n}$ of Section~\ref{sec:LOLO_notation}, $\mathcal E^{\mathrm{op}}_{\delta/n}$ of Lemma \ref{lemma:apriori} and that of Lemma \ref{lemma:InfNorm} when $\mathbb V>0$, which hold with probability at least $1-\delta/n$. The second kind are from Lemmas~\ref{lemma:Fluct} and~\ref{lemma:MC}. They hold with conditional probability at least $1-\delta/n$ given $\mathcal F_W$, almost surely on the $\mathcal F_W$-measurable event $\mathcal E^W:=\mathcal E^W_{\delta/n}$ of Lemma~\ref{lemma:embed_event}. Since $\mathcal E^W\in\mathcal F_W$, an event $F$ of the second kind satisfies $\mathbb P(\mathcal E^W\setminus F)=\mathbb E\big[\mathbbm 1_{\mathcal E^W}\,\mathbb P(F^c\,|\,\mathcal F_W)\big]\le\delta/n$. Counting $\mathcal E^W$ itself among the $n$ events, a union bound shows that all of them hold simultaneously with probability at least $1-\delta$.

\subsection{Error Decomposition}
The numerical scheme and the
limiting equation give the exact decomposition
\begin{equation}\label{eqn:error_decomp}
h_{l,k}-H_{\frac lL,k}
=\big(h_{0,k}-H_{0,k}\big)+\mathcal E^{H}_{\mathrm{T.D.},l,k}+\mathcal E^{H}_{\mathrm{m.c.},l,k}
+\mathcal E^{H}_{\mathrm{prop},l,k},
\end{equation}
where
\begin{align*}
\mathcal E^{H}_{\mathrm{T.D.},l,k}
&:=\sum_{r=1}^l\int_{\frac{r-1}L}^{\frac rL}
\Big(\frac1M\mathbb E\big[\hat\Phi^r_{k}\bm\rho(\hat P^r_{k})\big]
-\mathbb E\big[\phi^\infty_{t,k}\,\rho(P_{t,k})\big]\Big)\,dt,\\
\mathcal E^{H}_{\mathrm{m.c.},l,k}
&:=\frac1{LM}\sum_{r=1}^l\Big(\hat\Phi^r_{k}\bm\rho(\hat P^r_{k})
-\mathbb E\big[\hat\Phi^r_{k}\bm\rho(\hat P^r_{k})\big]\Big),\\
\mathcal E^{H}_{\mathrm{prop},l,k}
&:=\frac1{LM}\sum_{r=1}^l\Big(\Phi^r_{k}\bm\rho(p^r_{k})
-\hat\Phi^r_{k}\bm\rho(\hat P^r_{k})\Big).
\end{align*}

For the adjoint variables, write $\bm f^r_k:=\bm\rho'(p^r_k)\odot q^r_k$, as in Section~\ref{sec:LOLO_backward}, and $\hat{\bm f}^r_k:=\bm\rho'(\hat P^r_k)\odot\hat Q^r_k$. The same steps, downwards from the last layer, give
\[
a_{l,k}-A_{\frac lL,k}
=\big(a_{L,k}-A_{1,k}\big)
+\mathcal E^{A}_{\mathrm{T.D.},l,k}+\mathcal E^{A}_{\mathrm{m.c.},l,k}
+\mathcal E^{A}_{\mathrm{prop},l,k},
\]
where
\begin{align*}
\mathcal E^{A}_{\mathrm{T.D.},l,k}
&:=\sum_{r=l+1}^L\int_{\frac{r-1}L}^{\frac rL}
\Big(\frac1M\mathbb E\big[\hat\Psi^r_{k}\hat{\bm f}^r_{k}\big]
-\mathbb E\big[\psi^\infty_{t,k}\,\rho'(P_{t,k})\,Q_{t,k}\big]\Big)\,dt,\\
\mathcal E^{A}_{\mathrm{m.c.},l,k}
&:=\frac1{LM}\sum_{r=l+1}^L\Big(\hat\Psi^r_{k}\hat{\bm f}^r_{k}
-\mathbb E\big[\hat\Psi^r_{k}\hat{\bm f}^r_{k}\big]\Big),\\
\mathcal E^{A}_{\mathrm{prop},l,k}
&:=\frac1{LM}\sum_{r=l+1}^L\Big(\Psi^r_{k}\bm f^r_{k}
-\hat\Psi^r_{k}\hat{\bm f}^r_{k}\Big).
\end{align*}

\paragraph{Time Discretisation and Monte Carlo errors.}
By the bounds on the derivatives in $t$ in the proof of Lemma~\ref{lemma:C1}, with $B=C_{\mathrm{lim}}$,
\[
\|\partial_t H_{t,k}\|_{\bar2}\le o_\infty(C_{\mathrm{lim}})\Lambda_\rho\big(1+o_\infty(C_{\mathrm{lim}})C_{\mathrm{lim}}\big),\qquad\|\partial_tA_{t,k}\|_{\bar2}\le\Lambda_\rho\,o_\infty(C_{\mathrm{lim}})^2C_{\mathrm{lim}},
\]
uniformly over $t\in[0,1]$ and $k\le K$. Hence $H_{\cdot,k}$ and $A_{\cdot,k}$ are Lipschitz on $[0,1]$, on $\mathcal E^W_\delta$. Hence, for $t\in[\frac{r-1}L,\frac rL]$, the history $\bm Z^\infty_r$ is within $C'/L$ of the history at depth $t$, and Lemma~\ref{lemma:PQ_Lip_psi2} gives $\|\hat P^r_k-P_{t,k}\|_{\psi_2}\vee\|\hat Q^r_k-Q_{t,k}\|_{\psi_2}\le C'/L$. Writing the parameters through their update equations, the Cauchy--Schwarz inequality bounds the $\bar 2$-norm of each integrand in $\mathcal E^H_{\mathrm{T.D.},l,k}$ by $C'/L$, uniformly in $t$, $r$ and $k$. Summing at most $L$ integrals over intervals of length $1/L$, and arguing similarly for the adjoint variables, gives
\begin{equation}\label{eqn:B_TD}
\max_{l,k}
\left(
\|\mathcal E^H_{\mathrm{T.D.},l,k}\|_{\bar2}
\vee
\|\mathcal E^A_{\mathrm{T.D.},l,k}\|_{\bar2}
\right)
\le\frac{C_{\mathrm{T.D.}}}{L},
\end{equation}
for a constant $C_{\mathrm{T.D.}}$ depending only on the constants of the limiting model.

\paragraph{} The Monte Carlo errors are the sums \eqref{eqn:MC_as_sums} of Section~\ref{sec:MC}, along the histories $\bm Z^\infty_r$. By Lemma~\ref{lemma:MC}, with $\mathcal E=\mathcal E^W$ and $B= C_{\mathrm{lim}}$, applied at level $\delta/2$ to the hidden states and to the adjoint variables, for each $\mathcal F_W$-measurable direction $v$, the following holds with conditional probability at least $1-\delta$ given $\mathcal F_W$, almost surely on $\mathcal E^W$:
\begin{equation}
\label{eqn:MC_projection}
\max_{0\le l\le L}\max_{0\le j\le K}|\langle v,\mathcal E^{H}_{\mathrm{m.c.},l,j}\rangle_{\bar2}|\vee|\langle v,\mathcal E^{A}_{\mathrm{m.c.},l,j}\rangle_{\bar2}|
\le C\|v\|_{\bar2}\,\mathcal T^{LM}_{\delta,1},
\end{equation}
and the same holds for the $\bar2$-norms:
\begin{equation}
\label{eqn:MC_norm_bound}
\max_{0\le l\le L}\max_{0\le j\le K}\|\mathcal E^{H}_{\mathrm{m.c.},l,j}\|_{\bar2}\vee\|\mathcal E^{A}_{\mathrm{m.c.},l,j}\|_{\bar2}
\le C\,\mathcal T^{D,LM}_{\delta,1}.
\end{equation}

\subsection{Proof of Theorem \ref{thm:bar 2-norm}}\label{sec:ProofThm2bar}
We work on the intersection of $\mathcal E_\delta$, $\mathcal E^{\mathrm{op}}_\delta$, $\mathcal E^W$ and the Monte Carlo event \eqref{eqn:MC_norm_bound}, so that $n=4$ in the probability bookkeeping. For $k\le K$, write
\[
\varepsilon^{\bar2}_k:=\max_{0\le l\le L}\max_{k'\le k}\|h_{l,k'}-H_{\frac lL,k'}\|_{\bar2}\vee\|a_{l,k'}-A_{\frac lL,k'}\|_{\bar2},
\qquad \varepsilon^{\bar2}_{-1}:=0,
\]
and $\mathfrak u:=C_{\mathrm{T.D.}}/L+C\,\mathcal T^{D,LM}_{\delta,1}$ for the bound on the time discretisation and Monte Carlo errors. By definition, $\mathcal E^{H}_{\mathrm{prop}}$ is the decoupling
error $\mathcal E^H(\bm Z^\infty)$ of Section~\ref{sec:decoupling} with the history
$\bm Z^\infty$, which is bounded by $C_{\mathrm{lim}}\le C_B$, and similarly for the adjoint variables. Fix $k\le K$. Since the block distances $d^{(j)}_{k}(\bm Z^\infty)$ of Corollary~\ref{cor:MainLemma} are $\ell^8$-averages over the block $I_j$, they are at most the corresponding maxima over the block. Let
\[
\mathfrak h_i:=\max_{0\le l\le l_i}\max_{k'\le k}\|h_{l,k'}-H_{\frac lL,k'}\|_{\bar2},\qquad 0\le i\le N_0,
\]
so that $\mathfrak h_0\le\eta_{\mathrm{in}} K|x|^2\varepsilon^{\bar2}_{k-1}$, by \eqref{eqn:boundary_input}. The hidden-state errors entering $d^{H,(j)}_{k}$ are at most $\mathfrak h_{j+1}$, and $d^{(j)}_{k-1}\le\varepsilon^{\bar2}_{k-1}$. For $l\in I_i$, the decomposition of Section~\ref{sec:ProofMain} and \eqref{eqn:decoupling_blocks_H}, in which the blocks with $l_j<l$ are those with $j\le i$, give
\[
\|h_{l,k}-H_{\frac lL,k}\|_{\bar2}\le \mathfrak h_0+\mathfrak u+\frac12\sum_{j=0}^{i}\mathfrak h_{j+1}+\frac{N_0}2\varepsilon^{\bar2}_{k-1}.
\]
The right-hand side is non-decreasing in $k$, so the same bound holds at every $k'\le k$, and taking the maximum over $l\le l_{i+1}$ and $k'\le k$ gives the same bound for $\mathfrak h_{i+1}$. Since $\mathfrak h_{i+1}$ is finite, absorbing $\frac12\mathfrak h_{i+1}$ gives
\[
\mathfrak h_{i+1}\le b+\sum_{j<i} \mathfrak h_{j+1},\qquad b:=2\mathfrak u+\big(N_0+2\eta_{\mathrm{in}} K|x|^2\big)\varepsilon^{\bar2}_{k-1},
\]
and the discrete Gr\"onwall inequality, upwards over the $N_0$ blocks, gives $\mathfrak h_{N_0}\le2^{N_0}b$.
For the adjoint variables, \eqref{eqn:boundary_terminal} bounds the terminal condition via
\[
\|a_{L,k'}-A_{1,k'}\|_{\bar2}\le C_\partial\big(\mathfrak h_{N_0}+\tilde\eta_{\mathrm{out}} K\varepsilon^{\bar2}_{k-1}\big),\qquad k'\le k.
\] Let \[\mathfrak a_i:=\max_{ l_i\le l \le L}\max_{k'\le k}\|a_{l,k'}-A_{\frac lL,k'}\|_{\bar2}.\] For $l_i\le l<l_{i+1}$, the blocks with $l_{j+1}>l$ in \eqref{eqn:decoupling_blocks_A} are those with $j\ge i$, and $d^{(j)}_k\le \mathfrak h_{N_0}+\mathfrak a_j$, so that
\[
\mathfrak a_i\le C_\partial\big(\mathfrak h_{N_0}+\tilde\eta_{\mathrm{out}} K\varepsilon^{\bar2}_{k-1}\big)+\mathfrak u+\frac12\sum_{j= i}^{N_0-1}\big(\mathfrak h_{N_0}+\mathfrak a_j\big).
\]
Absorbing $\frac12\mathfrak a_i$ gives \[\mathfrak a_i\le b'+\sum_{j=i+1}^{N_0-1}\mathfrak a_j, \qquad b':=2\mathfrak u+(2C_\partial+N_0)\mathfrak h_{N_0}+2C_\partial\tilde\eta_{\mathrm{out}} K\varepsilon^{\bar2}_{k-1},\] and the discrete Gr\"onwall inequality, now downwards over the blocks, gives $\mathfrak a_0\le2^{N_0}b'$. Altogether,
\[
\varepsilon^{\bar2}_k
\le \mathfrak h_{N_0}+\mathfrak a_0
\le C_{N_0}\big(\mathfrak u+\varepsilon^{\bar2}_{k-1}\big),
\]
where $C_{N_0}$ depends only on $N_0$, $C_\partial$, $\widetilde\eta_{\mathrm{out}}K$ and $\eta_{\mathrm{in}}K|x|^2$. Iterating over the training steps, starting from $\varepsilon^{\bar2}_{-1}=0$, gives
\[
\varepsilon^{\bar2}_K
\le (K+1)(1\vee C_{N_0})^{K+1}\mathfrak u.
\]
Since $N_0$ is independent of $L,M,D$, and the shape constraints \eqref{eqn:shape_main} imply \[D+\mathfrak{l}_\delta \le (1+c_0+c_1+\log(K+1)) LM,\] we may absorb the linear term in $\mathcal T^{D,LM}_{\delta,1}$ into the square-root term. Thus,
\[
\varepsilon^{\bar2}_K\le C\left(\frac1L
+\mathcal T^{D,LM}_{\delta,1}\right)
\le C\left(\frac1L+\sqrt{\frac{D+\mathfrak l_\delta}{LM}}\right).
\]
This is Theorem~\ref{thm:bar 2-norm}.

\subsection{Proof of Theorem \ref{thm:MainTheorem}}\label{sec:ProofThmMain}
Every direction $w\in\R^D$ below is $\mathcal F_W$-measurable.

\subsubsection*{Step 1: Splitting the projected errors}
\paragraph{Three histories for each layer.} After expanding the numerical scheme satisfied by our variables, the Propagation Error compares the residual at layer $r$ defined using the history $\bm z_r=(h_{r-1},a_r)$ of the discrete model with that using the history $\bm Z^\infty_r$ of the limiting model. Between the two, we insert the comparison history of Section~\ref{sec:LOLO_notation},
\[
\bm z^{\mathrm{diag}}_r:=\widetilde{\bm Z}_r ,
\]
which is independent of the parameters of layer $r$ and, on $\mathcal E^{\mathrm{op}}_\delta$, coincides with the leave-one-layer-out history $(h^{(-r)}_{r-1},a^{(-r)}_r)$. Thus, we consider two distances,
\begin{align}
\Delta^{\mathrm{diag}}_r&:=\max_{0\le k\le K}\|\bm z^{\mathrm{diag}}_{r,k}-\bm z_{r,k}\|_{\bar2}
=\max_{0\le k\le K}\|h^{(-r)}_{r-1,k}-h_{r-1,k}\|_{\bar2}\vee\|a^{(-r)}_{r,k}-a_{r,k}\|_{\bar2},\label{eqn:Delta_diag_def}\\
\Delta^Z_r&:=\max_{0\le k\le K}\|\bm z^{\mathrm{diag}}_{r,k}-\bm Z^\infty_{r,k}\|_{\bar2},\label{eqn:Delta_Z_def}
\end{align}

\paragraph{Three bounds from the earlier sections.} The following are the only estimates on the $\bar2$-errors that are used in this subsection.
\begin{align}
\varepsilon^{\bar2}_K&\le C\Big(\frac1L+\sqrt{\frac{D+\mathfrak l_\delta}{LM}}\Big)
&&\text{(Theorem~\ref{thm:bar 2-norm})},\label{eqn:B_eps2}\\
\frac1L\sum_{r=1}^L\Delta^{\mathrm{diag}}_r\le\Big(\frac1L\sum_{r=1}^L(\Delta^{\mathrm{diag}}_r)^2\Big)^{1/2}&\le2C_K\varepsilon_\delta
&&\text{(by \eqref{eqn:Final_LOLO_estimate})},\label{eqn:B_diag}\\
\Delta^Z_r\le\Delta^{\mathrm{diag}}_r+\varepsilon^{\bar2}_K,
\quad
\Big(\frac1L\sum_{r=1}^L(\Delta^Z_r)^2\Big)^{1/2}&\le 2 C_K\varepsilon_\delta+\varepsilon^{\bar2}_K
&&\text{(triangle ineq.)}.\label{eqn:B_DeltaZ}
\end{align}
For \eqref{eqn:B_diag}, note that $\Delta^{\mathrm{diag}}_r\le\|\Delta^{(-r)}_{\wedge K}\|_\infty$, that the $\ell^2$-average over a block is at most the $\ell^8$-average bounded in \eqref{eqn:Final_LOLO_estimate}, and that there are $N_0$ blocks, each of length at most $\mathfrak m\le2L/N_0$. For \eqref{eqn:B_DeltaZ}, we pass through the discrete history $\bm z_r$, and use the Minkowski inequality for the average over $r$.

\paragraph{The five pieces.} We split the propagation error of Section~\ref{sec:ProofMain} through the history $\bm z^{\mathrm{diag}}$,
\[
\mathcal E^{H}_{\mathrm{prop}}=\widehat{\mathcal E}^H+\widetilde{\mathcal E}^H,
\qquad
\widehat{\mathcal E}^H:=\mathcal E^H(\bm z^{\mathrm{diag}}),
\]
\[
\widetilde{\mathcal E}^H_{l,k}:=\frac1{LM}\sum_{r=1}^l
\Big(\Phi^r_{k}(\bm z^{\mathrm{diag}}_r)\bm\rho\big(P^r_{k}(\bm z^{\mathrm{diag}}_r)\big)
-\hat\Phi^r_{k}\bm\rho(\hat P^r_{k})\Big).
\]
The first is the decoupling error of Section~\ref{sec:decoupling} with the history $\bm z^{\mathrm{diag}}$. The second compares the residuals along two histories that are both independent of the parameters of layer $r$. With $X^H_r(w;j)$ as in Lemma~\ref{lemma:Fluct}, its projection onto a direction $w$ is
\[
\langle w,\widetilde{\mathcal E}^H_{l,k}\rangle_{\bar2}=\frac1L\sum_{r=1}^l X^H_r(w;k)
=\frac1L\sum_{r=1}^l X^{H,\mathrm{cen}}_r(w;k)+\frac1L\sum_{r=1}^l \mathbb E\big[X^H_r(w;k)\,\big|\,\mathcal F_{-r}\big],
\]
where $X^{H,\mathrm{cen}}_r:=X^H_r-\mathbb E[X^H_r|\mathcal F_{-r}]$.
Together with the decomposition of Section~\ref{sec:ProofMain}, this gives, for every direction $w$, layer $l$ and step $k$,
\begin{align}
\label{eqn:five_terms_H}
e^H_{l,k}(w)={}&\underbrace{\big\langle w,\mathcal E^H_{\mathrm{T.D.},l,k}\big\rangle_{\bar2}
+\big\langle w,\mathcal E^H_{\mathrm{m.c.},l,k}\big\rangle_{\bar2}
+\big\langle w,\widehat{\mathcal E}^H_{l,k}\big\rangle_{\bar2}
+\frac1L\sum_{r=1}^{l}X^{H,\mathrm{cen}}_r(w;k)}_{=:\,R^H_{l,k}(w)}\\
&+\frac1L\sum_{r=1}^ l\mathbb E\big[X^H_r(w;k)\,\big|\,\mathcal F_{-r}\big]+e^H_{0,k}(w).\notag
\end{align}
The last term is the error of the input condition, due to the learnable input embeddings. By \eqref{eqn:embedding_updates}, it is an adjoint error at the earlier steps, in the same direction:
\begin{equation}\label{eqn:input_projection}
e^H_{0,k}(w)=-\eta_{\mathrm{in}}|x|^2\sum_{k'<k}e^A_{0,k'}(w).
\end{equation}
For the adjoint variables, let $\widehat{\mathcal E}^A:=\mathcal E^A(\bm z^{\mathrm{diag}})$, and let $X^A_r$ and $X^{A,\mathrm{cen}}_r$ be the adjoint versions of $X^H_r$ and $X^{H,\mathrm{cen}}_r$ in Lemma~\ref{lemma:Fluct}. The analogous decomposition gives
\begin{align}
\label{eqn:five_terms_A}
e^A_{l,k}(w)={}&\underbrace{\big\langle w,\mathcal E^A_{\mathrm{T.D.},l,k}\big\rangle_{\bar2}
+\big\langle w,\mathcal E^A_{\mathrm{m.c.},l,k}\big\rangle_{\bar2}
+\big\langle w,\widehat{\mathcal E}^A_{l,k}\big\rangle_{\bar2}
+\frac1L\sum_{r=l+1}^L X^{A,\mathrm{cen}}_r(w;k)}_{=:\,R^A_{l,k}(w)}\\
&+\frac1L\sum_{r=l+1}^L\mathbb E\big[X^A_r(w;k)\,\big|\,\mathcal F_{-r}\big]
+\big\langle w,a_{L,k}-A_{1,k}\big\rangle_{\bar2}.\notag
\end{align}

\emph{The Decoupling Error.} The distances of Corollary~\ref{cor:MainLemma} between $\bm z^{\mathrm{diag}}$ and $\bm z$ are block averages of $\Delta^{\mathrm{diag}}_r$, which are bounded by \eqref{eqn:Final_LOLO_estimate}. Hence the corollary gives
\begin{equation}
\label{eqn:B_dec}
\max_{l,k}\|\widehat{\mathcal E}^H_{l,k}\|_{\bar2}\vee\|\widehat{\mathcal E}^A_{l,k}\|_{\bar2}
\le2C_{\mathrm{dec}}C_K\,\varepsilon_\delta .
\end{equation}

\emph{The Fluctuation.} By Lemma~\ref{lemma:Fluct}, with $\mathcal E=\mathcal E^W$, the following holds with conditional probability at least $1-\delta$ given $\mathcal F_W$, almost surely on $\mathcal E^W$:
\[
\max_{1\le l \le L}\max_{k'\le k}\Big|\frac1L\sum_{r =1}^l X^{H,\mathrm{cen}}_r(w;k')\Big|
\le C_{\mathrm{conc}}\|w\|_{\bar2}\Big(\bar\Delta(k)+\frac1L\Big)\,\mathcal T^{M}_{\delta,3/2}.
\]
Here $\bar\Delta(k)$ is the root mean square over $r$ of the distances between $\widetilde{\bm Z}_r$ and $\bm Z^\infty_r$ up to the step $k$, which is at most the root mean square of $\Delta^Z_r$. By \eqref{eqn:B_DeltaZ},
\begin{equation}\label{eqn:Fluc_H_error}
\max_{1\le l \le L}\max_{k'\le k}\Big|\frac1L\sum_{r=1}^l X^{H,\mathrm{cen}}_r(w;k')\Big|
\le C\|w\|_{\bar2}\Big(\varepsilon_\delta+\varepsilon^{\bar2}_K+\frac1L\Big)\,\mathcal T^{M}_{\delta,3/2}.
\end{equation}
The same argument holds for the adjoint variables.
\subsubsection*{Step 2: The bias}
Fix a layer $r$ and a direction $v$ with $\|v\|_{\bar2}\le C_{\mathrm{lim}}\vee1$. Conditionally on $\mathcal F_{-r}$, the histories $\widetilde{\bm Z}_r$ and $\bm Z^\infty_r$ are fixed and bounded by $C_B$, and the neurons of layer $r$ are i.i.d., so that $\mathbb E[X^\cdot_r(v;j)|\mathcal F_{-r}]$ is the expectation over a single neuron. In this step only, we use the notation of Section~\ref{sec:bias} for this pair of histories:
\[
(H^1_k,A^1_k):=\widetilde{\bm Z}_{r,k},\qquad
(H^0_k,A^0_k):=\bm Z^\infty_{r,k}=\big(H_{\frac{r-1}L,k},A_{\frac rL,k}\big),
\]
\[
\Delta H_k:=H^1_k-H^0_k,\qquad\Delta A_k:=A^1_k-A^0_k,
\]
and $P^i_k,Q^i_k$ for the features of the neuron trained along $(H^i,A^i)$, $i\in\{0,1\}$. Lemma~\ref{lemma:Lip_in_E} is applied conditionally on $\mathcal F_{-r}$ and at every deterministic step $j\le K$, as in Remark~\ref{rem:Lip_in_E_conditional}. The quantities $\mathfrak d^H_k(u_1)$, $\mathfrak d^{A,-}_k(u_2)$, $\mathfrak d^-_k(\bm Z^0)$, $\mathfrak e^-_k$ and $\mathfrak v^-_k(u)$ of Section~\ref{sec:bias} then measure the difference between the comparison history $\widetilde{\bm Z}_r$ and the limiting history $\bm Z^\infty_r$. To distinguish them from the errors $e^H_{l,k}$ and $e^A_{l,k}$ between the discrete and the limiting model, of Section~\ref{sec:LimitingModel}, we write them with a tilde: $\tilde{\mathfrak d}^H_k(u_1)$, $\tilde{\mathfrak d}^{A,-}_k(u_2)$, $\tilde{\mathfrak d}^-_k$, $\tilde{\mathfrak e}^-_k$ and $\tilde{\mathfrak v}^-_k(u)$.

\paragraph{The hidden states: three terms.} Since $\phi_j=\phi_0-\eta\sum_{k'<j}A_{k'}\rho(P_{k'})$ along either history,
\begin{equation}
\label{eqn:bias_three_terms}
\mathbb E\big[X^H_r(v;j)\,\big|\,\mathcal F_{-r}\big]=\mathrm I_j-\eta\sum_{k'<j}\big(\mathrm{II}_{k',j}+\mathrm{III}_{k',j}\big),
\end{equation}
where
\begin{align*}
\mathrm I_j&:=\mathbb E\Big[\langle v,\phi_0\rangle_{\bar2}\big(\rho(P^1_j)-\rho(P^0_j)\big)\Big],\\
\mathrm{II}_{k',j}&:=\langle v,\Delta A_{k'}\rangle_{\bar2}\;\mathbb E\big[\rho(P^1_{k'})\rho(P^1_j)\big],\\
\mathrm{III}_{k',j}&:=\langle v,A^0_{k'}\rangle_{\bar2}\;\mathbb E\big[\rho(P^1_{k'})\rho(P^1_j)-\rho(P^0_{k'})\rho(P^0_j)\big].
\end{align*}
% We bound each term by the quantities $e^A_{j-1}(v)$, $e^-_j(\bm Z^0)$, $\mathfrak e^-_j$ and $\mathfrak v^-_j$ of Section~\ref{sec:bias}, which measure the difference between the two histories of the pair. 
Since $\rho$ does not depend on the feature $Q$, part~3 of Lemma~\ref{lemma:Lip_in_E} applies, and the adjoint differences only enter at the steps before $j$.
\begin{itemize}
\item \emph{Term $\mathrm I$.} This is part~1 of Lemma~\ref{lemma:Lip_in_E}, with $u=(0,v)$, $R=C_{\mathrm{lim}}\vee1$ and $f=\rho$:
\[
|\mathrm I_j|\le C\Big[\tilde{\mathfrak d}^{A,-}_j(v)+\sqrt{\tfrac{\mathbb V}D}\,\tilde{\mathfrak v}^-_j((0,v))\Big]+C\Big[\tilde{\mathfrak d}^-_j+(\tilde{\mathfrak e}^-_j)^2+\sqrt{\tfrac{\mathbb V}D}\,\|\bm Z^0\|_{\bar\infty}\,\tilde{\mathfrak e}^-_j\Big].
\]
\item \emph{Term $\mathrm{II}$.} The expectation is at most $C$, by Lemma~\ref{lemma:PQ_psi_2} and the linear growth of $\rho$, and the prefactor is a projection of the adjoint difference at a step $k'<j$, so that
\[
\eta\sum_{k'<j}|\mathrm{II}_{k',j}|\le C\eta\sum_{k'<j}|\langle v,\Delta A_{k'}\rangle_{\bar2}|=C\,\tilde{\mathfrak d}^{A,-}_j(v).
\]
\item \emph{Term $\mathrm{III}$.} The prefactor is at most $C_B\|v\|_{\bar2}$. Writing $\Delta\rho_{k}:=\rho(P^1_{k})-\rho(P^0_{k})$,
\[
\rho(P^1_{k'})\rho(P^1_j)-\rho(P^0_{k'})\rho(P^0_j)
=\rho(P^0_{k'})\,\Delta\rho_j+\rho(P^0_j)\,\Delta\rho_{k'}+\Delta\rho_{k'}\,\Delta\rho_j .
\]
The expectations of the first two products are bounded by parts~2 and 3 of Lemma~\ref{lemma:Lip_in_E}, with $g=\rho$. The expectation of the third is at most $C(\tilde{\mathfrak e}^-_j)^2$, by the Cauchy--Schwarz inequality, Assumption \ref{assm:rho} and Lemma~\ref{lemma:PQ_Lip_psi2}. Hence
\[
|\mathrm{III}_{k',j}|\le C\Big[\tilde{\mathfrak d}^-_j+(\tilde{\mathfrak e}^-_j)^2+\sqrt{\tfrac{\mathbb V}D}\,\|\bm Z^0\|_{\bar\infty}\,\tilde{\mathfrak e}^-_j\Big].
\]
\end{itemize}

\paragraph{From the pair of histories to the errors of the discrete model.} The three bounds above involve the difference between $\widetilde{\bm Z}_r$ and $\bm Z^\infty_r$, through the tilded quantities. We now pass through the history $\bm z_r$ of the discrete model, at a step $j\le K$, where, on $\mathcal E^{\mathrm{op}}_\delta$, $\widetilde{\bm Z}_r$ is the leave-one-layer-out history. Each of the following costs one term of the triangle inequality.
\begin{align*}
\tilde{\mathfrak d}^{A,-}_j(v)&\le\eta\sum_{k'<j}|e^A_{r,k'}(v)|+\eta K\|v\|_{\bar2}\,\Delta^{\mathrm{diag}}_r,
&&\text{(projection onto $v$)}\\
\tilde{\mathfrak d}^-_j&\le G^H_j(r-1)+G^A_{j-1}(r)+C_{\mathrm{lim}}\,\Delta^{\mathrm{diag}}_r,
&&\text{(projections onto the limiting history)}\\
\tilde{\mathfrak e}^-_j&\le\Delta^{\mathrm{diag}}_r+\varepsilon^{\bar2}_K,
&&\text{($\bar2$-error)}\\
\|\bm Z^0\|_{\bar\infty}&\le C_\infty\sqrt{\mathfrak l^\infty_\delta}.
&&\text{(Lemma~\ref{lemma:InfNorm})}
\end{align*}
Here $G^H$ and $G^A$ collect the projections of the errors onto the directions of the limiting model according to
\begin{equation}
\label{eqn:G_def}
G^H_k(r):=\max_{ k''\le K}\max_{r'\le r}\max_{k'\le k}
\big|e^H_{r',k'}\big(H_{\frac rL,k''}\big)\big|,
\quad
G^A_k(r):=\max_{ k''\le K}\max_{r'\ge r}\max_{k'\le k}
\big|e^A_{r',k'}\big(A_{\frac rL,k''}\big)\big|.
\end{equation}
The terms that are quadratic in the errors or involving the leave-one-layer-out distances are collected as
\begin{equation}
\label{eqn:varrho_def}
\varrho_r:=\Delta^{\mathrm{diag}}_r+(\Delta^{\mathrm{diag}}_r)^2+(\varepsilon^{\bar2}_K)^2
+C_\infty\sqrt{\tfrac{\mathbb V\mathfrak l^\infty_\delta}D}\big(\varepsilon^{\bar2}_K+\Delta^{\mathrm{diag}}_r\big),
\qquad
\bar\varrho:=\frac1L\sum_{r=1}^L\varrho_r,
\end{equation}
The remaining term, from Term~$\mathrm I$, vanishes when $\mathbb V=0$. Since $\tilde{\mathfrak v}^-_k((0,v))\le\tilde{\mathfrak v}_k((v,v))$, it is at most $\sqrt{\mathbb V/D}\,\tilde{\mathfrak v}_{r,k}(v)$, where $\tilde{\mathfrak v}_{r,k}(v):=\tilde{\mathfrak v}_k((v,v))$ is taken for the pair $(\bm Z^\infty_r,\widetilde{\bm Z}_r)$ of layer $r$. Substituting the four inequalities into the bounds on $\mathrm I$, $\mathrm{II}$ and $\mathrm{III}$, and summing \eqref{eqn:bias_three_terms} over $k'<j$, we obtain for some constant $C_b$,
\begin{equation}
\label{eqn:bias_hidden}
\big|\mathbb E[X^H_r(v;k)|\mathcal F_{-r}]\big|
\le C_b\Big[\eta\sum_{k'<k}|e^A_{r,k'}(v)|+G^H_k(r-1)+G^A_{k-1}(r)+\varrho_r+\sqrt{\tfrac{\mathbb V}D}\,\tilde{\mathfrak v}_{r,k}(v)\Big].
\end{equation}

\paragraph{The adjoint variables.} The argument is the same, with $\psi_j=\psi_0-\eta\sum_{k'<j}H_{k'}\,g(P_{k'},Q_{k'})$, where $g(p,q):=\rho'(p)\mathrm{cl}(q)$, and with $f(p,q):=\rho'(p)q$ in place of $\rho$:
\[
\mathbb E\big[X^A_r(v;j)\,\big|\,\mathcal F_{-r}\big]=\mathrm I^A_j-\eta\sum_{k'<j}\big(\mathrm{II}^A_{k',j}+\mathrm{III}^A_{k',j}\big),
\]
where, writing $f^i_k:=f(P^i_k,Q^i_k)$ and $g^i_k:=g(P^i_k,Q^i_k)$,
\begin{align*}
\mathrm I^A_j&:=\mathbb E\big[\langle v,\psi_0\rangle_{\bar2}\,(f^1_j-f^0_j)\big],\\
\mathrm{II}^A_{k',j}&:=\langle v,\Delta H_{k'}\rangle_{\bar2}\;\mathbb E\big[g^1_{k'}f^1_j\big],\\
\mathrm{III}^A_{k',j}&:=\langle v,H^0_{k'}\rangle_{\bar2}\;\mathbb E\big[g^1_{k'}f^1_j-g^0_{k'}f^0_j\big].
\end{align*}
Arguing as for the hidden states, with parts~1 and~2 of Lemma~\ref{lemma:Lip_in_E} as stated, rather than in the form of part~3, since $f$ and $g$ depend on $Q$, the same passage through $\bm z_r$ yields
\begin{align}
\label{eqn:bias_adjoint}
\big|\mathbb E[X^A_r(v;k)|\mathcal F_{-r}]\big|
&\le C_b\Big[|e^H_{r-1,k}(v)|+\eta\sum_{k'<k}|e^H_{r-1,k'}(v)|\Big]\\&\qquad +C_b\Big[G^H_k(r-1)+G^A_k(r)+\varrho_r+\sqrt{\tfrac{\mathbb V}D}\,\tilde{\mathfrak v}_{r,k}(v)\Big].\notag
\end{align}

\paragraph{The Stein-kernel term for a single direction.} By the definition \eqref{eqn:coordinatewise_error} and \eqref{eqn:B_DeltaZ},
\begin{equation}
\label{eqn:Stein_depth}
\sqrt{\tfrac{\mathbb V}D}\,\tilde{\mathfrak v}_{r,k}(v)\le2(1+\eta K)\,\sqrt{\tfrac{\mathbb V}D}\,\|v\|_{\bar\infty}\,\Delta^Z_r .
\end{equation}
For a direction $v$ of the limiting model, $\|v\|_{\bar\infty}\le C_\infty\sqrt{\mathfrak l^\infty_\delta}$ by Lemma~\ref{lemma:InfNorm}, and thus this term is at most $C\varrho_r$. For the output directions of Step~3 we do not bound the coordinates of each direction individually (see \eqref{eqn:Stein_output}).

\subsubsection*{Step 3: The output directions}
Let $\Delta_k:=y_{k}-Y_{k}$ be the error of the output at the step $k$. The output embeddings of the two models differ, so we split
\[
\Delta_k=\Delta^h_k+\Delta^W_k,
\qquad
\Delta^h_k:=\frac1D(W^\infty_{\mathrm{out},k})^\top\big(h_{L,k}-H_{1,k}\big),
\qquad
\Delta^W_k:=\frac1D\big(W_{\mathrm{out},k}-W^\infty_{\mathrm{out},k}\big)^\top h_{L,k} .
\]

\paragraph{The Eigenvectors.} Since the loss is $C^2$ and convex (Assumption~\ref{assm:Obj}), its Hessian is symmetric and positive semi-definite. For each deterministic step $k\le K$, let
\[
\nabla^2\ell(Y_k)=\sum_{j=1}^{\mathfrak r_\ell}\lambda_{j,k}\,u_{j,k}u_{j,k}^\top
\]
be its spectral decomposition at the output of the limiting model, with $\lambda_{j,k}>0$, where $\mathfrak r_\ell$ is the rank of the Hessian and the $u_{j,k}\in\R^{d_{\mathrm{out}}}$ are orthonormal. Since $\nabla^2\ell(Y_k)$ is an $\mathcal F_W$-measurable positive semidefinite matrix, its eigenvalues and an orthonormal eigenbasis can be chosen $\mathcal F_W$-measurably \cite[Lemma~31]{page2019ivanov}. The rank does not depend on $k$, since, by \eqref{eqn:Hessian_comparable}, all the Hessians on $\mathcal Y$ share the same kernel. If $\mathrm{tr}\,\nabla^2\ell(Y_k)=0$, there are no output directions, and we set $\mathcal Q_k:=0$. The \emph{output directions} and their weights are 
\begin{equation}
\label{eqn:output_directions}
\omega_{j,k}:=\frac{W^\infty_{\mathrm{out},k}u_{j,k}}{\|W^\infty_{\mathrm{out},k}u_{j,k}\|_{\bar2}},\qquad
\alpha_{j,k}:=\frac{\lambda_{j,k}\,\|W^\infty_{\mathrm{out},k}u_{j,k}\|_{\bar2}^2}{\mathrm{tr}\,\nabla^2\ell(Y_k)},
\end{equation}
with $\omega_{j,k}:=0$ and $\alpha_{j,k}:=0$ if $W^\infty_{\mathrm{out},k}u_{j,k}=0$. They are $\mathcal F_W$-measurable, $\|\omega_{j,k}\|_{\bar2}\le1$, and, by \eqref{eqn:Hessian_comparable} and Lemma~\ref{lemma:apriori},
\begin{equation}\label{eqn:alpha_j_sum}
    \sum_j\alpha_{j,k}=\frac{\|W^\infty_{\mathrm{out},k}\nabla^2\ell(Y_k)^{1/2}\|_F^2}{D\,\mathrm{tr}\,\nabla^2\ell(Y_k)}\le\frac{e^{2R_*}\mathfrak f(W^\infty_{\mathrm{out},k})^2}{\tau_\star}\le\frac{C_{\mathscr H}C_{\mathfrak f}'^2}{\tau_\star}=:C_\omega^2 .
\end{equation}
Further define
\begin{equation}\label{eqn:mathcal Q}
   \mathcal Q_k:=\Big(\sum_{j=1}^{\mathfrak{r}_\ell}\alpha_{j,k}\,\big|e^H_{L,k}(\omega_{j,k})\big|^2\Big)^{1/2},
\end{equation}
so that
\begin{align}
\label{eqn:Hessian_projections}
(\Delta^h_k)^\top\nabla^2\ell(Y_k)\Delta^h_k
&=\sum^{\mathfrak{r}_\ell}_{j=1}\lambda_{j,k}\langle u_{j,k},\Delta^h_k\rangle^2
=\mathrm{tr}\,\nabla^2\ell(Y_k)\,\mathcal Q_k^2,
\end{align}
as $\langle u_{j,k},\Delta^h_k\rangle=\|W^\infty_{\mathrm{out},k}u_{j,k}\|_{\bar2}\,e^H_{L,k}(\omega_{j,k})$. Since the Hessian is PSD, the induced seminorm satisfies the triangle inequality. This implies that
\begin{align}\label{eqn:Dk_split}
\mathscr D_k:=\big(\Delta_k^\top\nabla^2\ell(Y_k)\Delta_k\big)^{1/2}
&\le\big(\mathrm{tr}\,\nabla^2\ell(Y_k)\big)^{1/2}\mathcal Q_k+|\Delta^W_k|_{\mathscr H(Y_k)}\\&\le e^{R_*/2}\big(\sqrt{\tau_\star}\,\mathcal Q_k+|\Delta^W_k|_{\mathscr H_\star}\big)\le C\big(\mathcal Q_k+|\Delta^W_k|_{\mathscr H_\star}\big),\notag
\end{align}
where the second inequality is \eqref{eqn:Hessian_comparable}. By Lemma~\ref{lemma:C1}, $y_k$ and $Y_k$ lie in $\mathcal Y$. Lemma~\ref{lemma:grad_metric} with $\bar y=Y_k$ gives the first bound below, and the second follows analogously using \eqref{eqn:grad_factor} and $|\Delta_k|_{\mathscr H_\star}\le e^{R_*/2}\mathscr D_k$ by \eqref{eqn:Hessian_comparable}, so that
\begin{equation}
\label{eqn:Bregman_projections}
d_\ell(y_k,Y_k)\le\sqrt{C_{\mathscr H}}\,\mathscr D_k,
\qquad
|\nabla\ell(y_k)-\nabla\ell(Y_k)|_{\mathscr H_\star}\le e^{3R_*/2}\lambda_\star\,\mathscr D_k .
\end{equation}

\paragraph{The output embeddings.} By \eqref{eqn:embedding_updates} for both models, and writing $e_{k'}:=h_{L,k'}-H_{1,k'}$
\begin{equation}\label{eqn:DeltaWout}
W_{\mathrm{out},k}-W^\infty_{\mathrm{out},k}=-\eta_{\mathrm{out}}\sum_{k'<k}\Big(e_{k'}\,\nabla\ell(y_{k'})^\top+H_{1,k'}\big(\nabla\ell(y_{k'})-\nabla\ell(Y_{k'})\big)^\top\Big).
\end{equation}
Therefore,
\[
\Delta^W_k=-\eta_{\mathrm{out}}\sum_{k'<k}\Big(\langle e_{k'},h_{L,k}\rangle_{\bar2}\,\nabla\ell(y_{k'})+\langle H_{1,k'},h_{L,k}\rangle_{\bar2}\big(\nabla\ell(y_{k'})-\nabla\ell(Y_{k'})\big)\Big).
\]
We bound the four factors in turn. Since $h_{L,k}=H_{1,k}+e_k$, the first inner product is bounded by $e^H_{L,k'}(H_{1,k})+(\varepsilon^{\bar2}_K)^2$. The third is at most $C_{\mathrm{lim}}C_B$. On $\mathcal Y$, $|\nabla\ell(y)|_{\mathscr H_\star}\le\sqrt{\lambda_\star}\,G_\ell$, and the fourth is bounded according to \eqref{eqn:Bregman_projections}. Hence, since $\eta_{\mathrm{out}}\sqrt{\lambda_\star}G_\ell$ and $\eta_{\mathrm{out}}\lambda_\star$ are at most $C\tilde\eta_{\mathrm{out}}$, with $C$ depending only on $C_{\mathrm{lim}}$, $C_B$, $R_*$ and $S_*$,
\begin{equation}\label{eqn:DeltaW_bound}
|\Delta^W_k|_{\mathscr H_\star}\le C\tilde\eta_{\mathrm{out}}\sum_{k'<k}\Big(|e^H_{L,k'}(H_{1,k})|+(\varepsilon^{\bar2}_K)^2+\mathscr D_{k'}\Big) .
\end{equation}
The direction $H_{1,k}$ belongs to the family $\mathcal W_L$ of Step~4.

\paragraph{The terminal condition.} For every $w\in\R^D$, we split
\[
\langle w,a_{L,k}-A_{1,k}\rangle_{\bar2}=\underbrace{\big\langle w,W^\infty_{\mathrm{out},k}\big(\nabla\ell(y_k)-\nabla\ell(Y_k)\big)\big\rangle_{\bar2}}_{=:\,\mathrm T^1_k(w)}+\underbrace{\big\langle w,\big(W_{\mathrm{out},k}-W^\infty_{\mathrm{out},k}\big)\nabla\ell(y_k)\big\rangle_{\bar2}}_{=:\,\mathrm T^2_k(w)}.
\]
By Lemma~\ref{lemma:grad_metric}, with $y=y_k$ and $y'=Y_k$, which lie in $\mathcal Y$, with $\bar y=Y_k$ and the direction $v:=\frac1D(W^\infty_{\mathrm{out},k})^\top w$,
\begin{equation}\label{eqn:1D_proj_DeltaA_terminal}
|\mathrm T^1_k(w)|=\big|v^\top\big(\nabla\ell(y_k)-\nabla\ell(Y_k)\big)\big|\le C_{\mathscr H}\,|v|_{\mathscr H(Y_k)}\,\mathscr D_k\le C_{\mathscr H}^{3/2}C'_{\mathfrak f}\,\|w\|_{\bar2}\,\mathscr D_k,
\end{equation}
since $|v|_{\mathscr H(Y_k)}\le\sqrt{C_{\mathscr H}}\,\mathfrak f(W^\infty_{\mathrm{out},k})\|w\|_{\bar2}$, by \eqref{eqn:Hessian_comparable} and Lemma~\ref{lemma:C1}.
In $\mathrm T^2_k(w)$, the output embeddings are projected onto $w$ from the left. By \eqref{eqn:DeltaWout}, the bound $G_\ell$ on the loss gradient, and Lemma~\ref{lemma:grad_metric} with $\bar y=Y_{k'}$ and direction $\nabla\ell(y_k)$, we obtain
\[
|\mathrm T^2_k(w)|\le\eta_{\mathrm{out}}G_\ell\sum_{k'<k}\Big(G_\ell|e^H_{L,k'}(w)|+C_{\mathrm{lim}}C_{\mathscr H}e^{R_*/2}\sqrt{\lambda_\star}\,\|w\|_{\bar2}\,\mathscr D_{k'}\Big),
\]
where $|\nabla\ell(y_k)|_{\mathscr H(Y_{k'})}\le e^{R_*/2}\sqrt{\lambda_\star}G_\ell$. Altogether, with \eqref{eqn:Dk_split},
\begin{equation}
\label{eqn:terminal_projections}
|\langle w,a_{L,k}-A_{1,k}\rangle_{\bar 2}|
\le C\|w\|_{\bar2}\max_{k'\le k}\big(\mathcal Q_{k'}+|\Delta^W_{k'}|_{\mathscr H_\star}\big)+\eta_{\mathrm{out}} G_\ell^2\sum_{k'<k}|e^H_{L,k'}(w)| .
\end{equation}

\paragraph{The Stein-kernel term of the output directions.} Fix a deterministic step $k'\le K$, and let $W_i\in\R^{d_{\mathrm{out}}}$ be the $i$-th row of $W^\infty_{\mathrm{out},k'}$, so that
\[
\sum_j\alpha_{j,k'}(\omega_{j,k'})_i^2=\frac{W_i^\top\nabla^2\ell(Y_{k'})W_i}{\mathrm{tr}\,\nabla^2\ell(Y_{k'})}\le C_\infty^2\,\mathfrak l^\infty_\delta ,
\]
where the final inequality results from Lemma~\ref{lemma:InfNorm}. Thus, for every $x\in\R^D$,
\[
\Big(\sum_j\alpha_{j,k'}\,\|\omega_{j,k'}\odot x\|_{\bar2}^2\Big)^{1/2}\le C_\infty\sqrt{\mathfrak l^\infty_\delta}\,\|x\|_{\bar2} .
\]
The quantity $\tilde{\mathfrak v}_k$ in \eqref{eqn:coordinatewise_error} is a sum of such norms, with $x=\Delta H_{k''}$ or $x=\Delta A_{k''}$ as in Step~2, and with total weight at most $2(1+\eta K)$. Hence, by Minkowski's inequality in the $\alpha$-weighted $\ell^2$-norm over $j$ and \eqref{eqn:B_DeltaZ},
\begin{equation}
\label{eqn:Stein_output}
\max_{k'\le K}\Big(\sum_j\alpha_{j,k'}\,\tfrac{\mathbb V}D\,\tilde{\mathfrak v}_{r,k}(\omega_{j,k'})^2\Big)^{1/2}
\le2(1+\eta K)\,C_\infty\sqrt{\tfrac{\mathbb V\mathfrak l^\infty_\delta}D}\,\Delta^Z_r\le C\varrho_r .
\end{equation}

\subsubsection*{Step 4: Closing the induction}
\paragraph{The directions.} The bias bounds \eqref{eqn:bias_hidden} and \eqref{eqn:bias_adjoint} involve projections onto the limiting hidden states and adjoint variables, and the terminal condition \eqref{eqn:terminal_projections} involves projections onto the output directions. Therefore, to close the induction over the training steps, we only need to consider these directions. A direction is \emph{admissible} if it belongs to one of the families
\begin{equation}\label{eqn:depth_directions}
\mathcal W_r=\big\{H_{\frac rL,k},\,A_{\frac rL,k}:0\le k\le K\big\},\ 0\le r\le L,
\qquad
\big\{\omega_{j,k}:j\le\mathfrak r_\ell,\ 0\le k\le K\big\}.
\end{equation}
Every admissible $w$ is $\mathcal F_W$-measurable and satisfies $\|w\|_{\bar2}\le C_{\mathrm{lim}}\vee1$, so that the bias bounds \eqref{eqn:bias_hidden} and \eqref{eqn:bias_adjoint} apply with $v=w$.

\paragraph{The remainder, frozen at the end of training.} For an admissible direction $w$ and a step $k$, let
\begin{gather}
\mathscr R(w):=\max_{0\le l\le L}\,\max_{k\le K}\Big(\big|R^H_{l,k}(w)\big|\vee\big|R^A_{l,k}(w)\big|\Big),
\qquad
\mathscr S_k(w):=\frac1L\sum_{r=1}^L\sqrt{\tfrac{\mathbb V}D}\,\tilde{\mathfrak v}_{r,k}(w),\notag\\
\mathscr L^\star_k(w):=\mathscr R(w)+\mathscr S_k(w)+\bar\eta\sum_{k'<k}\mathscr S_{k'}(w),\label{s4:L_def}
\end{gather}
with the remainders of \eqref{eqn:five_terms_H} and \eqref{eqn:five_terms_A} and the Stein-kernel term of Step~2. Further define
\begin{gather}
\mathscr L^\star(\mathcal W_r):=\max_{w\in\mathcal W_r}\max_{k\le K}\mathscr L^\star_k(w),
\qquad
\mathcal O^\star:=\max_{k,k'\le K}\Big(\sum_j\alpha_{j,k'}\,\mathscr L^\star_k(\omega_{j,k'})^2\Big)^{1/2},\notag\\
\bar{\mathscr L}^\star:=\frac1{L+1}\sum_{r=0}^L\mathscr L^\star(\mathcal W_r)+\mathscr L^\star(\mathcal W_L)+\mathcal O^\star+\bar\varrho .\label{s4:Lbar_def}
\end{gather}
These quantities are independent of the training step. By their definitions, for every admissible $w$ and $k\le K$,
\begin{equation}\label{s4:volterra}
\bar\eta\sum_{k'<k}\mathscr L^\star_{k'}(w)
\le(1+\bar\eta K)\mathscr L^\star_k(w).
\end{equation}
The time-independent remainder contributes at most $\bar\eta K\mathscr R(w)$, while the nested sums over $\mathscr S_{k'}(w)$ are bounded by $(1+\bar\eta K)$ times their sum up to step $k-1$. We shall also repeatedly use the following immediate consequences of the definition of $\bar{\mathscr L}^\star$:
\begin{equation}\label{s4:avg}
\frac1L\sum_{r=0}^L\mathscr L^\star(\mathcal W_r)\le2\bar{\mathscr L}^\star,
\qquad
\bar\varrho\le\bar{\mathscr L}^\star,
\qquad
\mathscr L^\star(\mathcal W_L)\vee\mathcal O^\star\le\bar{\mathscr L}^\star.
\end{equation}
\paragraph{The two recursions.} We insert the bias bounds \eqref{eqn:bias_hidden} and \eqref{eqn:bias_adjoint}, and the input condition \eqref{eqn:input_projection}, into \eqref{eqn:five_terms_H} and \eqref{eqn:five_terms_A}. The Stein-kernel terms are non-negative, so their average over any range of layers is at most $\mathscr S_k(w)$, by \eqref{s4:L_def}. For every admissible $w$, layer $l$ and step $k$, this gives
\begin{align}
|e^H_{l,k}(w)|&\le(1+C_b)\mathscr L^\star_k(w)+\eta_{\mathrm{in}}|x|^2\sum_{k'<k}|e^A_{0,k'}(w)|\notag\\
&\quad+\frac{C_b}L\sum_{r=1}^{l}\Big[\eta\sum_{k'<k}|e^A_{r,k'}(w)|+G^H_k(r-1)+G^A_{k-1}(r)+\varrho_r\Big],\label{s4:H}\\
|e^A_{l,k}(w)|&\le\big|\langle w,a_{L,k}-A_{1,k}\rangle_{\bar2}\big|+(1+C_b)\mathscr L^\star_k(w)\notag\\
&\quad+\frac{C_b}L\sum_{r=l+1}^{L}\Big[|e^H_{r-1,k}(w)|+\eta\sum_{k'<k}|e^H_{r-1,k'}(w)|+G^H_k(r-1)+G^A_k(r)+\varrho_r\Big].\label{s4:A}
\end{align}
At the current step $k$, the hidden-state error at layer $l$ depends only on $G^H_k$ at the layers below $l$. The adjoint error at layer $l$ depends on the hidden-state errors at the step $k$, on the terminal condition, and on $G^A_k$ at the layers above $l$. All other terms are errors at the earlier steps.

\paragraph{The Claim.} For a step $k$ and an admissible direction $w$, let
\[
E_k(w):=\max_{0\le l\le L}\Big(|e^H_{l,k}(w)|\vee|e^A_{l,k}(w)|\Big),
\qquad
\mathscr{P}_k:=\max_{k'\le k}\big(\mathcal Q_{k'}+|\Delta^W_{k'}|_{\mathscr H_\star}\big).
\]
We prove that there is a constant $\mathsf C\ge1$, which we write in sans-serif to distinguish it from the generic constants $C$ and $C'$, independent of $L,M,D$, $\delta$ and $k$, such that, for every $0\le k\le K$ and every admissible direction $w$,
\begin{equation}\label{s4:claim}
E_k(w)\le \mathsf C^{k+1}\big(\mathscr L^\star_k(w)+\bar{\mathscr L}^\star\big),
\qquad
\mathscr{P}_k\le \mathsf C^{k+1}\,\bar{\mathscr L}^\star .
\end{equation}
Since $G^H_k(r)$ and $G^A_k(r)$ of \eqref{eqn:G_def} are maxima of $|e^H_{r',k'}(w)|$ and $|e^A_{r',k'}(w)|$ over suitable layers $r'$, steps $k'\le k$ and directions $w\in\mathcal W_r$, the first bound of \eqref{s4:claim} implies
\begin{equation}\label{s4:G}
G^H_k(r)\vee G^A_k(r)\le \mathsf C^{k+1}\big(\mathscr L^\star(\mathcal W_r)+\bar{\mathscr L}^\star\big),\qquad 0\le r\le L .
\end{equation}

\paragraph{The Induction.} Fix $k$, and assume \eqref{s4:claim} at the steps $k'<k$. At $k=0$, the induction hypothesis is vacuous: all sums over $k'<k$ in \eqref{s4:H}--\eqref{s4:A} vanish, $G^A_{-1}:=0$ and $\Delta^W_0=0$. Thus $\mathscr P_{-1}:=0$, and the three stages below establish \eqref{s4:claim} at $k=0$. By \eqref{s4:claim} and \eqref{s4:G} at the steps $k'<k$, \eqref{s4:volterra} and \eqref{s4:avg}, the terms at the earlier steps satisfy, for every admissible $w$ and layer $r$,
\begin{gather}
\bar\eta\sum_{k'<k}\Big(|e^H_{r,k'}(w)|+|e^A_{r,k'}(w)|\Big)\le2(1+\bar\eta K)\,\mathsf C^k\big(\mathscr L^\star_k(w)+\bar{\mathscr L}^\star\big),\notag\\
\frac1L\sum_{r=1}^LG^A_{k-1}(r)\le3 \mathsf C^k\bar{\mathscr L}^\star,
\qquad
\mathscr P_{k-1}\le \mathsf C^k \bar{\mathscr L}^\star .\label{s4:past}
\end{gather}
Below, $C'$ denotes a constant, which may change from line to line. It may depend on $C_b$, $C_{\mathrm{lim}}$, $C_\omega$, $\bar\eta K$ and the constants of Step~3, but not on $k$ or $\mathsf C$. 

\emph{Stage 1.} By \eqref{s4:H}, \eqref{s4:past} and $\bar\varrho\le\bar{\mathscr L}^\star$,
\begin{equation}\label{s4:H1}
|e^H_{l,k}(w)|\le C'\mathsf C^k \big(\mathscr L^\star_k(w)+\bar{\mathscr L}^\star\big)+\frac{C_b}L\sum_{r=0}^{l-1}G^H_k(r).
\end{equation}
The right-hand side is non-decreasing in $l$. Take $w\in\mathcal W_l$, so that $\mathscr L^\star_k(w)\le\mathscr L^\star(\mathcal W_l)$ by \eqref{s4:Lbar_def}, and the maximum over the layers $r'\le l$. The RHS of the above estimate also bounds $e^H_{r,k'}(w)$ at step $k'<k$  by the hypothesis \eqref{s4:G} at the step $k-1$ as $C'\ge 1$. This gives
\[
G^H_k(l)\le C'\mathsf C^k\big(\mathscr L^\star(\mathcal W_l)+\bar{\mathscr L}^\star\big)+\frac{C_b}L\sum_{r=0}^{l-1}G^H_k(r),\qquad 0\le l\le L .
\]
By the discrete Gr\"onwall inequality together with \eqref{s4:avg}, we obtain
\begin{equation}\label{s4:GH}
G^H_k(l)\le C'\mathsf C^k\big(\mathscr L^\star(\mathcal W_l)+\bar{\mathscr L}^\star\big),\qquad 0\le l\le L .
\end{equation}
Substituting \eqref{s4:GH} into \eqref{s4:H1} and using \eqref{s4:avg} once more gives, for every admissible $w$,
\begin{equation}\label{s4:Hdone}
\max_{0\le l\le L}|e^H_{l,k}(w)|\le C'\mathsf C^k\big(\mathscr L^\star_k(w)+\bar{\mathscr L}^\star\big).
\end{equation}

\emph{Stage 2.} By \eqref{s4:Hdone} at $l=L$ in the output directions $\omega_{j,k}$, Minkowski's inequality in the $\alpha$-weighted $\ell^2$-norm over $j$, and \eqref{eqn:alpha_j_sum},
\[
\mathcal Q_k\le C'\mathsf C^k\Big(\Big(\sum_j\alpha_{j,k}\mathscr L^\star_k(\omega_{j,k})^2\Big)^{1/2}+C_\omega\bar{\mathscr L}^\star\Big)\le C'\mathsf C^k\big(\mathcal O^\star+C_\omega\bar{\mathscr L}^\star\big)\le C'\mathsf C^k\bar{\mathscr L}^\star .
\]
The bound \eqref{eqn:DeltaW_bound} on $\Delta^W_k$ only involves the earlier steps. Its first term is a hidden-state error in the direction $H_{1,k}\in\mathcal W_L$, at most $2\mathsf C^k\bar{\mathscr L}^\star$ by \eqref{s4:claim} at the steps $k'<k$, where $\mathscr L^\star_{k'}(H_{1,k})\le\mathscr L^\star(\mathcal W_L)$ by \eqref{s4:avg}. Its second term satisfies $(\varepsilon^{\bar2}_K)^2\le\bar\varrho\le\bar{\mathscr L}^\star$, since $(\varepsilon^{\bar2}_K)^2$ is a term of every $\varrho_r$. Its third satisfies 
\[\mathscr D_{k'}\le C'(\mathcal Q_{k'}+|\Delta^W_{k'}|_{\mathscr H_\star})\le C'\mathsf C^{k'+1}\bar{\mathscr L}^\star,\]
by \eqref{eqn:Dk_split} and \eqref{s4:past}. Summing over $k'<k$,
\[
|\Delta^W_k|_{\mathscr H_\star}\le C'\mathsf C^k\bar{\mathscr L}^\star,
\qquad\text{and so}\qquad
\mathscr{P}_k\le C'\mathsf C^k\bar{\mathscr L}^\star .
\]
By \eqref{eqn:terminal_projections}, $\|w\|_{\bar2}\le C_{\mathrm{lim}}\vee1$, \eqref{s4:volterra} and \eqref{s4:past}, for every admissible $w$,
\begin{equation}\label{s4:T}
\big|\langle w,a_{L,k}-A_{1,k}\rangle_{\bar2}\big|\le C'\mathsf C^k\big(\mathscr L^\star_k(w)+\bar{\mathscr L}^\star\big).
\end{equation}

\emph{Stage 3: The adjoint variables.} In \eqref{s4:A}, we bound:
\begin{itemize}
\item the terminal condition by \eqref{s4:T};
\item the hidden-state errors in the direction $w$ by \eqref{s4:Hdone} at the step $k$ and \eqref{s4:past} at the earlier steps;
\item the average of $G^H_k(r-1)$ by \eqref{s4:GH} and \eqref{s4:avg};
\item the average of $\varrho_r$ by $\bar\varrho\le\bar{\mathscr L}^\star$.
\end{itemize}
Hence
\[
|e^A_{l,k}(w)|\le C'\mathsf C^k\big(\mathscr L^\star_k(w)+\bar{\mathscr L}^\star\big)+\frac{C_b}L\sum_{r=l+1}^{L}G^A_k(r).
\]
The right-hand side is non-increasing in $l$. As in Stage~1, take the maximum over $w\in\mathcal W_l$ and the layers $r'\ge l$, and use \eqref{s4:G} at the step $k-1$ for the earlier steps. The discrete Gr\"onwall inequality, now downwards in depth, and \eqref{s4:avg} then give
\[
G^A_k(l)\le C'\mathsf C^k\big(\mathscr L^\star(\mathcal W_l)+\bar{\mathscr L}^\star\big),
\qquad\text{and so}\qquad
\max_{0\le l\le L}|e^A_{l,k}(w)|\le C'\mathsf C^k\big(\mathscr L^\star_k(w)+\bar{\mathscr L}^\star\big).
\]
\emph{Conclusion.} Stages~1 and~3 bound the errors at the step $k$ in every admissible direction, and Stage~2 bounds $\mathscr P_k$. Hence \eqref{s4:claim} holds at the step $k$ for any $\mathsf C\ge C'$. This completes the induction. In particular,
\begin{equation}\label{s4:Q_final}
\max_{k\le K}\big(\mathcal Q_k+|\Delta^W_k|_{\mathscr H_\star}\big)\le \mathsf C^{K+1}\,\bar{\mathscr L}^\star.
\end{equation}

\subsubsection*{Step 5: The rates}
By \eqref{eqn:Bregman_projections}, \eqref{eqn:Dk_split} and \eqref{s4:Q_final},
\begin{equation}
\label{eqn:S_by_Lbar}
\max_{k\le K} d_\ell(y_k,Y_k)
\le C\,\mathsf C^{K+1}\,\bar{\mathscr L}^\star ,
\end{equation}
so it remains to prove that
\begin{equation}
\label{eqn:Lbar_target}
\bar{\mathscr L}^\star \le C\Big[\frac1L+\sqrt{\frac{\mathfrak l_\delta}{LM}}+\frac{D}{LM}\Big]
+C\,C_\infty\sqrt{\mathbb V}\Big[\sqrt{\frac{\mathfrak l^\infty_\delta}{LM}}
+\frac1L\sqrt{\frac{\mathfrak l^\infty_\delta}D}\Big].
\end{equation}

\paragraph{The remainders.} Let $w$ be an admissible direction, so that $\|w\|_{\bar2}\le C_{\mathrm{lim}}\vee1$. Each term of \eqref{eqn:five_terms_H} uniformly in $l$ and $k$ can be bounded in the following manner,
\begin{align}
\label{eqn:RH_bound}
\max_{0\le l\le L}\max_{k\le K}|R^H_{l,k}(w)|\le{}&
\underbrace{\max_{l,k}\big|\big\langle w,\mathcal E^H_{\mathrm{T.D.},l,k}\big\rangle_{\bar2}\big|}_{\le\,C/L\ \text{by \eqref{eqn:B_TD}}}
+\underbrace{\max_{l,k}\big|\big\langle w,\widehat{\mathcal E}^H_{l,k}\big\rangle_{\bar2}\big|}_{\le\,C\varepsilon_\delta\ \text{by \eqref{eqn:B_dec}}}\notag\\
&+\underbrace{\max_{l,k}\big|\big\langle w,\mathcal E^H_{\mathrm{m.c.},l,k}\big\rangle_{\bar2}\big|}_{\le\,C\mathcal T^{LM}_{\delta,1}\ \text{by \eqref{eqn:MC_projection}}}+\underbrace{\max_{l,k}\Big|\frac1L\sum_{r=1}^{l}X^{H,\mathrm{cen}}_r(w;k)\Big|}_{\le\,C(\frac1L+\varepsilon_\delta+\varepsilon^{\bar2}_K)\mathcal T^{M}_{\delta,3/2}\ \text{by \eqref{eqn:Fluc_H_error}}},
\end{align}
where each maximum is over $0\le l\le L$ and $k\le K$. Each of the four bounds holds uniformly in $l$ and $k$, since each of \eqref{eqn:B_TD}, \eqref{eqn:MC_projection}, \eqref{eqn:B_dec} and \eqref{eqn:Fluc_H_error} are maxima over the layers and training steps. The maximum of $|R^A_{l,k}(w)|$ in \eqref{eqn:five_terms_A} satisfies the same bound, by the adjoint versions of the same four equations.

\paragraph{From one direction to $\bar{\mathscr L}^\star$.} The first two bounds in \eqref{eqn:RH_bound} hold on $\mathcal E_\delta$ for all $w$ at once. The last two hold with probability $1-\delta$ for each fixed $w$, with tails of the form required by Lemma~\ref{lemma:LogRemoval_mixed}, conditionally on $\mathcal F_W$. By \eqref{s4:Lbar_def}, the depths enter $\bar{\mathscr L}^\star$ through an average over $r$, and the output directions through an $\alpha$-weighted root mean square. Lemma~\ref{lemma:LogRemoval_mixed}, with uniform weights and $p=1$ over $r$, and with the normalised weight $\alpha_{j,k'}/\sum_{j'}\alpha_{j',k'}$ and $p=2$ over $j$ (Remark~\ref{rem:LogRemoval_weights}), therefore gives the last two bounds of \eqref{eqn:RH_bound} for these averages, after a union bound over the $2(K+1)$ directions of each $\mathcal W_r$ and the $K+1$ steps $k'$. The same bounds hold for the single family $\mathcal W_L$, directly from the tails for each direction and a union bound over its $2(K+1)$ directions. Moreover, $\mathscr L^\star_k(w)$ contains the Stein-kernel part $\mathscr S_k(w)$ and the summation of $\bar\eta\mathscr S_{k'}(w)$ over $k'<k$, which are bounded differently for the two kinds of directions.
\paragraph{}\emph{Depth directions.} For $w\in\bigcup_r\mathcal W_r$, including $\mathcal W_L$, the bound \eqref{eqn:Stein_depth} is uniform in $k$. Since $\|w\|_{\bar\infty}\le C_\infty\sqrt{\mathfrak l^\infty_\delta}$ by Lemma~\ref{lemma:InfNorm}, and $\Delta^Z_r\le\Delta^{\mathrm{diag}}_r+\varepsilon^{\bar2}_K$ by \eqref{eqn:B_DeltaZ}, each summand of $\mathscr S_k(w)$ is at most $C\varrho_r$, by \eqref{eqn:varrho_def}. $\mathscr{S}_k(w)$ averages $\tilde{\mathfrak{v}}_{r,k}(w)$ over the layers, and so $\mathscr S_k(w)\le C\bar\varrho$ for every $k$. Hence the Stein-kernel part of $\mathscr L^\star_k(w)$ is at most $C(1+\bar\eta K)\bar\varrho$, uniformly in $k$, and so is its contribution to the first two terms of \eqref{s4:Lbar_def}.
\paragraph{}\emph{Output directions.} The $\bar\infty$-norms of the $\omega_{j,k'}$ are not controlled one direction at a time, and \eqref{eqn:Stein_output} only controls their $\alpha$-weighted root mean square, at a fixed step $k$. This suffices, since the output directions enter $\bar{\mathscr L}^\star$ only through $\mathcal O^\star$, in which the maxima over $k$ and $k'$ are taken outside the root mean square. By Minkowski's inequality in the $\alpha$-weighted $\ell^2$-norm over $j$,
\begin{align*}
\mathcal O^\star\le{}&\max_{k'\le K}\Big(\sum_j\alpha_{j,k'}\,\mathscr R(\omega_{j,k'})^2\Big)^{\frac{1}{2}}+\max_{k,k'\le K}\Big(\sum_j\alpha_{j,k'}\Big(\mathscr S_k(\omega_{j,k'})+\bar\eta\sum_{k''<k}\mathscr S_{k''}(\omega_{j,k'})\Big)^2\Big)^{\frac12}.
\end{align*}
In the first term, the first two bounds of \eqref{eqn:RH_bound} hold for all directions at once, and $\|\omega_{j,k'}\|_{\bar2}$ is $1$ by construction, so they contribute at most $(\sum_j\alpha_{j,k'})^{1/2}\le C_\omega$ times their right-hand sides, by \eqref{eqn:alpha_j_sum}. The last two bounds of \eqref{eqn:RH_bound} hold for each fixed direction, with tails satisfying the hypothesis of Lemma~\ref{lemma:LogRemoval_mixed}. This Lemma can then be applied with $p=2$ and the weights $\alpha_{j,k'}$ to bound the final two terms of \eqref{eqn:RH_bound}. Hence, the first term is bounded by $C_\omega$ times the sum of the terms underbraced on the right-hand side of \eqref{eqn:RH_bound}. For the second term, for fixed $k,k'\le K$, Minkowski's inequality in the $\alpha$-weighted $\ell^2$-norm over $j$, over the average in $r$ and over the sum in $k''$ gives
\begin{align*}
&\Big(\sum_j\alpha_{j,k'}\Big(\mathscr S_k(\omega_{j,k'})+\bar\eta\sum_{k''<k}\mathscr S_{k''}(\omega_{j,k'})\Big)^2\Big)^{1/2}\\
&\qquad\le\frac1L\sum_{r=1}^L\Big[\Big(\sum_j\alpha_{j,k'}\tfrac{\mathbb V}D\,\tilde{\mathfrak v}_{r,k}(\omega_{j,k'})^2\Big)^{1/2}+\bar\eta\sum_{k''<k}\Big(\sum_j\alpha_{j,k'}\tfrac{\mathbb V}D\,\tilde{\mathfrak v}_{r,k''}(\omega_{j,k'})^2\Big)^{1/2}\Big]\\
&\qquad\le C(1+\bar\eta K)\,\bar\varrho,
\end{align*}
where the final inequality uses \eqref{eqn:Stein_output} at each step. Taking the maxima over $k$ and $k'$, we see that the Stein-kernel part contributes at most $C(1+\bar\eta K)\bar\varrho$ to $\mathcal O^\star$.
Finally, by \eqref{eqn:varrho_def}, \eqref{eqn:B_diag} and $\varepsilon_\delta^2\le C\varepsilon_\delta$,
\begin{equation}
\label{eqn:B_varrho}
\bar\varrho\le C\Big(\varepsilon_\delta+(\varepsilon^{\bar2}_K)^2
+C_\infty\sqrt{\tfrac{\mathbb V\mathfrak l^\infty_\delta}D}\,\big(\varepsilon^{\bar2}_K+\varepsilon_\delta\big)\Big).
\end{equation}
Altogether,
\begin{equation}
\label{eqn:Lbar_collected}
\bar{\mathscr L}^\star\le C\Big[\frac1L+\varepsilon_\delta
+\mathcal T^{LM}_{\delta,1}
+\Big(\frac1L+\varepsilon_\delta+\varepsilon^{\bar2}_K\Big)\mathcal T^{M}_{\delta,3/2}
+(\varepsilon^{\bar2}_K)^2
+C_\infty\sqrt{\tfrac{\mathbb V\mathfrak l^\infty_\delta}D}\,\big(\varepsilon^{\bar2}_K+\varepsilon_\delta\big)\Big].
\end{equation}

\paragraph{Simplification.} 
By \eqref{eqn:shape_main},
\[
\mathfrak l_\delta\le 1+\log (K+1)+c_1M^{1/3},
\qquad D\le c_0M,
\]
and hence
\[
\mathcal T^M_{\delta,3/2}
=\sqrt{\frac{\mathfrak l_\delta}{M}}
\left(1+\frac{\mathfrak l_\delta}{\sqrt M}\right)\le C,
\qquad
\frac{D+\mathfrak l_\delta}{LM}\le C,
\qquad
\mathfrak l_\delta\le C D.
\]
Combining these bounds with \eqref{eqn:Leps_bound} and \eqref{eqn:B_eps2} gives
\begin{align*}
\frac1L+\varepsilon_\delta+\mathcal T^{LM}_{\delta,1}
&\le C\left[\frac1L+\sqrt{\frac{\mathfrak l_\delta}{LM}}+\frac{D+\mathfrak l_\delta}{LM}\right],\\
\varepsilon^{\bar2}_K\,\mathcal T^M_{\delta,3/2}
&\le C\left[\frac1L+\sqrt{\frac{\mathfrak l_\delta}{LM}}\right],\\
(\varepsilon^{\bar2}_K)^2
&\le C\left[\frac1{L^2}+\frac{D+\mathfrak l_\delta}{LM}\right],\\
\sqrt{\frac{\mathbb V\mathfrak l^\infty_\delta}{D}}\,
(\varepsilon^{\bar2}_K+\varepsilon_\delta)
&\le C\sqrt{\mathbb V}\left[
\sqrt{\frac{\mathfrak l^\infty_\delta}{LM}}
+\frac1L\sqrt{\frac{\mathfrak l^\infty_\delta}{D}}
+\sqrt{\frac{\mathfrak l_\delta\mathfrak l^\infty_\delta}{DLM}}
\right].
\end{align*}
Finally, since $\mathfrak l_\delta/(LM)\le C$ and $\mathfrak l_\delta\le CD$,
\[
\frac{\mathfrak l_\delta}{LM}
\le C\sqrt{\frac{\mathfrak l_\delta}{LM}},
\qquad
\sqrt{\frac{\mathfrak l_\delta\mathfrak l^\infty_\delta}{DLM}}
\le C\sqrt{\frac{\mathfrak l^\infty_\delta}{LM}}.
\]
Summing the four estimates therefore gives \eqref{eqn:Lbar_target}, which, together with \eqref{eqn:S_by_Lbar}, proves Theorem~\ref{thm:Rates}. The bound holds on the intersection of $\mathcal E_\delta$, $\mathcal E^{\mathrm{op}}_\delta$, the events of Theorem~\ref{thm:bar 2-norm} and of Lemma~\ref{lemma:InfNorm}, and the averaged Monte Carlo and fluctuation events for the two families of directions, whose number depends only on $K$. For the output directions, the events are for the $\alpha_{j,k}$-weighted average over $j\le\mathfrak{r}_\ell$, not a union bound over the $\mathfrak{r}_\ell$ directions, so their number does not depend on $d_{\mathrm{out}}$.

\section{Conclusion \& Future Research}
We have established quantitative convergence rates for the outputs of deep and wide ResNets trained in the MLU parameterisation. For Gaussian initialisation, the outputs converge to their mean-field limits in the loss-induced pseudometric $d_\ell$ at rate
\[
C\left(\frac1L+\sqrt{\frac{\mathfrak l_\delta}{LM}}+\frac{D}{LM}\right),
\]
where $\mathfrak l_\delta=\log(e(K+1)/\delta)$, with no logarithmic dependence on the architectural dimensions. For general initialisations, an additional Stein-kernel correction appears, of order $\sqrt{\log D/(LM)}$ up to the remaining terms of Theorem \ref{thm:MainTheorem}.

\paragraph{}We believe that our analysis could be extended to unclipped gradient descent, though at the cost of a less favourable dependence on the number of training steps and more restrictive conditions on the failure probability. Such an extension is obtained, at slower convergence rates, in \cite{chaintron2026infiniteresnetsgaussianmean}. The importance of controlling the magnitude of parameter updates via clipping in our analysis may reflect what is observed in practice. Large neural networks are often difficult to train with standard SGD, whereas optimisers such as AdamW and Muon \cite{liu2025muon}, which regulate the magnitude of parameter updates, tend to offer more stable training and require less careful hyperparameter tuning \cite{zhao2025deconstructingmakesgoodoptimizer}. Our bounds also extend to finite training datasets (Remark~\ref{rem:multi_data}), with the constants depending on the number of samples as they do on $K$, and an additional logarithmic factor from the union bound over the samples. Another research direction is to obtain estimates uniform over the initial condition, as in \cite{gibson2026uniform}, while retaining our faster convergence rates. Finally, much of the dependence of our constants on the training horizon is through $\eta K$, the natural time scale for gradient flow. However, some additional dependence on $K$ remains, notably the logarithmic factors arising from union bounds over training steps, which prevents our estimates from passing directly to the gradient-flow limit.

\section*{AI Usage Disclosure}
Claude Opus 5.5 was used to accelerate the writing process and to assist in checking the validity of the arguments. All mathematical ideas, arguments and proofs are the Authors’ own, except for a suggestion by Claude during proofreading that led to a simpler approach, replacing the $C^3$ regularity assumption on the loss with $C^2$ regularity and the Hessian stability condition of Assumption \ref{assm:Loss}.

\acks{This work was supported by the Laboratory for AI Security Research (LASR). The views expressed in this paper are those of the authors and do not necessarily reflect the position of LASR or His Majesty’s Government.}

\newpage

\appendix
\section{Notation}\label{sec:Notation}\label{app:notation}
The RMS inner product and norm, and the logarithmic factors $\mathfrak l_\delta$ and $\mathfrak l^\infty_\delta$, at the start of Section \ref{sec:Setup}, and the effective learning rates $\bar\eta$ and $\tilde\eta_{\mathrm{out}}$ in \eqref{eqn:eta_bar}. For $N_{\mathrm{data}}$ training samples, $\delta$ is replaced by $\delta/N_{\mathrm{data}}$ (Remark~\ref{rem:multi_data}).

\paragraph{Norms.}
For $x,y\in\R^D$, we consider the RMS inner product and $\bar p$-norm, for $p\in[1,\infty)$, defined respectively by \[\langle x,y\rangle_{\bar 2}:=D^{-1}x^\top y, \qquad \|x\|^p_{\bar p}:=\frac1D\sum_i|x_i|^p,\]  with the convention $\|x\|_{\bar\infty}:=\max_i|x_i|$, and $\|u\|_{\bar\infty}:=\|u_1\|_{\bar\infty}\vee\|u_2\|_{\bar\infty}$ for $u=(u_1,u_2)\in\R^{2D}$. We write $|\cdot|$ for the Euclidean norm, which is not normalised. We write 
\[
\mathbb B^D_p :=\{ v\in \R^D\colon \|v\|_p\le 1\}
\]
for the unit $p$-norm ball in $\R^D$. The Orlicz (quasi-)norms (see \cite[Definition~2.2]{kuchibhotla2022moving}) are denoted by $\|\cdot\|_{\psi_\alpha}$, and $\|\cdot\|_{\psi_\alpha|\mathcal F}$ denotes the conditional version. For a random variable $\psi$ taking values in $\R^D$, we use the convention
\[
\|\psi\|_{\psi_2} = \sup_{u\in \mathbb B^D_2}\|\langle u,\psi\rangle\|_{\psi_2}.
\]

\paragraph{Conventions.}
We write $a\vee b$ and $a\wedge b$ for the maximum and the minimum, and, for integers $l\le l'$, $[l,l']:=\{l,\ldots,l'\}$ and $(l,l']:=[l+1,l']$. A maximum over the empty set is zero. A function in bold is applied entrywise to a vector, so that $\bm\rho(p):=(\rho(p_m))_{m\le M}$ for $p\in\R^M$, and $\odot$ is the entrywise product. Unless stated otherwise, $C$ denotes a constant independent of $L,M,D$ and $\delta$, which may change from line to line.

\paragraph{$\sigma$-algebras.}
For a layer $r$,  denote $\mathcal F_{-r}$ for the $\sigma$-algebra generated by all initial parameters other than those of layer $r$, and by the initial embeddings. That is,
\[
\mathcal F_{-r} = \sigma \left((\Psi^l_0,\Phi^l_0)_{ l\ne r},W_{\mathrm{in},0},W_{\mathrm{out},0}\right).
\]
We further write $\mathcal F_W:=\sigma(W_{\mathrm{in},0},W_{\mathrm{out},0})$ for the $\sigma$-algebra generated by the initial embeddings. All expectations over the residual parameters below are conditional on $\mathcal F_W$.

\paragraph{Histories and layers.}
We consider the following norms on  a history $\bm Z$: \[
\|\bm Z\|_{\infty,\bar2}:=\max_k\|H_k\|_{\bar2}\vee\|A_k\|_{\bar 2}, \qquad \|\bm Z\|_{\bar\infty}:=\max_k\|H_k\|_{\bar\infty}\vee\|A_k\|_{\bar\infty},
\] 
and say that $\bm Z$ is \emph{bounded by} $B$ if $\|\bm Z\|_{\infty,\bar2}\le B$.
 For layer $r$, whose neurons have initial parameters $(\phi^{rm}_0,\psi^{rm}_0)_{m\le M}$, we write $\phi^{rm}_k(\bm Z),\psi^{rm}_k(\bm Z)$ and $P^{rm}_k(\bm Z),Q^{rm}_k(\bm Z)$ for the parameters and features in a neuron $m$, and write $\Phi^r_k(\bm Z),\Psi^r_k(\bm Z)\in\R^{D\times M}$, $P^r_k(\bm Z),Q^r_k(\bm Z)\in\R^M$ for corresponding whole layer quantities.

\paragraph{Rates.} For $d\ge0$, $N\ge1$ and $\beta\ge1$, define
\[
\mathcal T^{d,N}_{\delta,\beta}:=\sqrt{\frac{d+\mathfrak l_\delta}{N}}+\frac{(d+\mathfrak l_\delta)^\beta}{N},\qquad \mathcal T^{N}_{\delta,\beta}:=\mathcal T^{0,N}_{\delta,\beta},
\]
for the rate of concentration of an average of $N$ independent $\beta^{-1}$-sub-Weibull variables, in a fixed direction ($d=0$) or in the $\bar2$-norm in dimension $d$. Since $\mathfrak l_\delta=\log(e(K+1)/\delta)$, these rates include a union bound over the $K+1$ training steps. This complements the notation $\mathcal K^{D,M,p}_\delta$ of \eqref{eqn:mathcal K} for the $\bar p$-norms of the features.

\section{Probabilistic Tools}\label{app:prob}
\begin{lemma}[Averaging mixed tails]
\label{lemma:LogRemoval_mixed}
Let $\mathcal Z_1,\ldots,\mathcal Z_n$ be random variables, possibly
dependent, and let $c_1,c_2,c_3\ge0$ and $\beta\ge1$ be such that,
for every $\delta\in(0,1]$ and every $j$, with probability at
least $1-\delta$,
\[
|\mathcal Z_j|\le c_1+c_2\sqrt{\log(e/\delta)}+c_3\log(e/\delta)^\beta .
\]
Then there is $C_\beta$, depending only on $\beta$, such that,
for every $p\ge1$ and all weights $\gamma_j\ge0$ with
$\sum_j\gamma_j=1$, with probability at least $1-\delta$,
\[
\Big(\sum_j\gamma_j|\mathcal Z_j|^p\Big)^{1/p}
\le c_1+C_\beta\Big(c_2\sqrt{p+\log(1/\delta)}+c_3\big(p+\log(1/\delta)\big)^\beta\Big).
\]
\end{lemma}

\begin{proof}
Integrating the tail bound gives, for every $q\ge1$,
$\|(|\mathcal Z_j|-c_1)_+\|_{L^q}\le C_\beta(c_2\sqrt q+c_3q^\beta)$. Let
$X:=(\sum_j\gamma_j(|\mathcal Z_j|-c_1)_+^p)^{1/p}$. For $q\ge p$,
Jensen's inequality for the probability weights $\gamma_j$, applied to the convex map $x\mapsto x^{q/p}$ on $[0,\infty)$, gives
$$\left(\sum_j\gamma_j(|\mathcal Z_j|-c_1)_+^p\right)^{q/p}\le\sum_j\gamma_j(|\mathcal Z_j|-c_1)_+^q$$ pointwise, and taking expectations,
$\|X\|_{L^q}\le C_\beta(c_2\sqrt q+c_3q^\beta)$.
Markov's inequality with $q=p+\log(1/\delta)$ gives
$\mathbb P(X>e\|X\|_{L^q})\le e^{-q}\le\delta$. By the sub-additivity of $x\mapsto (x)_+$ and  Minkowski's
inequality, we obtain
$(\sum_j\gamma_j|\mathcal Z_j|^p)^{1/p}\le c_1+X$.
\end{proof}

\begin{remark}[Random weights]
\label{rem:LogRemoval_weights}
Let $\mathcal G$ be a $\sigma$-algebra and $\mathcal E\in\mathcal G$, and let the weights $\gamma_j$ and the constants $c_1,c_2,c_3$ be $\mathcal G$-measurable. If, for every $\delta\in(0,1]$ and every $j$, the tail bound holds with conditional probability at least $1-\delta$ given $\mathcal G$, almost surely on $\mathcal E$, then so does the conclusion, by the same proof with conditional expectations. In Section~\ref{sec:ProofThmMain}, $\mathcal G=\mathcal F_W$ and $\mathcal E=\mathcal E^W$, as in Lemma~\ref{lemma:MC}.
\end{remark}

\section{Operator Norms of Random Matrices}\label{app:opnorm}
\begin{lemma}\label{lemma:OpNormLayerAvg}
Let $\theta=(\theta^{rm})_{1\le r\le L,\,1\le m\le M}$ be vectors in $\R^D$ whose coordinates $\theta^{rmi}$ are i.i.d., centred, and satisfy $\|\theta^{rmi}\|_{\psi_2}\le\sigma\sqrt D$. Let $\mathcal J\subseteq\{1,\ldots,L\}$ be a set of $n\ge1$ layers, and $C^{\mathcal J}_{p,q}(\theta)$ be the layer-averaged operator norm \eqref{eqn:C_{p,q}^{l}}, with $p,q\ge1$. Then, there exists an absolute constant $C'$, independent of $p$ and $q$, such that, for every $\delta\in(0,1]$, with probability at least $1-\delta$,
\[
C^{\mathcal J}_{p,q}(\theta)\le \sigma C'\left(n^{1/p}\sqrt{p \vee q}+ n^{(1/p-1/2)_+}\frac{\sqrt{D}+\sqrt{\log(1/\delta)}}{M^{\frac{1}{q\vee 2}}}\right).
\]
\end{lemma}
\begin{proof}
Since the layers are i.i.d., we may assume that $\mathcal J=\{1,\ldots,l\}$ with $l:=n$. Replacing $\theta^{rm}$ by $\theta^{rm}/\sqrt D$, we may assume that the coordinates have $\psi_2$-norm at most $\sigma$, and that
\[
C^{\mathcal J}_{p,q}(\theta)=\sup_{u\in \mathbb B^D_{2}}\left(\sum_{r=1}^l\left(\frac{1}{M}\sum_{m=1}^M|\langle \theta^{rm},u \rangle|^q\right)^{p/q}\right)^{1/p}.
\]
By duality, this expression can be written as
\[
\sup_{(u,v)\in \mathbb B^{D}_2\times \mathcal V_{p,q} } \frac{Y_{u,v}}{M^{1/q}},\quad Y_{u,v}:= \sum_{r,m=1}^{l,M} v_{rm} \langle u, \theta^{rm} \rangle, \quad\
    \mathcal V_{p,q}:=\{v\in \R^{l\times M} \colon \| v\|_{p^*,q^*}\le 1\}.
\]
Here, $p^*,q^*$ refer to the H\"older dual exponents of $p,q$. Since each coordinate of $\theta^{rm}$ is independent, centred, and sub-Gaussian, there exists a universal constant $C'$ such that
\begin{align*}
    \E \left[\exp\left\{\lambda (Y_{u^1,v^1}-Y_{u^2,v^2})\right\} \right]&\le \exp\left(\lambda^2 C' \sigma^2 d^2((u^1,v^1),(u^2,v^2))\right),
\end{align*}
for every $u^1,u^2\in \R^D, v^1,v^2\in \R^{l\times M}$ and $\lambda\in \R$, where $d$ is the canonical metric
\begin{align*}
&G_{u,v }= \sum_{r=1}^l\sum_{m=1}^M v_{rm}\langle u , g_{rm}\rangle, \qquad g_{rm}\overset{i.i.d.}{\sim}\mathcal N(0,I_D),\\& d((u^1,v^1),(u^2,v^2)) = \| G_{u^1,v^1}-G_{u^2,v^2}\|_{L^2}.  
\end{align*}
% Indeed, $Y_{u^1,v^1}-Y_{u^2,v^2}=\sum_{r,m,i}(u^1_iv^1_{rm}-u^2_iv^2_{rm})\theta^{rmi}$ is a sum of independent centred sub-Gaussian variables, so that Hoeffding's inequality (Vershynin, \emph{op.\ cit.}, Proposition~2.6.1) bounds its $\psi_2$-norm by an absolute constant times
% \[
% \sigma\Big(\sum_{r,m,i}(u^1_iv^1_{rm}-u^2_iv^2_{rm})^2\Big)^{1/2}=\sigma d((u^1,v^1),(u^2,v^2)).
% \] 
Therefore, $(Y_{u,v})$ for $(u,v)\in \mathbb B_2^D\times \mathcal V_{p,q}$ is a process with sub-Gaussian increments with respect to the metric $C'\sigma d$. Denote $\mathbb B_2^D\times \mathcal V_{p,q}$ by $\mathbb T$. Since $Y_{u,0}=0$, the generic chaining tail bound \cite[Theorem~3.2]{Dirksen}, applied to $\sup_{t\in\mathbb T}|Y_t-Y_{t_0}|$ with $t_0=(u,0)$, gives a universal constant $C$ such that
\begin{align}\label{eqn:P1-delta}
&\mathbb P\left[\sup_{(u,v)\in \mathbb T}|Y_{u,v}|>C\sigma\left(\gamma_2(\mathbb T)  +\mathrm{diam}(\mathbb T)\sqrt{\log(1/\delta)} \right)\right] \le \delta.
\end{align}
where $\gamma_2(\mathbb T)$ and $\mathrm{diam}(\mathbb T)$ are Talagrand's generic-chaining functional on $(\mathbb T,d)$ \cite{Talagrand2014} and the diameter of $(\mathbb T,d)$, respectively. The latter is defined by
\[
    \mathrm{diam}(\mathbb T) = \sup_{(u^1,v^1),(u^2,v^2)\in \mathbb T} d((u^1,v^1),(u^2,v^2)).
\]
Since for any $(u^1,v^1),(u^2,v^2)\in \mathbb T$, we have
\begin{align*}
    d((u^1,v^1),(u^2,v^2))&\le  \left(\sum_{r,m,i=1}^{l,M,D}(u^1_i v_{rm}^1 -u^2_i v_{rm}^2 )^2 \right)^{1/2}\le\|v^1\|_\mathrm{F}+\|v^2\|_\mathrm{F}.
\end{align*}
This implies that $\mathrm{diam}(\mathbb T)\le2\,\mathrm{rad}(\mathcal V_{p,q})$, where $\mathrm{rad}(T):=\sup_{t\in T}\|t\|_2$. For $v\in\mathcal V_{p,q}$, H\"older's inequality gives $$\left(\sum_m v_{rm}^2\right)^{1/2}\le M^{(1/2-1/q^*)_+}\|v_{r\cdot}\|_{q^*}=M^{(1/q-1/2)_+}\|v_{r\cdot}\|_{q^*}$$ for each $r$, and similarly over the layers, so that
\[
    \mathrm{rad}(\mathcal V_{p,q})\le l^{(1/p-1/2)_+}M^{(1/q-1/2)_+},\qquad \mathrm{diam}(\mathbb T)\le 2l^{(1/p-1/2)_+}M^{(1/q-1/2)_+}.
\]
Whereas, we bound $\gamma_2(\mathbb T)$ using Talagrand's Majorizing Measure Theorem \cite{Talagrand2014}, which states that 
\begin{align*}
    \gamma_2(\mathbb T)& \le C''\E \sup_{(u,v)\in \mathbb T} G_{u,v}
\end{align*}
for some universal constant $C''$. Since $\gamma_2$ can be bounded above by the expected supremum of a Gaussian process that is linear in each argument, Chevet's inequality \cite[Theorem 8.6.1]{Vershynin_2026} applies to this process and yields
\[
\gamma_2(\mathbb T) \le C'' \left( w(\mathbb B_2^D) \mathrm{rad}(\mathcal V_{p,q}) +w(\mathcal V_{p,q}) \mathrm{rad}(\mathbb B_2^D)\right),
\]
where $w(T)$ is the Gaussian width of $T\subseteq \R^n$, defined by
\[
w(T) = \E \sup_{t\in T}\langle t, g\rangle, \qquad g\sim \mathcal N(0,I_n).
\]
The Gaussian widths of $\mathbb B_2^D$ and of $\mathcal V_{p,q}$, the latter being $w(\mathcal V_{p,q})=\E\|g\|_{p,q}$ by duality, can be estimated using Jensen's inequality via
\begin{align*}
    w(\mathbb B_2^D)&\le \sqrt{D} ,\\
    w(\mathcal V_{p,q})&\le l^{1/p}\left(\E \left[\left(\sum_{m=1}^M |g_{m}|^q\right)^{p/q}\right]\right)^{1/p}\le Cl^{1/p}M^{1/q}\left(\sqrt{q\vee p}\right),
\end{align*}
where $g_m$ are i.i.d. standard Gaussian random variables. For the last inequality, Jensen's inequality when $p\le q$, and $(\sum_m|g_m|^q)^{p/q}\le M^{p/q-1}\sum_m|g_m|^p$ when $p>q$, together with the Gaussian moment bound $(\mathbb E|g_1|^r)^{1/r}\le C\sqrt r$, give the result. Finally, as $\mathrm{rad}(\mathbb B_2^D)=1$,
we have that
\[
\gamma_2(\mathbb T) \le C''\left( l^{(1/p-1/2)_+}M^{(1/q-1/2)_+}\sqrt{D}+l^{1/p}M^{1/q}\sqrt{p \vee q}\right).
\]
Substituting the above bounds on $\gamma_2(\mathbb T)$ and $\mathrm{diam}(\mathbb T)$ into \eqref{eqn:P1-delta}, dividing by $M^{1/q}$ and using $(1/q-1/2)_+-1/q=-1/(q\vee2)$, yields
\[
\sup_{(u,v)\in \mathbb T} \frac{Y_{u,v}}{M^{1/q}} \le \sigma C\left(l^{1/p}\sqrt{p \vee q}+ l^{(1/p-1/2)_+}M^{-\frac{1}{q\vee 2}}\left(\sqrt{D}+\sqrt{\log(1/\delta)}\right)\right) ,
\]
with probability at least $1-\delta$.
\end{proof}

\section{Bounds on the Features, the Response Matrix and the Limiting Model}\label{app:features}
This appendix collects the auxiliary estimates on the linear response matrix of Section~\ref{sec:LinearResponse}, on the features $P,Q$ of a single layer, and on the coordinates of the limiting model.

\subsection{The Linear Response Matrix}
\begin{lemma}\label{lemma:BoundMathcalR}
Suppose Assumptions~\ref{assm:rho} and~\ref{assm:clipp} hold. Let $G>0$, and let $\bm Z=(\bm H,\bm A) \in \R^{(K+1)\times 2D}$ be any history, possibly random and depending arbitrarily on the initial parameters, whose Gram matrices \eqref{eqn:Gamma} satisfy
\[
\max_{0\le t,t'\le K}|\Gamma^{\bm H}_{t,t'}|\vee|\Gamma^{\bm A}_{t,t'}|\le G .
\]
Then, with $C_{\mathcal R}:=\sqrt{2}(\sqrt{2}c\vee 1)\Lambda_\rho G$, for every realisation of the initial parameters and all $0\le k'<k\le K$,
    \[
    \mathcal R^*_{k',k}:=\max_{1\le l\le L}\max_{1\le m \le M}\|\mathcal R^{lm}_{k',k}(\bm Z)\|_\mathrm{op} \le \eta C_{\mathcal R}(1+\eta C_{\mathcal R})^{k-k'-1},
    \]
    with $\mathcal R^*_{k,k}=1$, and
    \[
     \sum_{k' \le r \le k}\mathcal R^*_{r,k} \le  (1+\eta C_{\mathcal R})^{k-k'}.
    \]
\end{lemma}
\begin{proof}
    Let us first analyse $\mathcal M_{k',k}$. By applying the boundedness of the derivatives of $\rho$, $\mathrm{cl}$ and $\mathrm{cl}'$, we find that
    \[
    \| \mathcal M_{k',k}(\bm Z)\|_\mathrm{op}^2 \le \eta^2\Lambda_\rho^2\left(2c^2|\Gamma^{\bm H}_{k',k}|^2 + |\Gamma^{\bm A}_{k',k}|^2\right).
    \]
Since $2c^2|\Gamma^{\bm H}_{k',k}|^2+|\Gamma^{\bm A}_{k',k}|^2\le2(2c^2\vee1)G^2$,
\[
\| \mathcal M_{k',k}(\bm Z)\|_\mathrm{op} \le \sqrt{2}(\sqrt{2}c\vee 1)\eta \Lambda_\rho G= \eta C_{\mathcal R}.
\]
 Notice that the right hand side is independent of $k',k$. Since there are $\binom{k-k'-1}{j-1}$ chains $k'=t_0<\ldots<t_j=k$ of length $j$, namely the compositions of $k-k'$ into $j$ positive parts, we obtain that
 \begin{align*}
      \|\mathcal R_{k',k}(\bm{Z})\|_\mathrm{op} &\le \sum_{j=1}^{k-k'} \begin{pmatrix}
     k-k'-1 \\ j-1
 \end{pmatrix} ( \eta C_{\mathcal R} )^j =\eta C_{\mathcal R}\;(1+ \eta C_{\mathcal R})^{k-k'-1},
 \end{align*}
 where the second equality results from the Binomial Theorem. Since $l,m$ are arbitrary here, the bound holds on the maximum over them. The final claim results from evaluating the geometric series.
\end{proof}

\begin{lemma}\label{lemma:MathcalRDerivs}
Suppose Assumptions~\ref{assm:rho} and~\ref{assm:clipp} hold. Let $B>0$, and let $\bm{Z}=(\bm{H},\bm{A})\in \R^{(K+1)\times 2D}$ be any deterministic history with
\[
\| \bm{Z}\|_{\infty, \bar 2}\le B .
\]
Let $C'=C'(\Lambda_\rho,c,B):=C_{\mathcal R}\vee\sqrt3(c\vee1)\Lambda_\rho B^2$, with $C_{\mathcal R}$ as in Lemma~\ref{lemma:BoundMathcalR}, which is non-decreasing in $B$, and let $\beta:=1+\eta C'$. For every $0\le k'<k\le K$, there exist random matrices
\[
\mathcal N^{\psi}_{k''},\ \mathcal N^{\phi}_{k''}\in\R^{2\times2},\qquad 0\le k''<k,
\]
which depend on $(k',k)$, and on the initial parameters only through the features $(P_t,Q_t)_{t<k}$, such that the following holds for every realisation of the initial parameters.
\begin{enumerate}
\item For every coordinate $1\le i\le D$,
\begin{equation}\label{eqn:MathcalRDerivs_form}
\partial_{\psi^i}\mathcal R_{k',k}(\bm{Z})=\frac1D\sum_{k''<k}H^i_{k''}\,\mathcal N^{\psi}_{k''},
\qquad
\partial_{\phi^i}\mathcal R_{k',k}(\bm{Z})=\frac1D\sum_{k''<k}A^i_{k''}\,\mathcal N^{\phi}_{k''} .
\end{equation}
In particular, the matrices $\mathcal N^{\psi}_{k''}$ and $\mathcal N^{\phi}_{k''}$ do not depend on $i$.
\item 
\begin{equation}\label{eqn:MathcalRDerivs_bound}
\sum_{k''<k}\|\mathcal N^{\psi}_{k''}\|_{\mathrm{op}}\vee\sum_{k''<k}\|\mathcal N^{\phi}_{k''}\|_{\mathrm{op}}\le\sqrt2\,\eta C'\beta^{2(k-1)-k'} .
\end{equation}
\item For every $u\in\R^D$,
\[
\Big\|\sum_{i=1}^Du_i\,\partial_{\psi^i}\mathcal R_{k',k}(\bm{Z})\Big\|_{\mathrm{op}}\le\sqrt2\,\eta C'\beta^{2(k-1)-k'}\max_{0\le k''<k}|\langle u,H_{k''}\rangle_{\bar2}|,
\]
and the same bound holds for $\partial_{\phi^i}$, with $H_{k''}$ replaced by $A_{k''}$.
\end{enumerate}
If $k'=k$, then the derivatives are zero, since $\mathcal R_{k,k}=I_2$.
\end{lemma}
\begin{proof}
We prove the claims for $\partial_{\psi^i}$. First, differentiating \eqref{eqn:mathcal M} gives, for $t<t'$,
\[
\partial_{\psi^i}\mathcal M_{t,t'} = \mathfrak{M}^1_{t,t'}\, \partial_{\psi^i}P_{t}+\mathfrak{M}^2_{t,t'}\, \partial_{\psi^i}Q_{t},
\]
where
\[
\mathfrak M^1_{t,t'}:=-\eta\begin{pmatrix}\Gamma^{\bm H}_{t,t'}\rho'''\mathrm{cl}&\Gamma^{\bm H}_{t,t'}\rho''\mathrm{cl}'\\ \Gamma^{\bm A}_{t,t'}\rho''&0\end{pmatrix},\qquad
\mathfrak M^2_{t,t'}:=-\eta\begin{pmatrix}\Gamma^{\bm H}_{t,t'}\rho''\mathrm{cl}'&\Gamma^{\bm H}_{t,t'}\rho'\mathrm{cl}''\\ 0&0\end{pmatrix},
\]
with $\rho^{(n)}$ evaluated at $P_{t}$ and $\mathrm{cl}^{(n)}$ at $Q_{t}$. By Assumption~\ref{assm:rho} and $|\Gamma^{\bm H}|\vee|\Gamma^{\bm A}|\le B^2$, bounding the operator norm by the Frobenius norm, we find
\[
\|\mathfrak{M}^1_{t,t'}\|_{\mathrm{op}}\vee\|\mathfrak{M}^2_{t,t'}\|_{\mathrm{op}} \le \eta\Lambda_\rho B^2\sqrt{2c^2+1}\le\eta C' .
\]
Note the form of the derivative in \eqref{eqn:PQ_Deriv_theta},
\[
\begin{pmatrix}\partial_{\psi^i}P_t\\ \partial_{\psi^i}Q_t\end{pmatrix}=\frac1D\sum_{k''\le t}H^i_{k''}\,(\mathcal R_{k'',t})_{\cdot1},
\]
so that
\begin{equation}\label{eqn:DerivM_form}
\partial_{\psi^i}\mathcal M_{t,t'}=\frac1D\sum_{k''\le t}H^i_{k''}\,\mathfrak N_{k'',t,t'},
\qquad
\mathfrak N_{k'',t,t'}:=(\mathcal R_{k'',t})_{11}\,\mathfrak M^1_{t,t'}+(\mathcal R_{k'',t})_{21}\,\mathfrak M^2_{t,t'} .
\end{equation}
Next, by the definition \eqref{eqn:mathcal R} of $\mathcal R$ and the product rule, the derivative of $\mathcal R$ is given by
\[
\partial_{\psi^i}\mathcal R_{k',k}= \sum_{j=1}^{k-k'}\sum_{k' = t_0< \ldots< t_j=k} \sum_{l=1}^j \mathcal M_{t_{j-1},t_j}\cdots \mathcal M_{t_{l},t_{l+1}}\big(\partial_{\psi^i}\mathcal M_{t_{l-1},t_l}\big) \mathcal M_{t_{l-2},t_{l-1}}\cdots \mathcal M_{t_0,t_1}.
\]
Substituting \eqref{eqn:DerivM_form} and exchanging the sums gives the first identity in \eqref{eqn:MathcalRDerivs_form}, with
\[
\mathcal N^{\psi}_{k''}:=\sum_{j=1}^{k-k'}\sum_{k' = t_0< \ldots< t_j=k} \sum_{l=1}^j\mathbbm 1_{\{k''\le t_{l-1}\}}\, \mathcal M_{t_{j-1},t_j}\cdots \mathcal M_{t_{l},t_{l+1}}\,\mathfrak N_{k'',t_{l-1},t_l}\, \mathcal M_{t_{l-2},t_{l-1}}\cdots \mathcal M_{t_0,t_1}.
\]
Since $t_{l-1}<k$, only the indices $k''<k$ contribute. Every factor depends on the initial parameters only through the features $(P_t,Q_t)_{t<k}$, and none depends on $i$.  As seen in the proof of Lemma~\ref{lemma:BoundMathcalR}, each $\mathcal M$ has operator norm at most $\eta C_{\mathcal R}\le\eta C'$. Hence, we introduce
\[
\mathfrak r_t:=\sum_{k''\le t}\big(|(\mathcal R_{k'',t})_{11}|+|(\mathcal R_{k'',t})_{21}|\big),
\qquad\text{so that}\qquad
\sum_{k''\le t}\|\mathfrak N_{k'',t,t'}\|_{\mathrm{op}}\le\eta C'\,\mathfrak r_t .
\]
Since $\mathcal R_{t,t}=I_2$ and, for $k''<t$, the $\ell^1$-norm of a column of $\mathcal R_{k'',t}$ is at most $\sqrt2$ times its operator norm, Lemma~\ref{lemma:BoundMathcalR} gives
\[
\mathfrak r_t\le1+\sqrt2\sum_{k''<t}\eta C'\beta^{t-k''-1}=1+\sqrt2(\beta^t-1)\le\sqrt2\beta^t .
\]
In the chains above, the differentiated factor has lower end point $t_{l-1}$. For $l=1$ this is $t_0=k'$, and there are $\binom{k-k'-1}{j-1}$ chains of length $j$. For $l\ge2$ it is an interior point $k'<t<k$, and, by Vandermonde's identity, there are $\binom{k-k'-2}{j-2}$ chains of length $j$ through $t$. Hence, by the Binomial Theorem,
\begin{align*}
\sum_{k''<k}\|\mathcal N^{\psi}_{k''}\|_{\mathrm{op}}
&\le\sum_{j=1}^{k-k'}(\eta C')^j\binom{k-k'-1}{j-1}\mathfrak r_{k'}+\sum_{j=2}^{k-k'}(\eta C')^j\binom{k-k'-2}{j-2}\sum_{t=k'+1}^{k-1}\mathfrak r_t\\
&=\eta C'\beta^{k-k'-1}\,\mathfrak r_{k'}+(\eta C')^2\beta^{k-k'-2}\sum_{t=k'+1}^{k-1}\mathfrak r_t\\
&\le\sqrt2\,\eta C'\beta^{k-k'-2}\Big[\beta^{k'+1}+\eta C'\sum_{t=k'+1}^{k-1}\beta^t\Big]
=\sqrt2\,\eta C'\beta^{2(k-1)-k'},
\end{align*}
where the second sum is absent if $k-k'=1$, and the last step uses $\eta C'\sum_{t=k'+1}^{k-1}\beta^t=\beta^k-\beta^{k'+1}$. This is \eqref{eqn:MathcalRDerivs_bound}. Finally, by \eqref{eqn:MathcalRDerivs_form},
\[
\sum_{i=1}^Du_i\,\partial_{\psi^i}\mathcal R_{k',k}=\sum_{k''<k}\langle u,H_{k''}\rangle_{\bar2}\,\mathcal N^{\psi}_{k''},
\]
and the third claim follows from \eqref{eqn:MathcalRDerivs_bound}. For $\partial_{\phi^i}$, \eqref{eqn:PQ_Deriv_theta} gives the same formulas with $H^i_{k''}$ and $(\mathcal R_{k'',t})_{\cdot1}$ replaced by $A^i_{k''}$ and $(\mathcal R_{k'',t})_{\cdot2}$, and the argument is identical.
\end{proof}

\subsection{Moments and Tails of the Features}
\begin{lemma}\label{lemma:p-norm_on_PQ}
Suppose Assumptions~\ref{assm:MLU}, \ref{assm:rho} and~\ref{assm:clipp} hold. Let $B>0$ and $p\ge2$, and let \\$\widetilde{\bm{Z}}\in \R^{(K+1)\times 2D}$ be a history, possibly random, such that, almost surely,
\begin{align*}
    &\max_{0\le k \le K} \| \widetilde{H}_k\|_{\bar 2}\vee\| \widetilde{A}_k\|_{\bar 2} \le B.
\end{align*}
Further assume that $\widetilde{\bm{Z}}$ is independent of $(\Psi^l_0,\Phi^l_0)$. Then there exists a constant $C_3=C_3(\Lambda_\rho,c,\eta K,B,\sigma)$, non-decreasing in $B$, such that, for every $\delta\in(0,1]$, with probability at least $1-\delta$, simultaneously for every $0\le k\le K$,
    \begin{align*}
            &\|P^l_k(\widetilde{\bm Z})\|_{\bar p}\vee \|Q^l_k(\widetilde{\bm Z})\|_{\bar p}\le C_3 \left(\sqrt{p}+ \frac{\sqrt{\log(2(K+1)/\delta)}}{M^{1/p}} \right).
    \end{align*}
 Since $\log((K+1)/\delta)\le\mathfrak l_\delta$, the right-hand side is at most $C_3(\sqrt{p}+M^{-\frac{1}{p}}\sqrt{\log(2)+\mathfrak l_\delta})$. In addition, if we no longer assume that $\widetilde{\bm{Z}}$ is independent of $(\Psi^l_0,\Phi^l_0)$, then we still have, almost surely, for every $0\le k\le K$,
    \[
    \|P^l_k(\widetilde{\bm Z})\|_{\bar p}\vee \|Q^l_k(\widetilde{\bm Z})\|_{\bar p}\le C_3\Big(1+\|(\Psi^l_0)^\top\|_{\bar 2\to \bar p}\vee \|(\Phi^l_0)^\top\|_{\bar 2\to \bar p}\Big).
    \]
\end{lemma}
\begin{proof} Fix $0\le k\le K$ and a neuron $m$. By \eqref{eqn:p}, the boundedness of $\rho'$ and $\mathrm{cl}'$ (Assumptions \ref{assm:rho} and \ref{assm:clipp}), and $|\Gamma^{\widetilde{\bm H}}_{k',k}|\le\|\widetilde H_{k'}\|_{\bar2}\|\widetilde H_{k}\|_{\bar2}\le B^2$,
\begin{align*}
    |P^{lm}_k(\widetilde{\bm Z})|&\le|\langle\psi^{lm}_0,\widetilde H_k\rangle_{\bar2}| +\eta c\Lambda_\rho B^2\sum_{0\le k'<k} (1+|Q^{lm}_{k'}(\widetilde{\bm Z})|),\\|Q^{lm}_k(\widetilde{\bm Z})|&\le|\langle\phi^{lm}_0,\widetilde A_k\rangle_{\bar2}| +\eta\Lambda_\rho B^2\sum_{0\le k'<k} (1+|P^{lm}_{k'}(\widetilde{\bm Z})|),
\end{align*}
where the sums are zero when $k=0$. Taking the $\bar p$-norm over $m$ and applying Minkowski's and discrete Gr\"onwall inequalities, we obtain, with
\[
\mathcal H^*_k:=\|(\langle\psi^{lm}_0,\widetilde H_k\rangle_{\bar2})_{m}\|_{\bar p}\vee\|(\langle\phi^{lm}_0,\widetilde A_k\rangle_{\bar2})_{m}\|_{\bar p},\qquad C_0:=\eta K\Lambda_\rho (c\vee 1)B^2,
\]
that, for every $0\le k\le K$,
\begin{equation}\label{eqn:pnorm_pathwise}
\|P^l_k(\widetilde{\bm Z})\|_{\bar p}\vee \|Q^l_k(\widetilde{\bm Z})\|_{\bar p}\le (C_0+\max_{0\le k'\le k}\mathcal H^*_{k'})e^{C_0} .
\end{equation}
This bound is deterministic and holds for any history bounded by $B$, independent of the layer or not.

\emph{Dependent histories.} By the definition of the operator norms, \[\mathcal H^*_{k'}\le B\big(\|(\Psi^l_0)^\top\|_{\bar 2\to \bar p}\vee \|(\Phi^l_0)^\top\|_{\bar 2\to \bar p}\big)\] for every $k'$, which, substituted into \eqref{eqn:pnorm_pathwise}, gives the second claim.

\emph{Independent histories.} Fix $k$ and consider $$\mathcal H(w):=\sum_{m=1}^Mw_m\langle\psi^{lm}_0,\widetilde H_k\rangle_{\bar2},$$ with $w\in\R^M$. By duality,
\[
\|(\langle\psi^{lm}_0,\widetilde H_k\rangle_{\bar2})_m\|_{\bar p}=M^{-1/p}\sup_{w\in\mathbb B^M_{p^*}}\mathcal H(w),\qquad p^*=\frac p{p-1},
\]
where $\mathbb B^M_{p^*}$ is the unit ball of $\ell^{p^*}$. Since $\mathcal H$ is linear and the coordinates $\psi^{lmi}_0$ are independent, centred and sub-Gaussian (Assumption~\ref{assm:MLU}), the standard sub-Gaussian sum bound gives
\[
\|\mathcal H(w)-\mathcal H(w')\|_{\psi_2|\widetilde{\bm Z}}\le C_{\mathrm H}\sigma\|\widetilde H_k\|_{\bar2}\|w-w'\|_2\le C_{\mathrm H}\sigma B\|w-w'\|_2 ,
\]
for a universal constant $C_H$. Thus, $(\mathcal H(w))_{w\in\mathbb B^M_{p^*}}$ has sub-Gaussian increments with respect to $C_{\mathrm H}\sigma B\|\cdot\|_2$, and $\mathcal H(0)=0$. Arguing similarly to the proof of Lemma~\ref{lemma:OpNormLayerAvg}), the generic chaining tail bound \cite[Theorem~3.2]{Dirksen} and Talagrand's majorizing measure Theorem \cite{Talagrand2014} give an absolute constant $C$ such that, with probability at least $1-\delta'$,
\[
\sup_{w\in\mathbb B^M_{p^*}}\mathcal H(w)\le C\sigma B\Big(\mathbb E\sup_{w\in\mathbb B^M_{p^*}}\langle w,g\rangle+\mathrm{diam}(\mathbb B^M_{p^*})\sqrt{\log(1/\delta')}\Big),\qquad g\sim\mathcal N(0,I_M).
\]
Since $p^*\le2$, $\mathrm{diam}(\mathbb B^M_{p^*})\le2$. By duality, $\mathbb E\sup_{w}\langle w,g\rangle=\mathbb E\|g\|_p$, and
\[
\mathbb E\|g\|_p\le (\mathbb E\|g\|_p^p)^{1/p}\le CM^{1/p}\sqrt p,
\]
using standard bounds on the $p$-th moment of a Gaussian random variable. Dividing by $M^{1/p}$, with probability at least $1-\delta'$,
\begin{equation}\label{eqn:HP on mathcal H}
    \|(\langle\psi^{lm}_0,\widetilde H_k\rangle_{\bar2})_m\|_{\bar p} \le C\sigma B\left(\sqrt{p}+\frac{\sqrt{\log(1/\delta')}}{M^{1/p}}\right).
\end{equation}
The same holds for $\|(\langle\phi^{lm}_0,\widetilde A_k\rangle_{\bar2})_m\|_{\bar p}$. Since the RHS does not depend on $\widetilde{\bm Z}$, these bounds also hold unconditionally. A union bound over the $2(K+1)$ events, for $0\le k\le K$, each with $\delta'=\delta/(2(K+1))$, gives, with probability at least $1-\delta$, simultaneously for every $0\le k\le K$,
\[
\mathcal H^*_k\le C\sigma B\left(\sqrt{p}+\frac{\sqrt{\log(2(K+1)/\delta)}}{M^{1/p}}\right).
\]
Substituting this into \eqref{eqn:pnorm_pathwise}, proves the first claim with $C_3$ that is non-decreasing in $B$.
\end{proof}
\begin{lemma}\label{lemma:PQ_psi_2}
Let $B>0$, and let $\bm{Z}\in \R^{(K+1)\times 2D}$ be any deterministic history with \\$\| \bm{Z}\|_{\infty,\bar 2}\le B$. Under Assumptions~\ref{assm:MLU}, \ref{assm:rho} and~\ref{assm:clipp}, for every $0\le k \le K$, $1\le l\le L$ and $1\le m\le M$,
    \[
    \|P^{lm}_k(\bm Z)\|_{\psi_2}\vee \|Q^{lm}_k(\bm Z)\|_{\psi_2} \le C_\psi(B).
    \]
    The right-hand side is non-decreasing in $B$. Moreover, if $\mathcal G$ is a $\sigma$-algebra independent of $(\phi^{lm}_0,\psi^{lm}_0)$ and $\bm Z$ is a $\mathcal G$-measurable history with $\|\bm Z\|_{\infty,\bar2}\le B$ a.s., then the same bound holds for the conditional norms $\|\cdot\|_{\psi_2|\mathcal G}$.
\end{lemma}
\begin{proof}
Write $a:=C_{\mathrm H}\sigma B$, where $C_{\mathrm H}$ is a universal constant from Hoeffding's inequality. By Assumption~\ref{assm:MLU}, the coordinates of $\psi^{lm}_0$ are independent and centred with $\psi_2$-norm at most $\sigma\sqrt D$, so that, by \cite[Lemma~3.4.2]{Vershynin_2026}, $\|\langle\psi^{lm}_0,H_k\rangle_{\bar2}\|_{\psi_2}\le a$, and likewise $\|\langle\phi^{lm}_0,A_k\rangle_{\bar2}\|_{\psi_2}\le a$. The argument seen in the proof of Lemma~\ref{lemma:p-norm_on_PQ}, can be re-run with the $\psi_2$-norm as opposed to the $\bar p$-norm to give
\begin{align*}
           \| P^{lm}_k(\bm Z)\|_{\psi_2} \vee \| Q^{lm}_k(\bm Z)\|_{\psi_2} &\le (C_0(\log(2))^{-1/2}+a)e^{C_0},
    \end{align*}
where we have used $\|1\|_{\psi_2}=(\log 2)^{-1/2}$ . For the conditional version, conditionally on $\mathcal G$ the history is fixed and the initial parameters of the neuron keep their law, so the same argument applies.
\end{proof}

\subsection{Coordinatewise Growth of the Limiting Model}
The fluctuation term in Lemma~\ref{lemma:Lip_in_E} involves the
$\bar\infty$-norms, that is the largest coordinates, of the
directions and of the limiting history. These grow at most like
$\sqrt{\log D}$. For the output directions $\omega_j$, defined in Section~\ref{sec:ProofThmMain}, we only bound the weighted average of their squared coordinates, which avoids any control of the individual Hessian eigenvectors.

\begin{lemma}[Coordinatewise Growth of the Limiting Model]
\label{lemma:InfNorm}
Suppose Assumptions~\ref{assm:SteinMLU}--\ref{assm:Loss} hold, and that $\delta\ge\exp(-c_1D)$. Then there exists $C_\infty$, depending only on $K$, $\eta$, $\eta_{\mathrm{in}}|x|^2$, $\tilde\eta_{\mathrm{out}}$, $C_{\mathrm{lim}}$, $\Lambda_\rho$, $c$, $\sigma$, $\|\kappa(\xi)\|_{L^2}$, $\sigma_{\mathrm{in}}$, $\sigma_{\mathrm{out}}$, $\tau_\star$, $G_0$, $G_\ell$, $R_*$ and $S_*$, such that, on an event of probability at least $1-\delta$, contained in $\mathcal E^W_{\delta/2}$,
\begin{align}
\sup_{t\in[0,1]}\max_{k\le K}\|H_{t,k}\|_{\bar\infty}\vee\|A_{t,k}\|_{\bar\infty}
&\le C_\infty\sqrt{\log(12D/\delta)},\label{eqn:InfNorm_history}\\
\max_{k\le K}\max_{1\le i\le D}\sum_j\alpha_j(Y_k)\big(\omega_j(Y_k)\big)_i^2
&\le C_\infty^2\log(12D/\delta) .\label{eqn:InfNorm_output}
\end{align}
Both bounds remain valid with $\mathfrak l^\infty_\delta$ in place of $\log(12D/\delta)$, after adjusting $C_\infty$.
\end{lemma}

\begin{proof}
The limiting model of Section~\ref{sec:LimitingModel} is an $\mathcal F_W$-measurable function of the initial embeddings. Since $\delta/2 \ge \exp(-(c_1+\log2)D)$, Lemma~\ref{lemma:embed_event} applies at the level $\delta/2$ with $c_1$ replaced by $c_1+\log 2$. Thus, we work on $\mathcal E^W_{\delta/2}$, on which the limiting model history is bounded by $C_{\mathrm{lim}}$ (Lemma~\ref{lemma:C1}) and, by Stage~1 of the proof of Lemma~\ref{lemma:apriori} with $B=C_{\mathrm{lim}}$, $\mathfrak f(W^\infty_{\mathrm{out},k})\le C'_{\mathfrak f}$ and $\mathfrak n(W^\infty_{\mathrm{out},k})\le C'_{\mathfrak n}$, so that its outputs $Y_k$ lie in $\mathcal Y$, where \eqref{eqn:Hessian_comparable} and \eqref{eqn:grad_factor} apply.  Throughout, $C\ge1$ denotes a constant depending only on the quantities in the statement, which may change from line to line. Fix a coordinate $i$, let $W_{0,i}\in\R^{d_{\mathrm{out}}}$ be the $i$-th row of $W_{\mathrm{out},0}$, let $g_0:=\nabla\ell(0,\mathfrak t)$, and
\[
u_k:=\sup_{t\in[0,1]}|H^i_{t,k}|,\quad w_k:=\sup_{t\in[0,1]}|A^i_{t,k}|,\quad
a_i:=|\langle(W_{\mathrm{in},0})_{i\cdot},x\rangle|+|\langle W_{0,i},g_0\rangle|+|W_{0,i}|_{\mathscr H_\star} .
\]
Furthermore, for ease of notation, we drop the superscript $\infty$ of Section \ref{sec:LimitingModel}, which indicates the parameters and features are of the infinite model.
Firstly, the $i$-th coordinate of the limiting equation is
\[
\partial_tH^i_{t,k}=\mathbb E\big[\phi^i_0\,\rho(P_{t,k})\big| \mathcal F_W\big]-\eta\sum_{k'<k}A^i_{t,k'}\,\mathbb E\big[\rho(P_{t,k'})\rho(P_{t,k})\big|\mathcal F_W\big],
\]
where the conditional expectation in the final term is bounded, on $\mathcal E^W_{\delta/2}$, using the linear growth of $\rho$ (Assumption \ref{assm:rho}) and  Lemma \ref{lemma:PQ_psi_2}, which is justified by Lemma \ref{lemma:C1}. By the Stein identity in $\phi^i_0$ (Assumption \ref{assm:SteinMLU}) and \eqref{eqn:PQ_Deriv_theta},
\[
\mathbb E\big[\phi^i_0\,\rho(P_{t,k})\big|\mathcal F_W\big]
=\sum_{k'<k}A^i_{t,k'}\,\mathbb E\big[\kappa(\phi^i_0)\,\rho'(P_{t,k})\,(\mathcal R_{k',k})_{12}\big|\mathcal F_W\big],
\]
where the term $k'=k$ is absent since $\mathcal R_{k,k}=I_2$. This is valid as $\phi^{\infty,i}_0$ is independent of $\mathcal F_W$. By Lemma~\ref{lemma:BoundMathcalR}, $\|\mathcal R_{k',k}\|_{\mathrm{op}}\le C\eta$ for $k'<k$ a.s. By the boundedness of $\rho'$ (Assumption \ref{assm:rho}) and finite second moment of $\kappa$ (Assumption \ref{assm:Stein}), we have on $\mathcal E^W_{\delta/2}$,
\[
|\partial_tH^i_{t,k}|\le C\eta\sum_{k'<k}|A^i_{t,k'}| .
\]
Similarly,
\begin{align*}
&-\partial_tA^i_{t,k}=\mathbb E\big[\psi^i_0\,\rho'(P_{t,k})Q_{t,k}\big|\mathcal F_W \big]-\eta\sum_{k'<k}H^i_{t,k'}\,\mathbb E\big[\rho'(P_{t,k'})\,\mathrm{cl}(Q_{t,k'})\,\rho'(P_{t,k})Q_{t,k}\big|\mathcal F_W\big],\\
&\mathbb E\big[\psi^i_0\,\rho'(P_{t,k})Q_{t,k}\big|\mathcal F_W\big]=\sum_{k'\le k}H^i_{t,k'}\,
\mathbb E\Big[\kappa(\psi^i_0)\big(\rho''(P_{t,k})Q_{t,k},\,\rho'(P_{t,k})\big)(\mathcal R_{k',k})_{\cdot1}\big|\mathcal F_W\Big].
\end{align*}
Here the term $k'=k$ is present, with $| (\mathcal R_{k,k})_{\cdot,1}|=1$. Applying the same argument as done for the hidden states, on $\mathcal E^W_{\delta/2}$ we have
\[
|\partial_tA^i_{t,k}|\le C|H^i_{t,k}|+C\eta\sum_{k'<k}|H^i_{t,k'}| .
\]
Let us write $W_{k,i}$ for the $i$-th row of $W^\infty_{\mathrm{out},k}$. The embeddings of the limiting model are trained along $A_{0,k}$ and $H_{1,k}$, so that the initial hidden state and terminal adjoint satisfy
\begin{align*}
    H^i_{0,k}&=\langle(W_{\mathrm{in},0})_{i\cdot},x\rangle-\eta_{\mathrm{in}}|x|^2\sum_{k'<k}A^i_{0,k'},\\
A^i_{1,k}&=\langle W_{k,i},\nabla\ell(Y_k)\rangle=\langle W_{0,i},\nabla\ell(Y_k)\rangle-\eta_{\mathrm{out}}\sum_{k'<k}H^i_{1,k'}\,\nabla\ell(Y_{k'})^\top\nabla\ell(Y_k)
\end{align*}
 By \eqref{eqn:grad_factor} with $y'=0$ and the Cauchy--Schwarz inequality, \[|\langle W_{0,i},\nabla\ell(Y_k)\rangle|\le|\langle W_{0,i},g_0\rangle|+e^{R_*}|W_{0,i}|_{\mathscr H_\star}|Y_k|_{\mathscr H_\star}\le(1+e^{R_*}S_*)\,a_i,\] where the final inequality results from the definition of $a_i$ and that $Y\in \mathcal Y$  (Lemma \ref{lemma:C1}).  
 Since $|\nabla\ell|\le G_\ell$ on $\mathcal Y$,
\begin{equation}\label{eqn:InfNorm_boundary}
|H^i_{0,k}|\le a_i+\eta_{\mathrm{in}}|x|^2\sum_{k'<k}w_{k'},
\qquad
|A^i_{1,k}|\le C\,a_i+\eta_{\mathrm{out}}G_\ell^2\sum_{k'<k}u_{k'} .
\end{equation}
With the above estimates, we can now close a Gr\"onwall argument over the training steps. Integrating the bound on the derivatives of $H^i_t,A^i_t$ over $t$ and using \eqref{eqn:InfNorm_boundary},
\[
u_k\le a_i+C\bar\eta\sum_{k'<k}w_{k'},
\qquad
w_k\le C a_i+C\,u_k+C\bar\eta\sum_{k'<k}u_{k'} .
\]
Substituting the first bound into the second, and applying the discrete Gr\"onwall inequality yields
\begin{equation}\label{eqn:InfNorm_induction}
\max_{0\le k \le K} u_k\vee w_k\le C\,a_i.
\end{equation}
\emph{The Output Directions.} This step is only needed when $\mathbb V>0$. By \eqref{eqn:output_directions},
\[
\sum_j\alpha_{j,k}(\omega_{j,k})_i^2=\frac{W_{k,i}^\top \nabla^2\ell(Y_k)W_{k,i}}{\mathrm{tr}\,\nabla^2\ell(Y_k)}\le C_{\mathscr H}\,\frac{|W_{k,i}|_{\mathscr H_\star}^2}{\tau_\star},
\]
since $\mathscr{H}^0_k\preceq e^{R_*}\mathscr H_\star$ and $\mathrm{tr}(\mathscr{H}^0_k- e^{-R_*}\mathscr{H}_\star)\ge 0$, by \eqref{eqn:Hessian_comparable}. Since $Y_{k'}\in \mathcal Y$, we have  $|\nabla\ell(Y_{k'})|_{\mathscr H_\star}\le\sqrt{\lambda_\star}G_\ell$, which, together with \eqref{eqn:InfNorm_induction}, yields
\begin{equation}\label{eqn:InfNorm_output_det}
|W_{k,i}|_{\mathscr H_\star}\le|W_{0,i}|_{\mathscr H_\star}+\eta_{\mathrm{out}}\sqrt{\lambda_\star}G_\ell\sum_{k'<k}u_{k'}\le C\,a_i .
\end{equation}
% Unlike a bound through the largest entries of $W_{k,i}$, this involves neither the $(1,1)$-norm of the Hessian nor the output dimension.

\emph{Probability Estimates} The bounds \eqref{eqn:InfNorm_induction} and \eqref{eqn:InfNorm_output_det} hold on $\mathcal E^W_{\delta/2}$ for every $t$, $k\le K$ and $i\le D$, and only involve the $3D$ variables $\langle(W_{\mathrm{in},0})_{i\cdot},x\rangle$, $\langle W_{0,i},g_0\rangle$ and $|W_{0,i}|_{\mathscr H_\star}$. The first two are sub-Gaussian, with $\psi_2$-norms at most $\sigma_{\mathrm{in}}$ and $C\sigma_{\mathrm{out}}G_0$, and, by the Hanson--Wright inequality \cite[Theorem 1.1]{HansonWright} as in the proof of Lemma~\ref{lemma:embed_event}, \[|W_{0,i}|_{\mathscr H_\star}\le C\sigma_{\mathrm{out}}\sqrt{\tau_\star}(1+\sqrt{\log(2/\delta)})\] with probability at least $1-\delta$. A union bound over the $3D$ variables gives, with probability at least $1-\delta/2$,
\[
\max_{1\le i\le D}a_i\le C\big(\sigma_{\mathrm{in}}+\sigma_{\mathrm{out}}(G_0+\sqrt{\tau_\star})\big)\sqrt{\log(12D/\delta)} .
\]
Intersecting with $\mathcal E^W_{\delta/2}$ gives an event of probability at least $1-\delta$, on which \eqref{eqn:InfNorm_induction} gives \eqref{eqn:InfNorm_history}, and the \emph{Output Directions} step establishes \eqref{eqn:InfNorm_output}. The last claim holds since $\log(12D/\delta)\le\mathfrak l^\infty_\delta$.
\end{proof}

\section{Proofs for the Output Geometry}\label{app:output_geometry}
This appendix contains the proofs of Lemmas~\ref{lemma:embed_event} and~\ref{lemma:grad_metric} of Section~\ref{sec:BD}.

\begin{proof}[Proof of Lemma~\ref{lemma:embed_event}]
The variables $\langle(W_{\mathrm{in},0})_{i\cdot},x\rangle^2$ are independent, with $\psi_1$-norm at most $\sigma_{\mathrm{in}}^2$, and Bernstein's inequality \cite[Theorem~2.9.1]{Vershynin_2026} gives, with probability at least $1-\delta/4$,
\[
\|W_{\mathrm{in},0}\,x\|_{\bar2}^2=\frac1D\sum_{i=1}^D\langle(W_{\mathrm{in},0})_{i\cdot},x\rangle^2\le C\sigma_{\mathrm{in}}^2\Big(1+\frac{\log(8/\delta)}D\Big).
\]
Let $W_{0,i}\in\R^{d_{\mathrm{out}}}$ be the $i$-th row of $W_{\mathrm{out},0}$, and $g_0:=\nabla\ell(0,\mathfrak t)$. The variables $\langle W_{0,i},g_0\rangle$ are independent and centred, with $\psi_2$-norm at most $C\sigma_{\mathrm{out}}|g_0|$ \cite[Proposition~2.6.1]{Vershynin_2026}, so that the same argument bounds $\|W_{\mathrm{out},0}g_0\|_{\bar2}^2$ with probability at least $1-\delta/4$, with $ \sigma_{\mathrm{out}}G_0$ in  place of $\sigma_\mathrm{in}$. Recall the definition of $\mathfrak{f}$ in \eqref{eqn:mathfrak f&n}, whose square is an average of $D$ independent sub-Exponential random variables \[\mathfrak f(W_{\mathrm{out},0})^2=\frac1D\sum_iX_i, \qquad X_i:=|W_{0,i}|^2_{\mathscr H_\star}.\] By the Hanson--Wright inequality \cite[Theorem~1.1]{HansonWright}, and since $\|\mathscr H_\star\|_F\vee\|\mathscr H_\star\|_{\mathrm{op}}\le\tau_\star$,
\[
\mathbb P\big(|X_i-\mathbb EX_i|>u\big)\le2\exp\Big(-c\min\Big(\frac{u^2}{\sigma_{\mathrm{out}}^4\tau_\star^2},\frac u{\sigma_{\mathrm{out}}^2\tau_\star}\Big)\Big),\qquad u\ge0,
\]
so that $\|X_i-\mathbb EX_i\|_{\psi_1}\le C\sigma_{\mathrm{out}}^2\tau_\star$, and Bernstein's inequality produces the same bound on $\mathfrak f(W_{\mathrm{out},0})^2$ as on the previous variables with probability at least $1-\delta/4$, but with $\sigma_{\mathrm{out}}\sqrt{\tau_\star}$ in place of $\sigma_\mathrm{in}$. \\

Finally, $\|\cdot\|_\ell$ is a finite, symmetric, sublinear function on $\R^{d_{\mathrm{out}}}$. Positive homogeneity follows by rescaling $s$, and subadditivity by the triangle inequality for $d_T$ and taking the supremum over all base points. Hence it is the support function $\|v\|_\ell=\sup_{z\in \hat K}\langle z,v\rangle$ of a non-empty bounded convex set $\hat K$ \cite[Corollary~13.2.2]{ConvexAnalysis}. By (L2), its radius $\sup_{z\in\hat K}|z|$ is $\Lambda_\ell$ and Gaussian width $\mathbb E\sup_{z\in \hat K}\langle z,g\rangle=\mathcal{GW}(\|\cdot\|_\ell)$. Since $\|h\|_{\bar2}\le1$ if and only if $|h|\le\sqrt D$,
\[
\mathfrak n(W_{\mathrm{out},0})=\frac1{\sqrt D}\sup_{|h|\le1,\,z\in \hat K}h^\top W_{\mathrm{out},0}\,z .
\]
The entries of $W_{\mathrm{out},0}$ are independent and centred, with $\psi_2$-norm at most $\sigma_{\mathrm{out}}$. Therefore, the process $(h,z)\mapsto h^\top W_{\mathrm{out},0}z$ has sub-Gaussian increments with respect to the canonical metric of the Gaussian process $h^\top Gz$, where $G$ has i.i.d.\ standard Gaussian entries. It vanishes at $h=0$, and its index set has diameter at most $2\Lambda_\ell$. Accordingly, Talagrand's comparison inequality in its tail form \cite[Theorem~5.10]{TalagrandWithTail} applies and yields
\[
\mathfrak n(W_{\mathrm{out},0})\le \frac{C\sigma_\mathrm{out}}{\sqrt{D}}\left( \E\sup_{|h|\le1,\,z\in \hat K} h^\top G z + \Lambda_\ell\sqrt{\log(2/\delta)}\right),
\]
Chevet's inequality for the Gaussian process, \cite[Theorem~8.6.1]{Vershynin_2026}, gives, with probability at least $1-\delta/4$,
\[
\mathfrak n(W_{\mathrm{out},0})\le\frac{C\sigma_{\mathrm{out}}}{\sqrt D}\Big(\sqrt D\,\Lambda_\ell+\mathcal{GW}(\|\cdot\|_\ell)+\Lambda_\ell\sqrt{\log(8/\delta)}\Big)
.
\] 
Since $\log(8/\delta)\le(3+c_1)D$, a union bound proves the lemma.
\end{proof}

\begin{proof}[Proof of Lemma~\ref{lemma:grad_metric}]
The segment $\gamma(t):=y'+t(y-y')$ lies in the convex set $\mathcal Y$, so that, by \eqref{eqn:Hessian_comparable},
\[
d_\ell(y,y')^2\le\mathcal A_\ell(\gamma)=\int_0^1|y-y'|^2_{\mathscr H(\gamma(t))}\,dt\le C_{\mathscr H}\,|y-y'|^2_{\mathscr H(\bar y)} .
\]
By \eqref{eqn:grad_factor} and the Cauchy--Schwarz inequality,
\[
w^\top\big(\nabla\ell(y)-\nabla\ell(y')\big)=\big(\mathscr H_\star^{1/2}w\big)^\top \widetilde M_{y,y'}\,\mathscr H_\star^{1/2}(y-y')\le e^{R_*}\,|w|_{\mathscr H_\star}\,|y-y'|_{\mathscr H_\star},
\]
and, by the first bound of \eqref{eqn:Hessian_comparable}, $|v|_{\mathscr H_\star}\le e^{R_*/2}|v|_{\mathscr H(\bar y)}$ for every $v$. Since $e^{2R_*}=C_{\mathscr H}$, this proves the second bound.
\end{proof}

\section{Dependence on the Loss and the Output Dimension}\label{app:dout}
This appendix expands on Remark~\ref{rem:gw}. Section~\ref{app:dout_where} locates where the loss and the dimensions enter the constants, Section~\ref{app:dout_examples} treats two common loss function examples.

\subsection{Where the Loss Enters the Constants}\label{app:dout_where}
All constants of the paper are built from the a priori bounds of Lemmas~\ref{lemma:embed_event}, \ref{lemma:apriori} and~\ref{lemma:boundary}, from Lemma~\ref{lemma:InfNorm}, and from the effective learning rate $\bar\eta$. The loss enters these only through $\tau_\star$, $\Lambda_\ell$, $\mathfrak g_\ell$, $\varkappa_\ell$, $G_0$ and $\lambda_\star\le\tau_\star$, in the following places. We keep $\tau_\star$ explicit, rather than using its normalisation to make it clear how different normalisations of the loss affect these constants.
\begin{enumerate}
\item \emph{The initial output embedding.} By Appendix~\ref{app:output_geometry}, and since $\delta\ge e^{-c_1D}$, the constants of Lemma~\ref{lemma:embed_event} satisfy
\[
C_{\mathfrak f}\le C\sigma_{\mathrm{out}}\sqrt{\tau_\star},\qquad
C_r\le C\sigma_{\mathrm{out}}G_0,\qquad
C_{\mathfrak n}\le C\sigma_{\mathrm{out}}\big((1+\sqrt{3+c_1})\Lambda_\ell+\mathfrak g_\ell\big).
\]
The first bounds the size of the initial outputs in the geometry of $\mathscr H_\star$, the second the initial terminal adjoint, and the third how far the Hessian can move along the initial outputs. The same products $\sigma_{\mathrm{out}}(G_0+\sqrt{\tau_\star})$ bound the $\bar\infty$-norms in Lemma~\ref{lemma:InfNorm}.
\item \emph{The training of the output embedding.} In the proof of Lemma~\ref{lemma:apriori}, the update $-\eta_{\mathrm{out}}h_{L,j}g_j^\top$, with $|g_j|\le G_0+\sqrt{\lambda_\star}e^{n_jB}s_jB$ and $\|g_j\|_\ell\le\varkappa_\ell$, increases $\mathfrak f$, $\mathfrak n$ and the terminal adjoint by at most
\[
\eta_{\mathrm{out}}B\sqrt{\lambda_\star}\,|g_j|,\qquad
\eta_{\mathrm{out}}B\,\varkappa_\ell,\qquad
\eta_{\mathrm{out}}BG_0\,|g_j|,
\]
respectively. In Lemma~\ref{lemma:boundary}, the difference of the output embeddings of two models grows by $\eta_{\mathrm{out}}G_0G_\ell$ times the hidden-state error, and $\eta_{\mathrm{out}}G_\ell^2\le C\bar\eta$, where the effective learning rate $\bar\eta$ contains $\eta_{\mathrm{out}}(\lambda_\star+\varkappa_\ell+G_0^2)=\varsigma$.
\end{enumerate}
Write
\[
\mathfrak s:=\sigma_{\mathrm{out}}\big(\sqrt{\tau_\star}+\Lambda_\ell+\mathfrak g_\ell+G_0\big),\qquad
\varsigma:=\eta_{\mathrm{out}}\big(\lambda_\star+\varkappa_\ell+G_0^2\big).
\]
Every quantity in the first item is at most $C\mathfrak s$. In the second item, the learning rate always multiplies the curvature or the gradient of the loss: $\eta_{\mathrm{out}}\lambda_\star$, $\eta_{\mathrm{out}}G_0^2$ and $\eta_{\mathrm{out}}\varkappa_\ell$ are at most $\varsigma$, and $\eta_{\mathrm{out}}\sqrt{\lambda_\star}G_0\le\frac12\varsigma$. In particular, Lemma~\ref{lemma:boundary} and the bound \eqref{eqn:DeltaW_bound} are stated in terms of $\tilde\eta_{\mathrm{out}}=\eta_{\mathrm{out}}(\lambda_\star+G_0^2)\le\varsigma$, and Lemma~\ref{lemma:InfNorm} only involves $\eta_{\mathrm{out}}G_\ell^2$ and $\eta_{\mathrm{out}}\sqrt{\lambda_\star}G_\ell$. The learning rate never enters on its own, so that it may be large when the curvature and the gradient of the loss are small. Finally, $G_\ell\le G_0+\sqrt{\lambda_\star}e^{R_*}S_*$, where $R_*$ and $S_*$ are bounded in terms of $\mathfrak s$, $\varsigma$ and the remaining constants by Lemma~\ref{lemma:apriori}, so that $\eta_{\mathrm{out}}G_0G_\ell$ and $\eta_{\mathrm{out}}G_\ell^2$ are bounded in terms of $\mathfrak s$ and $\varsigma$ as well. Hence all constants depend on the loss and the dimensions $D,d_{\mathrm{out}}$ only through upper bounds on $\mathfrak s$ and $\varsigma$, and are non-decreasing in these bounds. In particular, the dimension-dependent quantity $\mathfrak g_\ell=\mathcal{GW}(\|\cdot\|_\ell)/\sqrt D$ satisfies, by (L2),
\begin{equation}\label{eqn:gw_trivial}
\mathfrak g_\ell
\le\frac{\Lambda_\ell\mathbb E|g|}{\sqrt D}
\le\Lambda_\ell\sqrt{\frac{d_{\mathrm{out}}}{D}},
\end{equation}
so $\mathfrak g_\ell\le\Lambda_\ell$ whenever $d_{\mathrm{out}}\le D$, for any loss. This bound is weak, as we will see in Examples \ref{ex:MSE}-\ref{ex:CE}.

\subsection{Examples}\label{app:dout_examples}
\begin{example}[Mean squared error and the RMS distance]\label{ex:MSE}
Consider the loss
\[
\ell(y,\mathfrak t)=\frac1{2d_{\mathrm{out}}}|y-\mathfrak t|^2.
\]
Its Hessian is constant, so that $\|\cdot\|_\ell=0$, $\varkappa_\ell=\Lambda_\ell=\mathfrak g_\ell=\mathfrak n=0$, $C_{\mathscr H}=1$, $\tau_\star=1$, $\lambda_\star=\frac1{d_{\mathrm{out}}}$ and $G_0=\sup_{\mathfrak{t}\in \mathfrak{T}}|\mathfrak{t}|/d_{\mathrm{out}}$, and $\mathfrak f(W)^2=\frac{\|W\|_F^2}{Dd_{\mathrm{out}}}$ is the mean square of the entries of $W$. The action is minimised by the segment, and
\[
d_\ell(y,y')=\frac{|y-y'|}{\sqrt{d_{\mathrm{out}}}},
\]
the root mean square distance between the outputs. For targets with entries of order one, $G_0$ is of order $d_{\mathrm{out}}^{-1/2}$. Hence $\mathfrak s$ is of order one for $\sigma_{\mathrm{out}}=\mathcal O(1)$, and $\varsigma\approx\eta_{\mathrm{out}}/d_{\mathrm{out}}$. With $\eta_{\mathrm{out}}\asymp d_{\mathrm{out}}$, $\varsigma$ is of order one, so that the constants remain uniform in $d_{\mathrm{out}}$.
\end{example}

\begin{example}[Cross-entropy and the Fisher--Rao distance]\label{ex:CE}
For the cross-entropy loss with label $c$, the predicted class probabilities are $\pi(y):=\mathrm{softmax}(y)$. Then $\nabla\ell(y)=\pi(y)-e_c$, and $\nabla^2\ell(y)=\mathrm{diag}(\pi)-\pi\pi^\top$, which is the Fisher information of the categorical distribution in the logits, with kernel spanned by $(1,\ldots,1)$. Along any curve of logits $\gamma$, the corresponding probability curve $p:=\pi\circ\gamma$ satisfies
\[
\dot\gamma^\top\nabla^2\ell(\gamma)\dot\gamma
=\sum_a\frac{\dot p_a^2}{p_a}
=\sum_a\dot q_a^2,
\qquad q_a:=2\sqrt{p_a}.
\]
Since $\|q\|_2=2$, the Hessian-induced action is the Euclidean energy of a curve on the sphere of radius $2$. Its minimum is attained by the constant-speed shorter great-circle arc joining $2\sqrt{\pi(y)}$ and $2\sqrt{\pi(y')}$, which remains in the positive orthant. As a result, $d_\ell$ is the Fisher--Rao distance between the predicted class probabilities,
\[
d_\ell(y,y')
=2\arccos\Big(\sum_a\sqrt{\pi_a(y)\pi_a(y')}\Big),
\]
independently of the label. Let $\mathrm{osc}(u):=\max_au_a-\min_au_a$, and compare $\pi:=\pi(y)$ with $\pi':=\pi(y+u)$. Since $v^\top\mathscr H(y)v=\mathrm{Var}_{\pi}(v)$, the definition of $d_T$ reduces to comparing variances under $\pi$ and $\pi'$. As $\pi'_a=\pi_ae^{u_a}/\mathbb E_\pi e^u$, the ratio $\pi'_a/\pi_a$ lies in $[e^{-\mathrm{osc}(u)},e^{\mathrm{osc}(u)}]$, and so
\[
\mathrm{Var}_{\pi'}(v)\le\mathbb E_{\pi'}\big(v-\mathbb E_\pi v\big)^2\le e^{\mathrm{osc}(u)}\,\mathrm{Var}_{\pi}(v),
\]
and symmetrically. Hence $d_T(\mathscr H(y+u),\mathscr H(y))\le\mathrm{osc}(u)$. Using that $\mathrm{osc}$ is positively homogeneous,
\[
\|u\|_\ell\le\mathrm{osc}(u)\le2\max_a|u_a|.
\]
Therefore Assumption~\ref{assm:Loss} holds with
\[
\varkappa_\ell\le\sup_{y,c}\mathrm{osc}(\pi(y)-e_c)\le2,\qquad\Lambda_\ell\le2,\qquad\mathcal{GW}(\|\cdot\|_\ell)\le2\sqrt{2\log(2d_{\mathrm{out}})}.
\]
 Hence $\mathfrak g_\ell\le2\sqrt{2\log(2d_{\mathrm{out}})/D}$, and $C_{\mathfrak n}=C\sigma_{\mathrm{out}}\big(1+\sqrt{c_1}+\sqrt{\log(2d_{\mathrm{out}})/D}\big)$ in Lemma~\ref{lemma:embed_event}. Moreover, $\tau_\star=1-\frac1{d_{\mathrm{out}}}$, $\lambda_\star=\frac1{d_{\mathrm{out}}}$ and $G_0\le1$. Hence, $\mathfrak s\le\sigma_{\mathrm{out}}(4+\mathfrak g_\ell)$ and $\varsigma\le4\eta_{\mathrm{out}}$, so that readout entries and learning rate of order one suffice whenever $\log d_{\mathrm{out}}\lesssim D$, a condition met by every vocabulary used in practice. Otherwise, $\sigma_{\mathrm{out}}\asymp\sqrt{D/\log d_{\mathrm{out}}}$ restores $\mathfrak s=\mathcal O(1)$.
\end{example}

\vskip 0.2in
\bibliography{DeepResNet}

\end{document}